\documentclass[journal]{IEEEtran}

\usepackage{float}
\usepackage{graphicx}
\usepackage{multirow}
\usepackage{amsmath} 
\usepackage{textcomp}
\usepackage{tabularx}
\usepackage[normalem]{ulem}
\usepackage{booktabs}
\usepackage{footnote}
\usepackage{threeparttable}
\usepackage[colorlinks,linkcolor=red]{hyperref}
\usepackage{cite}
\usepackage{booktabs}
\usepackage{caption}
\usepackage[utf8]{inputenc}
\usepackage{amssymb}             
\usepackage{amsfonts}            
\usepackage{mathrsfs} 
\usepackage{CJKutf8}
\usepackage[table]{xcolor}
\usepackage[noend]{algpseudocode}
\usepackage{booktabs}
\usepackage{indentfirst}
\usepackage{threeparttable}
\usepackage{algorithmicx}
\usepackage{algorithm}
\usepackage{array}
\usepackage{stfloats}
\usepackage{url}
\usepackage{verbatim}
\usepackage{graphicx}
\usepackage{float}
\usepackage{booktabs} 
\usepackage{subfigure}
\usepackage{cite}
\usepackage{colortbl}
\usepackage{cleveref}
\usepackage{makecell}
\usepackage{rotating}
\newcommand{\Tr}{^{\mathrm{T}}}

\usepackage[colorinlistoftodos]{todonotes}

\begin{document}

\title{Robustness-Aware 3D Object Detection in Autonomous Driving: A Review and Outlook}

\author{Ziying Song, Lin Liu, Feiyang Jia, Yadan Luo, Caiyan Jia,  Guoxin Zhang, Lei Yang,  Li Wang
\thanks{This work was supported in part by the National Key R\&D Program of China (2018AAA0100302), supported by the STI 2030-Major Projects under Grant 2021ZD0201404.\emph{(Corresponding author: Caiyan Jia.)}}
\thanks{Ziying Song,  Lin Liu, Feiyang Jia, Caiyan Jia are with School of Computer Science \& Technology, Beijing Key Lab of Traffic Data Analysis and Mining, Beijing Jiaotong University, Beijing 100044, China (e-mail: 22110110@bjtu.edu.cn,  liulin010811@gmail.com, feiyangjia@bjtu.edu.cn, cyjia@bjtu.edu.cn)
}
\thanks{Yadan Luo is with the School of Information Technology and Electrical
Engineering, The University of Queensland, St Lucia, QLD 4072, Australia
(e-mail: uqyluo@uq.edu.au)
}
\thanks{Guoxin Zhang is with School of Computer Science, Beijing University of Posts and Telecommunications, Beijing 100876, China (e-mail: zhangguoxincs@gmail.com)
}
\thanks{Lei Yang is with the State Key Laboratory of Intelligent Green Vehicle and Mobility, and the School of Vehicle and Mobility, Tsinghua University, Beijing 100084, China (e-mail: yanglei20@mails.tsinghua.edu.cn).}
\thanks{Li Wang is with School of Mechanical Engineering, Beijing Institute of Technology, Beijing 100081, China (e-mail: wangli\_bit@bit.edu.cn.)}


}

\maketitle

\begin{abstract}
In the realm of modern autonomous driving, the perception system is indispensable for accurately assessing the state of the surrounding environment, thereby enabling informed prediction and planning. The key step to this system is related to 3D object detection that utilizes vehicle-mounted sensors such as LiDAR and cameras to identify the size, the category, and the location of nearby objects. Despite the surge in 3D object detection methods aimed at enhancing detection precision and efficiency, there is a gap in the literature that systematically examines their resilience against environmental variations, noise, and weather changes. This study emphasizes the importance of robustness, alongside accuracy and latency, in evaluating perception systems under practical scenarios. Our work presents an extensive survey of camera-only, LiDAR-only, and multi-modal 3D object detection algorithms, thoroughly evaluating their trade-off between accuracy, latency, and robustness, particularly on datasets like KITTI-C and nuScenes-C to ensure fair comparisons. Among these, multi-modal 3D detection approaches exhibit superior robustness, and a novel taxonomy is introduced to reorganize \textcolor{black}{the} literature for enhanced clarity. This survey aims to offer a more practical perspective on the current capabilities and the constraints of 3D object detection algorithms in real-world applications, thus steering future research towards robustness-centric advancements.

\end{abstract}

\begin{IEEEkeywords}
3D Object Detection,  Perception,   Robustness,  Autonomous Driving
\end{IEEEkeywords}

\IEEEpeerreviewmaketitle

\section{Introduction}

\begin{figure*}[!t]
	\centering
	\includegraphics[width=\linewidth]{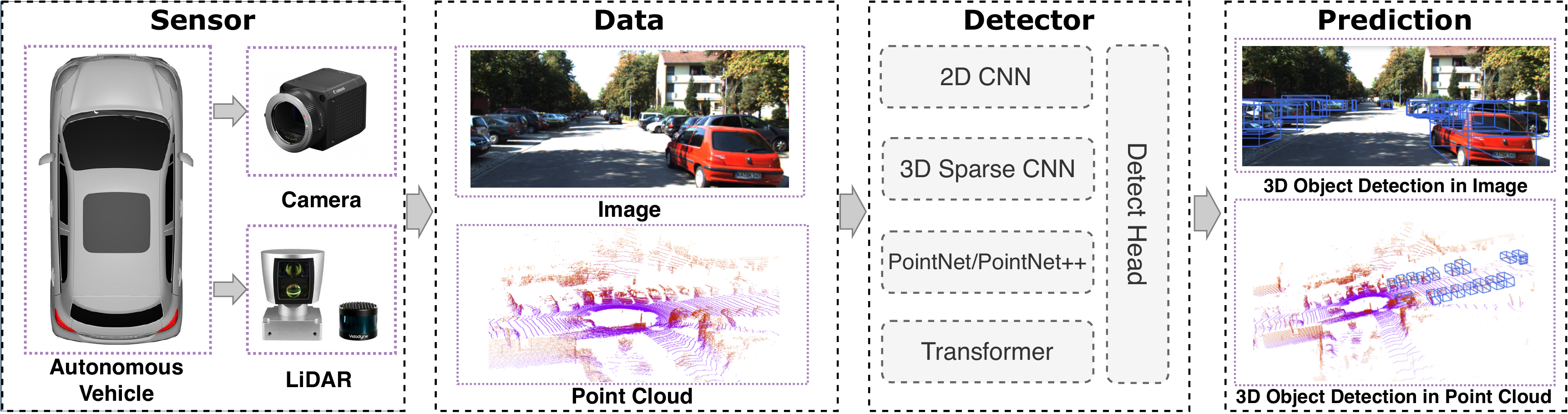}
	\caption{An illustration of 3D object detection in autonomous driving scenarios with different sensors.}
	\label{fig:vis_data}
\end{figure*}

\IEEEPARstart{A}{UTONOMOUS} driving systems, fundamental to the future of transportation, heavily rely on advanced perception, decision-making, and control technologies. These systems employ a range of sensors~\cite{arnold2019survey} such as camera, LiDAR and radar as depicted in Fig. \ref{fig:vis_data}, to effectively perceive surrounding environments. This capability is crucial for recognizing road signs, detecting and tracking vehicles, and predicting pedestrian behaviors, enabling safe operations amidst complex traffic conditions \cite{wanghong_PNNUAD_TITS}. 

The primary task of perception is to accurately understand the surrounding environment and minimize collision risks \cite{wanghong_Prediction_TITS}. This is where 3D object detection methods become essential. These approaches enable the autonomous systems to accurately identify objects in the vicinity, including their position, shape, and category~\cite{wang2023multi}. Such detailed environmental perception enhances the system's ability to comprehend the driving context and make more informed decisions.

The advancement of autonomous driving technologies has spared a wave of research in 3D object detection, leading to the development of diverse and innovative methods. These approaches are typically categorized based on their input types, including camera-only ~\cite{cao2022cman,pseudoLiDAR,li2020rtm3d,zhang2021objects,simonelli2019disentangling,lian2022monojsg,brazil2019m3d,cai2020monocular,chen2020monopair,liu2021yolostereo3d,bevdepth,bevformer,chen2020dsgn,liu2019deep,shi2020distance,ma2019accurate,ku2019monocular,CaDDN,MonoDTR,wang2021fcos3d,wang2022probabilistic,lu2021geometry,park2021pseudo,yang2023mix,zhang2022monodetr,wang2022detr3d,liu2022petr,liu2023petrv2,huang2022bevdet4d,liu2023sparsebev,yang2023bevheight++}, LiDAR-only~\cite{vpnet,CenterNet3D,AFDetV2,afdet,3dssd,3dcvf,Voxelnet,Second,Point-gnn,Pointrcnn,voxelnext,Voxelrcnn,VoxelFPN,pvcnn,pvrcnn++,Pointpillars,pillarNeXt,PDV,TED,Sparsetransformer,FSD,Centerpoint,voxeltransformer,vsettransformer,LargeKernel3D,Link,lai2023spherical,svganet,SWFormer,UVTR,3dobwithpointformer,CT3D,Behind_the_curtain,SE-SSD,ipod,Associate-3Ddet,vtop,Sienet,zhang2023simple,CBGS,hvnet,Cia-ssd,cvcnet2020,Reconfigurable_Voxels,DGCNN,SSN,PVGNet,HVPR,M3DETR,infofocus,pyramidrcnn,rangercnn,rangecd,rangeioudet,RangeDet,Seformer,casa,pillarnet,sasa,rbgnet,Frustumconvnet,tang2020searching,Std}, and multi-modal methods~\cite{virconv,Pointpainting,sparsefusion,SparseFusion3D,cmt,Epnet,epnetpami,FUTR3D,MVF,Transfusion,BEVFusion,UVTR,focalconv,graphalign,graphalign++,Graph-RCNN,sfd,DeepFusion,DeepInteraction,ObjectFusion,SupFusion,qian20223d,HMFI,Logonet,VoxelNextFusion}. The current landscape of 3D object detection methods is prolific, necessitating a comprehensive summarization to offer intriguing insights \textcolor{black}{to} the research community. While comprehensive, prior surveys, \textcolor{black}{such as} \cite{wang2023multi,wang2023multi_ijcv}, often overlook the safety aspects of autonomous driving perception, particularly in terms of the system's robustness against varying testing data \textcolor{black}{following} deployment.

In real-world testing scenarios, the conditions encountered usually greatly differ from those during training. The environmental variability, sensor discrepancies or noise, and spatial misalignment can cause a shift in the input sensory data distribution, leading to a significant drop in detector performance \cite{graphalign,Transfusion,RoboBEV,BR3D}. We identify and discuss three major factors critical for assessing detection \textbf{robustness}.
\begin{itemize}
\item \textbf{Environmental Variability}. A detection algorithm needs to perform well under different environmental conditions, including variations in lighting, weather, and seasons. The algorithms should exhibit adaptability, ensuring that it does not fail due to changes in the environment.
\item \textbf{Sensor Noise}. This includes handling noise introduced by sensor malfunctions, such as motion blur \textcolor{black}{in a} camera. An algorithm must possess the capability to effectively manage hardware noise, ensuring \textcolor{black}{the} accurate processing of input data.
\item \textbf{Misalignment}. In real-world scenarios, sensor calibration errors can complicate the synchronization of multi-modal input data, causing misalignment due to external factors (e.g., uneven road surfaces) or internal factors (e.g., system clock misalignment). An algorithm should be fault-tolerant and may incorporate an elastic alignment to mitigate \textcolor{black}{the impact of misalignment}  on detection performance.
\end{itemize}


To ensure safe operation in varying test environments, assessing the robustness of 3D object detection algorithms is essential. They must maintain efficient, accurate, and reliable performance across diverse scenarios. In this survey, we conduct extensive experimental comparisons among existing algorithms. Centered around \textcolor{black}{`Accuracy, Latency, Robustness',} we delve into existing solutions, offering insightful guidance for practical deployment in autonomous driving.
\begin{itemize}
\item  \textit{\textbf{Accuracy:}}
Current researches often prioritize accuracy as a key performance metric. However, a deeper understanding of these methods' performance in complex environments and extreme weather conditions is needed to ensure real-world reliability. A more detailed analysis of false positives and false negatives is necessary for improvement.

\item \textit{\textbf{Latency:}}
Real-time capability is vital for autonomous driving. The latency of a 3D object detection method impacts the system's ability to make timely decisions, particularly in emergencies.

\item \textit{\textbf{Robustness:}}
Robustness refers to the system's stability under various conditions, including weather, lighting, sensory and alignment changes. Many existing evaluations may not fully consider the diversity of real-world scenarios, necessitating a more comprehensive adaptability assessment.

 \end{itemize}

Through an in-depth analysis of extensive experimental results, with a focus on \textcolor{black}{`Accuracy, Latency, Robustness',} we have identified significant advantages in safety perception with Multi-modal 3D detection in safety perception. By integrating information from diverse sensors or data sources, Multi-modal methods provide a richer and more diverse perception capability for autonomous driving systems, thereby enhancing the understanding and responding to the surrounding environment. Our research provides practical guidance for the future deployment of autonomous driving technology. By discussing these key areas, we aim to align the technology more closely with real-world needs and enhance its societal benefits effectively.

The structure of this paper is organized as follows: First, we introduce the datasets and evaluation metrics for 3D object detection, with a particular focus on robustness in Section \ref{section:dataset}. \textcolor{black}{Subsequent sections} systematically examine existing 3D object detection methods, including camera-only approaches (Section \ref{section:camera-based}), LiDAR-only approaches (Section \ref{section:LiDAR-only}), and multi-modal approaches (Section \ref{section:multimodal-based}). The paper concludes with a comprehensive summary of our findings \textcolor{black}{in Section} \ref{section:conclusion}.


\section{Datasets}
\label{section:dataset}

\begin{table}[!t]
\scriptsize
\renewcommand\arraystretch{1.2}
\centering
\caption{Advantages and limitations of different modalities.}
\label{table:Advantages_Limitations}
\setlength{\tabcolsep}{0.4mm}{
\begin{tabular}{p{1.25cm}p{1cm}p{1.1cm}p{2.6cm}p{2.6cm}} \hline
\textbf{Type} & \textbf{Sensor} & \textbf{Hardware Cost(\$)} & \multicolumn{1}{c}{\textbf{Advantages}} & \multicolumn{1}{c}{\textbf{Limitations}} \\ \hline
Image & Camera &$10^{2}$\~{}$10^{3}$& + The dense data format incorporates additional color and texture information. & - Missing depth information the camera will be affected by light, weather, etc. \\
Point cloud & LiDAR     &$10^{4}$\~{}$10^{5}$& + With accurate depth information less affected by light +larger field of view & -High computational cost for sparse and disordered point cloud data and no color information. \\
\multirow{2}{*}{Multi-modal} &Camera, LiDAR & $10^{4}$\~{}$10^{5}$& + Simultaneous color and depth information & - Fusion methods can produce noise interference \\ \hline
\end{tabular}}
\end{table}

Currently, autonomous driving systems primarily rely on sensors such as cameras, and LiDAR, generating data in two modalities, point clouds and images. Based on these data types, existing public benchmarks predominantly manifest in three forms: camera-only, LiDAR-only, and multi-modal. Table \ref{table:Advantages_Limitations} delineates the advantages and the disadvantages of each of these three forms. Among them, there are many reviews \cite{mao20233d,wang2023multi_ijcv,alaba2023deep,singh2023surround,singh2023transformer,wang2023multi_TITS,peng2022survey,wu2020deep} providing a comprehensive overview of \textbf{clean} \textcolor{black}{autonomous driving datasets as shown in Table \ref{table:datasets}.} The most notable ones include KITTI\cite{kitti}, nuScenes\cite{nuscenes}, and Waymo\cite{Waymo}. 

In recent times, the pioneering work on clean autonomous driving datasets has provided rich resources for 3D object detection. As autonomous driving technology transitions from breakthrough stages to practical implementation, we have \textcolor{black}{conducted} some guided researches to \textcolor{black}{systematically} review the currently available robustness datasets. We \textcolor{black}{have focused} more on noisy scenarios and systematically reviewed datasets related to the robustness of 3D detection. Many studies \textcolor{black}{have collected} new datasets to evaluate model robustness under different conditions. Early research explored camera-only approaches under adverse conditions \cite{sakaridis2018semantic,cai2016dehazenet}, with datasets that were notably small in scale and exclusively applicable to camera-only visual tasks, rather than multi-modal sensor stacks that include LiDAR. Subsequently, a series of multi-modal datasets \cite{ort2021grounded,pitropov2021canadian,apolloscape,diaz2022ithaca365} have focused on noise concerns. For instance, the GROUNDED dataset \cite{ort2021grounded} focuses on ground-penetrating radar localization under varying weather conditions. Additionally, the ApolloScape open dataset \cite{apolloscape} incorporates LiDAR, camera, and GPS data, encompassing cloudy and rainy conditions as well as brightly lit scenarios. \textcolor{black}{The Ithaca365 dataset \cite{diaz2022ithaca365} is designed for robustness in autonomous driving research, providing scenarios under various challenging weather conditions, such as rain and snow.}

Due to the prohibitive cost of collecting extensive noisy datasets from the real world, rendering the formation of large-scale datasets impractical, many studies have shifted their focus to synthetic datasets. ImageNet-C \cite{hendrycks2019benchmarking} is a seminal work in corruption robustness research, benchmarking classical image classification models against prevalent corruptions and perturbations. This line of research has subsequently \textcolor{black}{been} extended to include robustness datasets tailored for 3D object detection in autonomous driving. Additionally, there are adversarial attacks \cite{xie2023adversarial,sun2020towards,liu2019extending} designed for studying the robustness of 3D object detection. However, these attacks may not exclusively concentrate on natural corruption, which is less \textcolor{black}{common} in autonomous driving scenarios.

To better emulate the distribution of noise data in the real world, several studies \cite{BR3D,robotrans,BRLCF,RoboBEV,kong2023robo3d,kong2023robodepth} have developed toolkits for robustness benchmarks. These benchmark toolkits \cite{BR3D,robotrans,BRLCF,RoboBEV,kong2023robo3d,kong2023robodepth} enable the simulation of various scenarios using clean autonomous driving datasets, such as KITTI \cite{kitti}, nuScenes \cite{nuscenes}, and Waymo \cite{Waymo}. Among them, Dong et al.\cite{BR3D} systematically designed 27 common corruptions in 3D object detection to benchmark the corruption robustness of existing detectors. By applying these corruptions comprehensively on public datasets, they established three corruption-robust benchmarks: KITTI-C, nuScenes-C, and Waymo-C. \cite{BR3D} denotes model performance on the original validation set as $AP_{\text{clean}}$. For each corruption type $c$ at each severity $s$, \cite{BR3D} adopts the same metric to measure model performance as $AP_{\text{c,s}}$. The corruption robustness of a model is calculated by averaging over all corruption types and severities as 
\begin{equation}
\begin{aligned}
\Lambda \mathrm{P}_{\text {cor }}=\frac{1}{|\mathcal{C}|} \sum_{c \in \mathcal{C}} \frac{1}{5} \sum_{s=1}^{5} \Lambda \mathrm{P}_{c, s}.
\end{aligned}
\end{equation}
Where $\mathcal{C}$ is the set of corruptions in evaluation. It should be noticed that for different kinds of 3D object detectors, the set of corruptions can be different (e.g., \cite{BR3D} has not evaluated camera noises for LiDAR-only models). Thus, the results of APcor are not directly comparable between different kinds of models, and \cite{BR3D} performs a fine-grained analysis under each corruption. 
It also calculates relative corruption error (RCE) by measuring the percentage of performance drop as
\begin{equation}
\begin{aligned}
\mathrm{RCE}_{c, s}=\frac{\mathrm{AP}_{\text {clean }}-\mathrm{AP}_{c, s}}{\mathrm{AP}_{\text {clean }}} ; \mathrm{RCE}=\frac{\mathrm{AP}_{\text {clean }}-\mathrm{AP}_{\text {cor }}}{\mathrm{AP}_{\text {clean }}}.
\end{aligned}
\label{equ:rce}
\end{equation}

Unlike KITTI-C and Waymo-C, nuScenes-C primarily assesses performance using the mean Average Precision (mAP) and nuScenes Detection Score (NDS) computed across ten object categories. The mAP is determined using the 2D center distance on the ground plane instead of the 3D Intersection over Union (IoU). The NDS metric consolidates mAP with other aspects, such as scale and orientation, into a unified score. Analogous to KITTI-C, 
\cite{BR3D} denotes the model's performance on the validation set as $\text{mAP}_{\text{clean}}$ and $\text{NDS}_{\text{clean}}$, respectively. The corruption robustness metrics, $\text{mAP}_{\text{cor}}$ and $\text{NDS}_{\text{cor}}$, are evaluated by averaging over all corruption types and severities. Additionally, 
\cite{BR3D} calculates the Relative Corruption Error (RCE) under both mAP and NDS metrics, similar to the formulation in Eq.\ref{equ:rce}.

\begin{table}[t]
\centering
\caption{Public datasets for 3D object detection in autonomous driving. `C', `L' and `R' denote Camera, LiDAR and Radar, respectively.}
\label{table:datasets}
\renewcommand\arraystretch{1.0}
\setlength{\tabcolsep}{1.0mm}{
\begin{tabular}{lcccccc}
\toprule
\multirow{2}{*}{\textbf{Dataset}} & \multirow{2}{*}{\textbf{Year}} & \multirow{2}{*}{\textbf{Sensors}} & \multicolumn{2}{c}{\textbf{Data Size}} & \multicolumn{2}{c}{\textbf{Diversity}}   \\ \cmidrule{4-5}  \cmidrule{6-7}
&  & & Frame & Annotation & Scenes & Category \\ 
\midrule
KITTI\cite{kitti} & 2012 & CL& 15K & 200K & 50 & 3  \\


nuScenes\cite{nuscenes} & 2019  & CLR & 40K & 1.4M & 1000 & 10  \\
Lyft L5\cite{kesten2019lyft} & 2019 & CL& 46K & 1.3M & 366 & 9 \\
H3D\cite{h3d} & 2019 & L & 27K & 1.1M & 160 & 8 \\
Appllo\cite{apolloscape} & 2019 & CL& 140K & - & 103 & 27 \\

Argoverse\cite{argoverse} & 2019 & CL& 46K & 993K & 366 & 9   \\
A*3D \cite{pham20203d} & 2019 & CL& 39K & 230K & - & 7 \\

Waymo\cite{Waymo} & 2020 & CL& 230K & 12M & 1150 & 3 \\ 

A2D2 \cite{geyer2020a2d2} & 2020 & CL& 12.5K & 43K & - & 38  \\
PandaSet \cite{xiao2021pandaset} & 2020 & CL& 14K & - & 179 & 28  \\
KITTI-360 \cite{liao2022kitti} & 2020 & CL& 80K & 68K & 11 & 19\\
Cirrus \cite{wang2021cirrus} & 2020 & CL& 6285 & - & 12 & 8 \\

ONCE \cite{mao2021one} & 2021 & CL& 15K & 417K & - & 5  \\

OpenLane \cite{chen2022persformer} & 2022 & CL& 200K & - & 1000 & 14\\ 
\bottomrule
\end{tabular}}
\end{table}

Additionally, some studies \cite{xie2023adversarial,robotrans,ren2022benchmarking} examine robustness in single-modal contexts. For instance, 
\cite{robotrans} proposes a LiDAR-only benchmark that utilizes \textcolor{black}{physically-aware} simulation methods to simulate degraded point clouds under various real-world common corruptions. This benchmark, tailored for point cloud detectors, includes 1,122,150 examples across 7,481 scenes, covering 25 common corruption types with six severity levels. Moreover, 
\cite{robotrans} \textcolor{black}{devises} a novel evaluation \textcolor{black}{metric}, including $\text{CE}_{\text{AP}}(\%)$ and \textbf{mCE}, 
and calculates corruption error (CE) to assess performance degradation based on Overall Accuracy (OA) by
\begin{equation}
\begin{aligned}
    \mathrm{CE}_{\mathrm{c}, \mathrm{s}}^{\mathrm{m}}=\mathrm{OA}_{\mathrm{clean}}^{\mathrm{m}}-\mathrm{OA}_{\mathrm{c}, \mathrm{s}}^{\mathrm{m}},
\end{aligned}
\label{equ:CEAP}
\end{equation}
where $\mathrm{OA}_{\mathrm{c}, \mathrm{s}}^{\mathrm{m}}$ is the overall accuracy of detector m under corruption $\mathrm{c}$ of severity level $\mathrm{s}$ (excluding ``clean," i.e., severity level 0) and clean represents the clean data. For detector m, we can calculate the mean CE ($\mathrm{mCE}$) for each detector by
\begin{equation}
\begin{aligned}
\mathrm{mCE}^{\mathrm{m}}=\frac{\sum_{\mathrm{s}=1}^{5} \sum_{\mathrm{c}=1}^{25} \mathrm{CE}_{\mathrm{c}, \mathrm{s}}^{\mathrm{m}}}{5 \mathrm{C}}.
\end{aligned}
\label{equ:mCE}
\end{equation}

\section{Camera-only 3D Object Detection}
\label{section:camera-based}

\begin{figure}[!t]
	\centering
	\includegraphics[width=\linewidth]{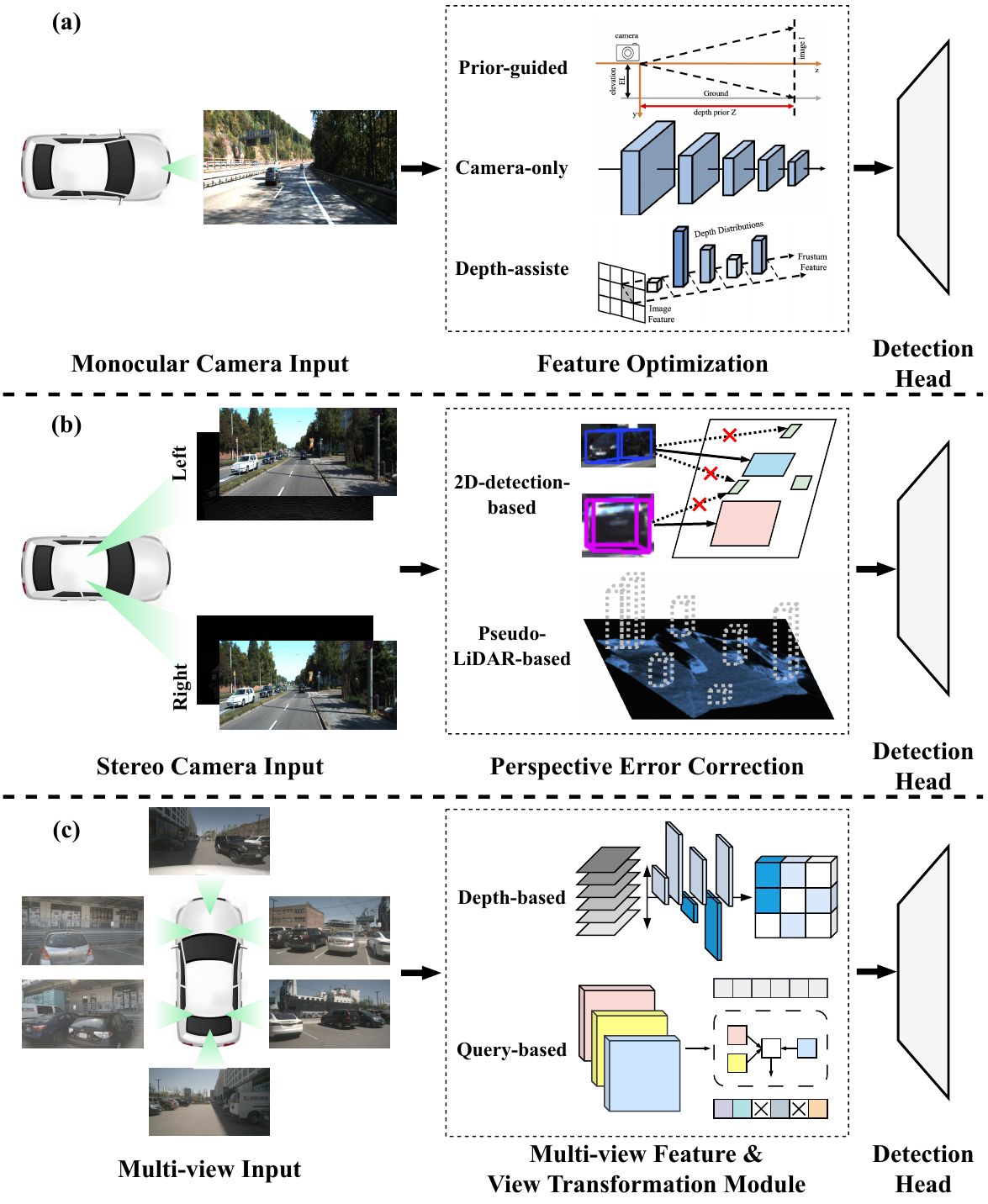}
	\caption{The general pipeline of Camera-only methods.}
	\label{fig:camera_only_pipeline}
\end{figure}

In this section, we introduce the Camera-only 3D object detection methods. 
Compared to LiDAR-only methods, the camera solution is more cost-effective and the images obtained from cameras require no complex preprocessing. 
Therefore, it is favored by many automotive manufacturers, particularly in the context of multi-view applications such as BEV (bird's-eye view) systems. 
Generally, as shown in Fig. \ref{fig:camera_only_pipeline}, Camera-only methods can be categorized into three types, monocular, stereo-based, and multi-view (bird's-eye view). 
Due to the excellent cost-effectiveness of Camera-only methods, there have been numerous reviews and investigations conducted to summarize and explore them. However, the majority of existing reviews on 3D object detection are limited to specific methodologies, with a predominant focus on accuracy. 
This survey aims to revisit the fundamental considerations of safety-perception deployment, redefining the discourse around existing categorizations, 
and exploring \textcolor{black}{`Accuracy, Latency, and Robustness',} as the core dimensions for an in-depth analysis of current methodologies. The objective is to provide additional insights to guide \textcolor{black}{the development of} existing technologies.

\begin{table*}[!t]
\scriptsize
\centering
\caption{Camera-only 3D object detection methods.}
\label{table:camera-based}
\renewcommand\arraystretch{1.2}
\setlength{\tabcolsep}{1.2mm}{
\begin{tabular}{p{1.2cm}p{4.5cm}p{11.6cm}}
\toprule
\textbf{Input Type} & \textbf{Keypoint} & \textbf{Methods} \\ \midrule
\multirow{12}{*}{Monocular} & 
\textbf{Prior-guided}: Direct regression using geometric prior knowledge. & 
Deep MANTA \textcolor[HTML]{708090}{\tiny{[CVPR2017]}}\cite{chabot2017deep}, 
Mono3D++ \textcolor[HTML]{708090}{\tiny{[AAAI2019]}} \cite{he2019mono3dplus}, 
3D-RCNN \textcolor[HTML]{708090}{\tiny{[CVPR2018]}} \cite{kundu20183d}, 
ROI-10D \textcolor[HTML]{708090}{\tiny{[CVPR2019]}}\cite{manhardt2019roi}, 
MonoDR \textcolor[HTML]{708090}{\tiny{[ECCV2020]}}\cite{beker2020monocular}, 
Autolabeling \textcolor[HTML]{708090}{\tiny{[CVPR2020]}}\cite{zakharov2020autolabeling}, 
MonoPSR \textcolor[HTML]{708090}{\tiny{[CVPR2019]}}\cite{ku2019monocular}, 
3DVP \textcolor[HTML]{708090}{\tiny{[CVPR2015]}}\cite{xiang2015data}, 
MultiBin \textcolor[HTML]{708090}{\tiny{[CVPR2017]}}\cite{mousavian20173d}, 
M3D-RPN \textcolor[HTML]{708090}{\tiny{[ICCV2019]}}\cite{brazil2019m3d}, 
SHIFT R-CNN \textcolor[HTML]{708090}{\tiny{[ICIP2019]}}\cite{naiden2019shift}, 
RTM3D \textcolor[HTML]{708090}{\tiny{[ECCV2020]}}\cite{li2020rtm3d}, 
UR3D \textcolor[HTML]{708090}{\tiny{[ECCV2020]}}\cite{shi2020distance}, 
Decoupled-3D \textcolor[HTML]{708090}{\tiny{[AAAI2020]}}\cite{cai2020monocular}, 
GUP Net \textcolor[HTML]{708090}{\tiny{[ICCV2021]}}\cite{lu2021geometry}, 
MonoFlex \textcolor[HTML]{708090}{\tiny{[CVPR2021]}}\cite{zhang2021objects}, 
Mix-Teaching \textcolor[HTML]{708090}{\tiny{[TCSVT2023]}}\cite{yang2023mix}, 

MonoPair \textcolor[HTML]{708090}{\tiny{[CVPR2020]}}\cite{chen2020monopair},
MonoJSG \textcolor[HTML]{708090}{\tiny{[CVPR2022]}}\cite{lian2022monojsg},
Geo Aug \textcolor[HTML]{708090}{\tiny{[CVPR2022]}}\cite{lian2022exploring},
Monoground \textcolor[HTML]{708090}{\tiny{[CVPR2022]}}\cite{qin2022monoground},
MonoPGC \textcolor[HTML]{708090}{\tiny{[ICRA2023]}}\cite{wu2023monopgc},
MonoEdge \textcolor[HTML]{708090}{\tiny{[WACV2023]}}\cite{zhu2023monoedge},
GPro3D \textcolor[HTML]{708090}{\tiny{[Neurocomputing2023]}}\cite{yang2023gpro3d},
MonoGAE \textcolor[HTML]{708090}{\tiny{[arXiv2023]}}\cite{yang2023monogae},
GUPNet++ \textcolor[HTML]{708090}{\tiny{[arXiv2023]}}\cite{lu2023gupnet++},
NeurOCS \textcolor[HTML]{708090}{\tiny{[CVPR2023]}}\cite{min2023neurocs}. 
\\ \cline{2-3}
& \textbf{Camera-only}: uses the RGB image information captured by the monocular.& 
Smoke \textcolor[HTML]{708090}{\tiny{[CVPR2020]}} \cite{liu2020smoke},
Kinematic3D \textcolor[HTML]{708090}{\tiny{[ECCV2020]}} \cite{brazil2020kinematic},
FQNet \textcolor[HTML]{708090}{\tiny{[CVPR2019]}} \cite{liu2019deep},
FCOS3D \textcolor[HTML]{708090}{\tiny{[CVPR2021]}} \cite{wang2021fcos3d}, 
PGD \textcolor[HTML]{708090}{\tiny{[CoRL2022]}} \cite{wang2022probabilistic}, 
CaDDN \textcolor[HTML]{708090}{\tiny{[CVPR2021]}} \cite{CaDDN},
MoVi-3D \textcolor[HTML]{708090}{\tiny{[ECCV2020]}} \cite{simonelli2020towards},
MonoDIS \textcolor[HTML]{708090}{\tiny{[ICCV2019]}} \cite{simonelli2019disentangling},
GS3D \textcolor[HTML]{708090}{\tiny{[CVPR2019]}} \cite{li2019gs3d},
MonoGRNet \textcolor[HTML]{708090}{\tiny{[TPAMI2021]}} \cite{qin2019monogrnet}, 
MonoRCNN \textcolor[HTML]{708090}{\tiny{[ICCV2021]}} \cite{shi2021geometry},
MonoFENet \textcolor[HTML]{708090}{\tiny{[TIP2019]}} \cite{bao2019monofenet},
MonoCon \textcolor[HTML]{708090}{\tiny{[AAAI2022]}} \cite{liu2022learning},
MonoXiver \textcolor[HTML]{708090}{\tiny{[ICCV2023]}} \cite{liu2023monocular},
SGM3D \textcolor[HTML]{708090}{\tiny{[RAL2022]}} \cite{zhou2022sgm3d},
MonoDETR \textcolor[HTML]{708090}{\tiny{[ICCV2023]}} \cite{zhang2022monodetr},
MonoDTR \textcolor[HTML]{708090}{\tiny{[CVPR2022]}} \cite{MonoDTR},
DiD-M3D \textcolor[HTML]{708090}{\tiny{[ECCV2022]}} \cite{peng2022did},
MonoNeRD \textcolor[HTML]{708090}{\tiny{[ICCV2023]}} \cite{xu2023mononerd},
MonoSAID \textcolor[HTML]{708090}{\tiny{[IRS2024]}} \cite{xia2024monosaid},
WeakMono3D \textcolor[HTML]{708090}{\tiny{[CVPR2023]}} \cite{tao2023weakly},
DDCDC \textcolor[HTML]{708090}{\tiny{[Neurocomputing2023]}} \cite{wu2023depth},
Obmo \textcolor[HTML]{708090}{\tiny{[TIP2023]}} \cite{huang2023obmo},
Shape-Aware \textcolor[HTML]{708090}{\tiny{[TITS2023]}} \cite{chen2021shape},
Lite-FPN \textcolor[HTML]{708090}{\tiny{[KBS2023]}} \cite{yang2023litefpn},
OOD-M3D \textcolor[HTML]{708090}{\tiny{[TCE2024]}} \cite{park2024odd},
MonoTDP \textcolor[HTML]{708090}{\tiny{[arXiv2023]}} \cite{li2023monotdp},
Cube R-CNN \textcolor[HTML]{708090}{\tiny{[CVPR2023]}} \cite{brazil2023omni3d},
M2S \textcolor[HTML]{708090}{\tiny{[TIP2023]}} \cite{kim2023stereoscopic}.
\\ \cline{2-3}
& \textbf{Depth-assisted}: extracting depth information via camera parallax.& 
PatchNet \textcolor[HTML]{708090}{\tiny{[ECCV2020]}} \cite{RPLR},
DD3D \textcolor[HTML]{708090}{\tiny{[ICCV2021]}} \cite{park2021pseudo},
Pseudo-LiDAR \textcolor[HTML]{708090}{\tiny{[CVPR2019]}} \cite{pseudoLiDAR},
DeepOptics \textcolor[HTML]{708090}{\tiny{[ICCV2019]}} \cite{chang2019deep},
AM3D \textcolor[HTML]{708090}{\tiny{[ICCV2019]}} \cite{ma2019accurate},
MonoTAKD \textcolor[HTML]{708090}{\tiny{[arXiv2024]}} \cite{liu2024monotakd},
MonoPixel \textcolor[HTML]{708090}{\tiny{[TITS2022]}} \cite{kim2022boosting},
DDMP-3D \textcolor[HTML]{708090}{\tiny{[CVPR2021]}} \cite{wang2021depth},
D4LCN \textcolor[HTML]{708090}{\tiny{[CVPRW2020]}} \cite{ding2020learning},
ADD \textcolor[HTML]{708090}{\tiny{[AAAI2023]}} \cite{wu2023attention},
PDR \textcolor[HTML]{708090}{\tiny{[TCSVT2023]}} \cite{sheng2023pdr},
Pseudo-Mono \textcolor[HTML]{708090}{\tiny{[ECCV2022]}} \cite{tao2023pseudo},
Deviant \textcolor[HTML]{708090}{\tiny{[ECCV2022]}} \cite{kumar2022deviant},
CMAN \textcolor[HTML]{708090}{\tiny{[TITS2022]}} \cite{cao2022cman},
ODM3D \textcolor[HTML]{708090}{\tiny{[WACV2024]}} \cite{zhang2024alleviating},
MonoGAE \textcolor[HTML]{708090}{\tiny{[arXiv2023]}} \cite{yang2023monogae},
FD3D \textcolor[HTML]{708090}{\tiny{[AAAI2023]}} \cite{wu2024fd3d},
MonoSKD \textcolor[HTML]{708090}{\tiny{[arXiv2023]}} \cite{wang2023monoskd}.
\\ \hline
\multirow{12}{*}{Stereo} & \textbf{2D-Detection-based}: Integrate 2D information about the object into the image. & 
Disp R-CNN \textcolor[HTML]{708090}{\tiny{[CVPR2020]}} \cite{disprcnn},
TL-Net \textcolor[HTML]{708090}{\tiny{[CVPR2019]}} \cite{qin2019triangulation},
ZoomNet \textcolor[HTML]{708090}{\tiny{[AAAI2020]}} \cite{xu2020zoomnet}, 
IDA-3D \textcolor[HTML]{708090}{\tiny{[CVPR2020]}} \cite{peng2020ida}, 
YOLOStereo3D \textcolor[HTML]{708090}{\tiny{[ICRA2021]}} \cite{liu2021yolostereo3d},
SIDE \textcolor[HTML]{708090}{\tiny{[WACV2022]}} \cite{peng2022side},
VPFNet \textcolor[HTML]{708090}{\tiny{[TMM2022]}} \cite{VPFNet},
FCNet \textcolor[HTML]{708090}{\tiny{[Entropy2022]}} \cite{wu2022fcnet},
MC-Stereo \textcolor[HTML]{708090}{\tiny{[arXiv2023]}} \cite{feng2023mc},
PCW-Net \textcolor[HTML]{708090}{\tiny{[ECCV2022]}} \cite{shen2022pcw},
ICVP \textcolor[HTML]{708090}{\tiny{[ICIP2023]}} \cite{kwon2023icvp},
MoCha-Stereo \textcolor[HTML]{708090}{\tiny{[arXiv2024]}} \cite{chen2024mocha},
UCFNet \textcolor[HTML]{708090}{\tiny{[TPAMI2023]}} \cite{shen2023digging},
IGEV-Stereo \textcolor[HTML]{708090}{\tiny{[CVPR2023]}} \cite{xu2023iterative},
NMRF-Stereo \textcolor[HTML]{708090}{\tiny{[arXiv2024]}} \cite{guan2024neural}.
\\ \cline{2-3}

&\textbf{Pseudo-LiDAR-only}: incorporate additional information from pseudo-LiDAR to simulate LiDAR depth. & 
Pseudo-LiDAR \textcolor[HTML]{708090}{\tiny{[CVPR2019]}} \cite{pseudoLiDAR},
Pseudo-LiDAR++ \textcolor[HTML]{708090}{\tiny{[ICLR2020]}} \cite{pseudoLiDAR++},
E2E-PL \textcolor[HTML]{708090}{\tiny{[CVPR2020]}} \cite{qian2020end},
CG-Stereo \textcolor[HTML]{708090}{\tiny{[IROS2020]}} \cite{CGStereo3D},
SGM3D \textcolor[HTML]{708090}{\tiny{[RAL2022]}} \cite{zhou2022sgm3d},
RTS3D \textcolor[HTML]{708090}{\tiny{[AAAI2023]}} \cite{li2021rts3d},
RT3DStereo \textcolor[HTML]{708090}{\tiny{[ITS2019]}} \cite{RT3DOD},
RT3D-GMP  \textcolor[HTML]{708090}{\tiny{[ITSC2020]}} \cite{konigshof2020learning},
CDN \textcolor[HTML]{708090}{\tiny{[NIPS2020]}} \cite{garg2020cdn}.
\\ \cline{2-3}
& \textbf{Volume-based}: perform 3D object detection directly on 3D stereo volumes. & 
GC-Net \textcolor[HTML]{708090}{\tiny{[ICCV2017]}} \cite{kendall2017end},
ESGN \textcolor[HTML]{708090}{\tiny{[TCSVT2022]}} \cite{gao2022esgn},
DSGN \textcolor[HTML]{708090}{\tiny{[CVPR2020]}} \cite{chen2020dsgn},
DSGN++ \textcolor[HTML]{708090}{\tiny{[TPAMI2022]}} \cite{chen2022dsgn++},
LIGA-Stereo \textcolor[HTML]{708090}{\tiny{[ICCV2021]}} \cite{guo2021liga},
PLUMENet \textcolor[HTML]{708090}{\tiny{[IROS2021]}} \cite{wang2021plumenet},
Selective-IGEV \textcolor[HTML]{708090}{\tiny{[arXiv2024]}} \cite{wang2024selective},
ViTAS \textcolor[HTML]{708090}{\tiny{[arXiv2024]}} \cite{liu2024playing},
LaC+GANet \textcolor[HTML]{708090}{\tiny{[AAAI2022]}} \cite{liu2022local},
DMIO \textcolor[HTML]{708090}{\tiny{[arXiv2024]}} \cite{shi2024rethinking},
HCR \textcolor[HTML]{708090}{\tiny{[IVC2024]}} \cite{yuan2024hourglass},
LEAStereo \textcolor[HTML]{708090}{\tiny{[NIPS2020]}} \cite{cheng2020hierarchical},
CREStereo \textcolor[HTML]{708090}{\tiny{[CVPR2022]}} \cite{li2022practical},
Abc-Net \textcolor[HTML]{708090}{\tiny{[TVC2022]}} \cite{li2022area},
AcfNet \textcolor[HTML]{708090}{\tiny{[AAAI2020]}} \cite{zhang2020adaptive},
CAL-Net \textcolor[HTML]{708090}{\tiny{[ICASSP2021]}} \cite{chen2021cost},
CFNet \textcolor[HTML]{708090}{\tiny{[CVPR2021]}} \cite{shen2021cfnet},
PFSMNet \textcolor[HTML]{708090}{\tiny{[TITS2021]}} \cite{zeng2021deep},
DCVSMNet \textcolor[HTML]{708090}{\tiny{[arXiv2024]}} \cite{tahmasebi2024dcvsmnet},
DPCTF \textcolor[HTML]{708090}{\tiny{[TIP2021]}} \cite{deng2021detail},
ACVNet \textcolor[HTML]{708090}{\tiny{[CVPR2022]}} \cite{xu2022acvnet},
.
\\ 
\hline
\multirow{7}{*}{Multi-view} & \textbf{Depth-based}: Convert 2D spatial features into 3D spatial features through depth estimation. &
BEVDepth \textcolor[HTML]{708090}{\tiny{[AAAI2023]}} \cite{bevdepth},
BEVDet \textcolor[HTML]{708090}{\tiny{[arXiv2021]}} \cite{huang2021bevdet},
BEVDet4D \textcolor[HTML]{708090}{\tiny{[arXiv2022]}} \cite{huang2022bevdet4d},
LSS \textcolor[HTML]{708090}{\tiny{[ECCV2020]}} \cite{lss},
BEVHeight \textcolor[HTML]{708090}{\tiny{[CVPR2023]}} \cite{yang2023bevheight},
BEVHeight++ \textcolor[HTML]{708090}{\tiny{[arXiv2023]}} \cite{yang2023bevheight++},
BEV-SAN \textcolor[HTML]{708090}{\tiny{[CVPR2023]}} \cite{chi2023bev},
BEVUDA \textcolor[HTML]{708090}{\tiny{[arXiv2022]}} \cite{liu2022multi},
BEVPoolv2 \textcolor[HTML]{708090}{\tiny{[arXiv2022]}} \cite{huang2022bevpoolv2},
BEVStereo \textcolor[HTML]{708090}{\tiny{[AAAI2023]}} \cite{li2023bevstereo},
BEVStereo++ \textcolor[HTML]{708090}{\tiny{[arXiv2023]}} \cite{li2023bevstereo++},
TiG-BEV \textcolor[HTML]{708090}{\tiny{[arXiv2022]}} \cite{huang2022tig},
DG-BEV \textcolor[HTML]{708090}{\tiny{[CVPR2023]}} \cite{wang2023towards},
HotBEV \textcolor[HTML]{708090}{\tiny{[NeuriPS2024]}} \cite{dong2024hotbev},
BEVNeXt \textcolor[HTML]{708090}{\tiny{[CVPR2024]}} \cite{li2023bevnext}.
\\ \cline{2-3}
& \textbf{Query-based}: Influenced by the transformer technology stack, there is a trend to explicitly or implicitly query Bird's Eye View (BEV) features. &
PolarFormer \textcolor[HTML]{708090}{\tiny{[AAAI2023]}} \cite{jiang2023polarformer},
SparseBEV \textcolor[HTML]{708090}{\tiny{[ICCV2023]}} \cite{liu2023sparsebev},
BEVFormer \textcolor[HTML]{708090}{\tiny{[ECCV2022]}} \cite{bevformer},
PETR \textcolor[HTML]{708090}{\tiny{[ECCV2022]}} \cite{liu2022petr},
PETRv2 \textcolor[HTML]{708090}{\tiny{[ICCV2023]}} \cite{liu2023petrv2},
M3DETR  \textcolor[HTML]{708090}{\tiny{[WACV2022]}} \cite{M3DETR},
FrustumFormer \textcolor[HTML]{708090}{\tiny{[CVPR2023]}} \cite{wang2023frustumformer},
DETR4D \textcolor[HTML]{708090}{\tiny{[arXiv2022]}} \cite{luo2022detr4d},
Sparse4D \textcolor[HTML]{708090}{\tiny{[arXiv2022]}} \cite{lin2022sparse4d},
Sparse4D v2 \textcolor[HTML]{708090}{\tiny{[arXiv2023]}} \cite{lin2023sparse4dv2},
Sparse4D v3 \textcolor[HTML]{708090}{\tiny{[arXiv2023]}} \cite{lin2023sparse4dv3},
SOLOFusion \textcolor[HTML]{708090}{\tiny{[ICLR2022]}} \cite{park2022time},
CAPE \textcolor[HTML]{708090}{\tiny{[CVPR2023]}} \cite{xiong2023cape},
VEDet \textcolor[HTML]{708090}{\tiny{[CVPR2023]}} \cite{chen2023viewpoint},
Graph-DETR3D \textcolor[HTML]{708090}{\tiny{[ACMMM]}} \cite{chen2022graph},
3DPPE \textcolor[HTML]{708090}{\tiny{[CVPR2023]}} \cite{shu20233dppe},
BEVDistill \textcolor[HTML]{708090}{\tiny{[ICLR2023]}} \cite{chen2022bevdistill},
StreamPETR \textcolor[HTML]{708090}{\tiny{[ICCV2023]}} \cite{wang2023exploring},
Far3D \textcolor[HTML]{708090}{\tiny{[ICCV2023]}} \cite{jiang2023far3d},
CLIP-BEVFormer \textcolor[HTML]{708090}{\tiny{[CVPR2024]}} \cite{pan2024clipbevformer},
BEVFormer v2 \textcolor[HTML]{708090}{\tiny{[CVPR2023]}} \cite{bevformerv2}.
\\ 
\bottomrule
\end{tabular}}
\end{table*}

\begin{table}[t]
\scriptsize
\caption []{A comprehensive performance analysis of various categories of Camera-only 3D object detection methods across different datasets. We report the inference time (ms) originally reported in the papers, and report AP$_{3D}(\%)$ for 3D car detection on the KITTI test benchmark, mAP (\%) and NDS scores on the nuScenes test set. `R.E.P.' denotes `Representation'. `PUB' denotes `Publication'. `M.V.' denotes `Multi-view'. `L.T.' denotes `Latency Time'.}
\label{table:lidarnoise}
\renewcommand\arraystretch{0.75}
\setlength{\tabcolsep}{0.3mm}{
\begin{tabular}{ll|c|c|cc|ccc|cc }
    \toprule
  \multicolumn{2}{c|}{\multirow{2}{*}{Method}} & \multirow{2}{*}{R.E.P.}& \multirow{2}{*}{PUB} &\multirow{2}{*}{L.T.} &\multirow{2}{*}{GPU}&\multicolumn{3}{c|}{KITTI Car} &\multicolumn{2}{c}{nuScenes}  \\           

&&&&&&Easy&Mod.&Hard& mAP  & NDS   \\ 
\midrule
&FQNet\cite{liu2019deep}&\multirow{16}{*}{Mono.} &\textcolor[HTML]{708090}{\tiny{CVPR2019}}&500&1080Ti& 2.77& 1.51& 1.01 & - & - \\
&ROI-10D\cite{manhardt2019roi}& &\textcolor[HTML]{708090}{\tiny{CVPR2019}}&200&-& 4.32& 2.02& 1.46 & -& - \\
&MonoGRNet\cite{qin2019monogrnet}& &\textcolor[HTML]{708090}{\tiny{AAAI2019}}&60&TITANX& 9.61& 5.74& 4.25 & -& -\\
&MonoDIS\cite{simonelli2019disentangling}& &\textcolor[HTML]{708090}{\tiny{CVPR2019}}&100&V100& 10.37& 7.94& 6.40& 30.4& 38.4 \\
&MonoPair\cite{chen2020monopair}& &\textcolor[HTML]{708090}{\tiny{CVPR2020}}&60&1080Ti&13.04& 9.99& 8.65& -& - \\
&SMOKE\cite{liu2020smoke}& &\textcolor[HTML]{708090}{\tiny{CVPR2020}}&30&TITANX& 14.03& 9.76 & 7.84 & -& - \\
&PatchNet\cite{RPLR}& &\textcolor[HTML]{708090}{\tiny{ECCV2020}}&400&1080& 15.68& 11.12& 10.17 & -& - \\
&CaDDN\cite{CaDDN}& &\textcolor[HTML]{708090}{\tiny{CVPR2021}}&-&-& 19.17 &13.41 &11.46 & - & - \\
&FCOS3D\cite{wang2021fcos3d}& &\textcolor[HTML]{708090}{\tiny{CVPR2021}}&-&-& -& -& - &35.8& 42.8 \\
&MonoFlex\cite{zhang2021objects}& &\textcolor[HTML]{708090}{\tiny{CVPR2021}}&30&2080Ti& 19.94& 13.89& 12.07 & -&- \\
&PGD\cite{wang2022probabilistic}& &\textcolor[HTML]{708090}{\tiny{CVPR2022}}&28&1080Ti& -& -& - & 38.6& 44.8 \\
&MonoDTR\cite{MonoDTR}& &\textcolor[HTML]{708090}{\tiny{CVPR2022}}&37&V100& 21.99& 15.39 &12.73 & -& - \\
&NeurOCS\cite{min2023neurocs}& &\textcolor[HTML]{708090}{\tiny{CVPR2023}}&-&-& 29.89 &18.94 & 15.90 & -& - \\
&MonoATT\cite{zhou2023monoatt}& &\textcolor[HTML]{708090}{\tiny{CVPR2023}}&56&3090& 24.72 &17.37& 15.00 & -& - \\
&MonoDETR\cite{zhang2022monodetr}& &\textcolor[HTML]{708090}{\tiny{ICCV2023}}&38&3090& 25.00 &16.47& 13.58 & -& - \\
&MonoCD\cite{yan2024monocd}& &\textcolor[HTML]{708090}{\tiny{CVPR2024}}&36&2080Ti& 25.53 & 16.59 & 14.53 & -& - \\
\cmidrule{2-11}
&RT3DStereo\cite{RT3DOD}& &\textcolor[HTML]{708090}{\tiny{ITS2019}}&79&TITANX& 29.90 &	23.28 &	18.96 & -& - \\
&Stereo R-CNN \cite{li2019stereo}&\multirow{14}{*}{Stereo} &\textcolor[HTML]{708090}{\tiny{CVPR2019}}&420&TITANXp& 47.58& 30.23& 23.72 & - & - \\
&Pseudo-LiDAR\cite{pseudoLiDAR}& &\textcolor[HTML]{708090}{\tiny{CVPR2019}}&-&-& 54.53 & 34.05& 28.25 & -& - \\
&OC-Stereo \cite{pon2020object}& &\textcolor[HTML]{708090}{\tiny{ICRA2020}}&350&TITANXp& 55.15 &37.60& 30.25 & -& - \\
&ZoomNet \cite{xu2020zoomnet}& &\textcolor[HTML]{708090}{\tiny{AAAI2020}}&-&-& 55.98 & 38.64 & 30.97 & -& - \\
&Disp R-CNN\cite{disprcnn}& &\textcolor[HTML]{708090}{\tiny{CVPR2020}}&-&-& 58.53 & 37.91 & 31.93 & -& - \\
&DSGN\cite{chen2020dsgn}& &\textcolor[HTML]{708090}{\tiny{CVPR2020}}&682&V100& 73.50& 52.18 &45.14 & -& -\\
&CG-Stereo\cite{CGStereo3D}& &\textcolor[HTML]{708090}{\tiny{IROS2020}}&570&2080Ti& 74.39 &53.58 &46.50 & -& -\\
&YoloStereo3D\cite{liu2021yolostereo3d}& &\textcolor[HTML]{708090}{\tiny{ICRA2021}}&80&1080Ti& 65.68 &41.25 &30.42 & -& -\\
&LIGA-Stereo\cite{guo2021liga}& &\textcolor[HTML]{708090}{\tiny{ICCV2021}}&400&TITANXp& 81.39 &64.66 &57.22 & -& -\\
&PLUMENet\cite{wang2021plumenet}& &\textcolor[HTML]{708090}{\tiny{IROS2021}}&150&V100& 83.00 &66.30 &56.70  & -& -\\
&ESGN\cite{gao2022esgn}& &\textcolor[HTML]{708090}{\tiny{TCSVT2022}}&62&3090& 65.80 &46.39&38.42 & -& -\\
&SNVC\cite{li2022stereosnvc}& &\textcolor[HTML]{708090}{\tiny{AAAI2022}}&-&-& 78.54 &61.34 &54.23 & -& -\\
&DSGN++\cite{chen2022dsgn++}& &\textcolor[HTML]{708090}{\tiny{TPAMI2022}}&281&2080Ti& 83.21 &67.37 &59.91 & -& -\\
&StereoDistill\cite{liu2023stereodistill}& &\textcolor[HTML]{708090}{\tiny{AAAI2023}}&-&-& 81.66 &66.39 &57.39 & -& -\\
\cmidrule{2-11}
&BEVDet\cite{huang2021bevdet}&\multirow{23}{*}{M.V.}  &\textcolor[HTML]{708090}{\tiny{arXiv2021}}&526&3090& - &- &- & 42.2& 48.2\\ 
&DETR3D\cite{wang2022detr3d}& &\textcolor[HTML]{708090}{\tiny{PMLR2022}}&-&-& - &- &- & 41.2& 47.9\\
&Graph-DETR3D\cite{chen2022graph}& &\textcolor[HTML]{708090}{\tiny{ACMMM2022}}&-&-& - &- &- & 42.5& 49.5\\
&BEVDet4D\cite{huang2022bevdet4d}& &\textcolor[HTML]{708090}{\tiny{arXiv2022}}&526&3090& - &- &- & 42.1& 54.5\\
&PETR\cite{liu2022petr}& &\textcolor[HTML]{708090}{\tiny{ECCV2022}}&93&V100& - &- &- & 44.1& 50.4\\
&BEVFormer\cite{bevformer}& &\textcolor[HTML]{708090}{\tiny{ECCV2022}}&588&V100& - &- &- & 48.1& 56.9\\
&Sparse4D\cite{lin2022sparse4d}& &\textcolor[HTML]{708090}{\tiny{arXiv2022}}&164&3090& - &- &- & 51.1& 59.5\\
&PolarFormer\cite{jiang2023polarformer}& &\textcolor[HTML]{708090}{\tiny{AAAI2023}}&-&-& - &- &- & 49.3& 57.2\\
&BEVDistill\cite{chen2022bevdistill}& &\textcolor[HTML]{708090}{\tiny{ICLR2023}}&-&-& - &- &- & 49.8& 59.4\\
&VEDet\cite{chen2023viewpoint}& &\textcolor[HTML]{708090}{\tiny{CVPR2023}}&-&-& - &- &- & 50.5& 58.5\\
&PETRv2\cite{liu2023petrv2}& &\textcolor[HTML]{708090}{\tiny{ICCV2023}}&53&3090& - &- &- & 51.9& 60.1\\
&BEVDepth\cite{bevdepth}& &\textcolor[HTML]{708090}{\tiny{AAAI2023}}&-&-& - &- &- & 52.0& 60.9\\
&BEVStereo\cite{li2023bevstereo}& &\textcolor[HTML]{708090}{\tiny{AAAI2023}}&-&-& - &- &- & 52.5& 61.0\\
&DistillBEV\cite{wang2023distillbev}& &\textcolor[HTML]{708090}{\tiny{ICCV2023}}&-&-& - &- &- & 52.5& 61.2\\
&BEVStereo++\cite{li2023bevstereo++}& &\textcolor[HTML]{708090}{\tiny{arXiv2023}}&-&-& - &- &- & 54.6& 62.5\\
&SparseBEV\cite{liu2023sparsebev}& &\textcolor[HTML]{708090}{\tiny{ICCV2023}}&43&3090& - &- &- & 55.6& 63.6\\
&CAPE\cite{xiong2023cape}& &\textcolor[HTML]{708090}{\tiny{CVPR2023}}&-&-& - &- &- & 52.5& 61.0\\
&Sparse4Dv2\cite{lin2023sparse4dv2}& &\textcolor[HTML]{708090}{\tiny{arXiv2023}}&49&3090& - &- &- & 55.7& 63.8\\
&Sparse4Dv3\cite{lin2023sparse4dv3}& &\textcolor[HTML]{708090}{\tiny{arXiv2023}}&51&3090& - &- &- & 57.0& 65.6\\
&StreamPETR\cite{wang2023exploring}& &\textcolor[HTML]{708090}{\tiny{CVPR2023}}&32&3090& - &- &- & 62.0& 67.6\\
&Far3D\cite{jiang2023far3d}& &\textcolor[HTML]{708090}{\tiny{ICCV2023}}&-&-& - &- &- & 63.5& 68.7\\
&BEVNeXt\cite{li2023bevnext}& &\textcolor[HTML]{708090}{\tiny{CVPR2024}}&227&3090& - &- &- & 55.7& 64.2\\
&CLIP-BEVFormer\cite{pan2024clipbevformer}& &\textcolor[HTML]{708090}{\tiny{CVPR2024}}&-&-& - &- &- & 44.7& 54.7\\

\bottomrule
  \end{tabular} }
  \label{tab:camera_latency}
\end{table}

\begin{table}[t]
\scriptsize
\caption []{\textcolor{black}{A comprehensive performance analysis of various categories of LiDAR-only  3D object detection methods across different datasets. `P.V.' denotes `Point-Voxel based'. The other settings are the same as Table \ref{tab:camera_latency}.
}}
\label{table:lidarnoise}
\renewcommand\arraystretch{0.75}
\setlength{\tabcolsep}{0.53mm}{
\begin{tabular}{l|c|c|cc|ccc|cc }
    \toprule
  \multicolumn{1}{c|}{\multirow{2}{*}{Method}} & \multirow{2}{*}{R.E.P.}& \multirow{2}{*}{PUB} &\multirow{2}{*}{L.T.} &\multirow{2}{*}{GPU}&\multicolumn{3}{c|}{KITTI Car} &\multicolumn{2}{c}{nuScenes}  \\           

&&&&&Easy&Mod.&Hard& mAP  & NDS   \\ 
\midrule
PIXOR \cite{pixor}&\multirow{7}{*}{View}  &\textcolor[HTML]{708090}{\tiny{CVPR2018}}&35&TitanXp& 81.70& 77.05& 72.95 & -& - \\
HDNet \cite{hdnet}& &\textcolor[HTML]{708090}{\tiny{CoRL2018}}&-  &-     & 89.14& 86.57& 78.32 & -& - \\
BirdNet \cite{birdnet}& &\textcolor[HTML]{708090}{\tiny{ITSC2018}}&-&-& 75.52& 50.81& 50.00 & -& - \\
RCD\cite{rangecd}& &\textcolor[HTML]{708090}{\tiny{arXiv2020}}&301&V100& 85.37& 82.61& 77.80 & -& - \\
RangeRCNN\cite{rangercnn}& &\textcolor[HTML]{708090}{\tiny{arXiv2020}}&45&V100& 88.47 &81.33 &77.09 & -& - \\
RangeIoUDet\cite{rangeioudet}& &\textcolor[HTML]{708090}{\tiny{CVPR2021}}&22&V100& 88.60 &79.80 &76.76 & -& - \\
RangeDet\cite{RangeDet}& &\textcolor[HTML]{708090}{\tiny{ICCV2021}}&83&2080Ti& 85.41& 77.36& 72.60 & -& - \\
\cmidrule{1-10}

IPOD \cite{ipod}&\multirow{16}{*}{Point}  &\textcolor[HTML]{708090}{\tiny{arXiv2018}}&-&-& 71.40& 53.46& 48.34 & -& - \\
PointRGCN\cite{PointRGCN}&  &\textcolor[HTML]{708090}{\tiny{arXiv2019}}&262&1080Ti& 85.97& 75.73& 70.60 & -& - \\
StarNet \cite{StarNet} &  &\textcolor[HTML]{708090}{\tiny{arXiv2019}}&-&-& 81.63& 73.99& 67.07 & -& - \\
PointRCNN \cite{Pointrcnn}&  &\textcolor[HTML]{708090}{\tiny{CVPR2019}}&-&-& 85.94& 75.76& 68.32 & -& - \\
STD\cite{Std}& &\textcolor[HTML]{708090}{\tiny{ICCV2019}}&80&TITANv& 87.95 &79.71& 75.09 & -& - \\
PI-RCNN\cite{PI-RCNN}&  &\textcolor[HTML]{708090}{\tiny{AAAI2020}}&11&TITAN& 84.37& 74.82& 70.03 & -& - \\
Point-GNN\cite{Point-gnn}& &\textcolor[HTML]{708090}{\tiny{CVPR2020}}&643&1070& 88.33&79.47&72.29 & -& - \\
3DSSD\cite{3dssd}& &\textcolor[HTML]{708090}{\tiny{CVPR2020}}&38&TITANv& 88.36 &79.57 &74.55 & -& - \\
3D-CenterNet\cite{3D-CenterNet}& &\textcolor[HTML]{708090}{\tiny{PR2021}}&19&TITANXp& 86.83& 80.17& 75.96 & -& - \\
DGCNN\cite{DGCNN}&  &\textcolor[HTML]{708090}{\tiny{NeuriPS2021}}&-&-& - &-& - & 53.3& 63.0 \\
PC-RGNN\cite{PC-RGNN}&  &\textcolor[HTML]{708090}{\tiny{AAAI2021}}&-&-& 89.13 &79.90& 75.54 & -& - \\
Pointformer\cite{3dobwithpointformer}&  &\textcolor[HTML]{708090}{\tiny{CVPR2021}}&-&-& 87.13& 77.06& 69.25 & -& - \\
IA-SSD\cite{IA-SSD}& &\textcolor[HTML]{708090}{\tiny{CVPR2022}}&12&2080Ti&  88.34& 80.13& 75.04 & -& - \\
SASA\cite{sasa}& &\textcolor[HTML]{708090}{\tiny{AAAI2022}}&36&V100&  88.76& 82.16& 77.16 & -& - \\
SVGA-Net\cite{svganet}&  &\textcolor[HTML]{708090}{\tiny{AAAI2022}}&-&-& 87.33 &80.47& 75.91 & -& - \\
PG-RCNN\cite{PG-RCNN}& &\textcolor[HTML]{708090}{\tiny{TGRS2023}}&60&3090&  89.38 &82.13 &77.33 & -& - \\

\cmidrule{1-10}
SECOND\cite{Second}&\multirow{18}{*}{Voxel}  &\textcolor[HTML]{708090}{\tiny{Sensors2018}}&50&1080Ti& 83.13 &73.66& 66.20 & -& - \\
VoxelNet\cite{Voxelnet}& &\textcolor[HTML]{708090}{\tiny{CVPR2018}}&220&TITANX& 77.47& 65.11& 57.73 & -& - \\
PointPillars\cite{Pointpillars}& &\textcolor[HTML]{708090}{\tiny{CVPR2019}}&16&1080Ti& 79.05& 74.99& 68.30 & -& - \\
CBGS\cite{CBGS}& &\textcolor[HTML]{708090}{\tiny{arXiv2019}}&-&-&-&-&-& 52.8& 63.3 \\
PartA2\cite{part2}& &\textcolor[HTML]{708090}{\tiny{TPAMI2020}}&71&ITANXp& 85.94& 77.86& 72.00 & -& - \\
Voxel-FPN\cite{VoxelFPN}& &\textcolor[HTML]{708090}{\tiny{Sensors2020}}&20&1080Ti&  85.64& 76.70& 69.44 & -& - \\
TANet\cite{Tanet}& &\textcolor[HTML]{708090}{\tiny{AAAI2020}}&35&TITANv&  83.81& 75.38& 67.66 & -& - \\
CVC-Net\cite{cvcnet2020}& &\textcolor[HTML]{708090}{\tiny{NIPS2020}}&-&-&-&-&-& 55.8& 64.2 \\
SegVoxelNet\cite{SegVoxelNet}& &\textcolor[HTML]{708090}{\tiny{ICRA2020}}&40&1080TI&84.19& 75.81& 67.80& -& - \\
HotSpotNet\cite{chen2020object}& &\textcolor[HTML]{708090}{\tiny{ECCV2020}}&40&V100&87.60& 78.31& 73.34& 59.3& 66.0 \\
Associate-3Ddet\cite{Associate-3Ddet}& &\textcolor[HTML]{708090}{\tiny{CVPR2020}}&60&1080TI&85.99& 77.40& 70.53& -& - \\
CenterPoint\cite{Centerpoint}& &\textcolor[HTML]{708090}{\tiny{CVPR2021}}&70&TITAN&-&-&-& 58.0& 65.5 \\
CIA-SSD\cite{Cia-ssd}& &\textcolor[HTML]{708090}{\tiny{AAAI2021}}&31&ITANXp&  89.59& 80.28& 72.87 & -& - \\
SIEV-NET\cite{SIEV-Net}& &\textcolor[HTML]{708090}{\tiny{TGRS2021}}&45&1080Ti&  85.21& 76.18& 70.06 & -& - \\
VoTr-TSD\cite{voxeltransformer}& &\textcolor[HTML]{708090}{\tiny{ICCV2021}}&139&V100&  89.90& 82.09& 79.14 & -& - \\
Voxel R-CNN\cite{Voxelrcnn}& &\textcolor[HTML]{708090}{\tiny{AAAI2021}}&40&2080TI&  90.90& 81.62& 77.06 & -& - \\
PillarNet\cite{pillarnet}& &\textcolor[HTML]{708090}{\tiny{ECCV2022}}&-&-&  -& -& - & 66.0& 71.4  \\
VoxelNeXt\cite{voxelnext}& &\textcolor[HTML]{708090}{\tiny{CVPR2023}}&-&-&  -& -& - & 64.5& 70.0 \\
\cmidrule{1-10}
PV-RCNN\cite{Pv-rcnn}&\multirow{15}{*}{P.V.} &\textcolor[HTML]{708090}{\tiny{CVPR2020}}&80&1080Ti& 90.25 &81.43& 76.82 & -& - \\
SA-SSD\cite{Structureawaresingle-stage_3d_object_detection}& &\textcolor[HTML]{708090}{\tiny{CVPR2020}}&40&2080Ti& 88.75 &79.79& 74.16 & -& - \\
HVPR\cite{HVPR}& &\textcolor[HTML]{708090}{\tiny{CVPR2021}}&28&2080Ti& 86.38& 77.92& 73.04 & -& - \\
VIC-NET\cite{jiang2021vic}& &\textcolor[HTML]{708090}{\tiny{ICRA2021}}&-&-& 88.60& 81.57& 77.09 & -& - \\
PVGNet\cite{PVGNet}& &\textcolor[HTML]{708090}{\tiny{CVPR2021}}&-&-& 89.94& 81.81& 77.09 & -& - \\
CT3D\cite{CT3D}& &\textcolor[HTML]{708090}{\tiny{ICCV2021}}&-&-& 87.83& 81.77& 77.16 & -& - \\
Pyramid R-CNN\cite{pyramidrcnn}& &\textcolor[HTML]{708090}{\tiny{ICCV2021}}&-&-& 88.39& 82.08& 77.49 & -& - \\
PV-RCNN++\cite{pvrcnn++}& &\textcolor[HTML]{708090}{\tiny{IJCV2023}}&-&-& 90.14& 81.88& 77.15 & -& - \\
VP-Net\cite{vpnet}& &\textcolor[HTML]{708090}{\tiny{TGRS2023}}&59&2080Ti&  90.46&82.03&79.65 & -& - \\
SASAN\cite{zhang2023sasan}& &\textcolor[HTML]{708090}{\tiny{TNNLS2023}}&104&V100&  90.40&81.90&77.20 & -& - \\
PVT-SSD\cite{PVT-SSD}& &\textcolor[HTML]{708090}{\tiny{CVPR2023}}&49&3080TI&  90.65 &82.29& 76.85 & -& - \\
HCPVF\cite{fan2023hcpvf}& &\textcolor[HTML]{708090}{\tiny{TCSVT2023}}&70&3090&  89.34& 82.63 &77.72 & -& - \\
APVR\cite{cao2023accelerating}& &\textcolor[HTML]{708090}{\tiny{TAI2023}}&-&-&  91.45& 82.17 &78.08 & 58.6& 65.9 \\
HPV-RCNN\cite{feng2023hpvrcnn}& &\textcolor[HTML]{708090}{\tiny{TCSS2023}}&81&A100&  89.33& 80.61 &75.53 & -& - \\
\bottomrule
  \end{tabular} }
  \label{tab:lidar_latency}
   
\end{table}

\begin{table}[t]
\scriptsize
\caption []{\textcolor{black}{A comprehensive performance analysis of various categories of multi-modal 3D object detection methods across different datasets. 
`P.P.' denotes Point-Projection. `F.P.' denotes Feature-Projection. `A.P.' denotes Auto-Projection. `D.P.' denotes Decision-Projection. `Q.L.' denotes Query-Learning. `U.F.' denotes Unified-Feature. The other settings are the same as Table \ref{tab:camera_latency}.
}}
\label{table:lidarnoise}
\renewcommand\arraystretch{0.75}
\setlength{\tabcolsep}{0.35mm}{
\begin{tabular}{l|c|c|cc|ccc|cc }
    \toprule
  \multicolumn{1}{c|}{\multirow{2}{*}{Method}} & \multirow{2}{*}{R.E.P.}& \multirow{2}{*}{PUB} &\multirow{2}{*}{L.T.} &\multirow{2}{*}{GPU}&\multicolumn{3}{c|}{KITTI Car} &\multicolumn{2}{c}{nuScenes}  \\           

&&&&&Easy&Mod.&Hard& mAP  & NDS   \\ 
\midrule
MVX-Net\cite{sindagi2019mvx}& \multirow{12}{*}{P.P.}&\textcolor[HTML]{708090}{\tiny{ICRA2019}}&-&-& 85.50& 73.30& 67.40 & -& - \\
RoarNet\cite{shin2019roarnet}& &\textcolor[HTML]{708090}{\tiny{IV2019}}&-&-& 83.71& 73.04& 59.16 & -& - \\
ComplexerYOLO\cite{simon2019complexer}& &\textcolor[HTML]{708090}{\tiny{CVPRW2019}} &16&1080i& 55.63& 49.44& 44.13 & -& - \\
PointPainting\cite{Pointpainting}&  &\textcolor[HTML]{708090}{\tiny{CVPR2020}}&-&-& 82.11& 71.70& 67.08 & 46.4& 58.1 \\
EPNet\cite{Epnet}& &\textcolor[HTML]{708090}{\tiny{ECCV2020}}&-&-& 89.81& 79.28& 74.59 & -& - \\
PointAugmenting\cite{wang2021pointaugmenting}& &\textcolor[HTML]{708090}{\tiny{CVPR2021}}&542& 1080Ti& -& -& - & 66.8& 71.0 \\
FusionPainting\cite{xu2021fusionpainting}& &\textcolor[HTML]{708090}{\tiny{ITSC2021}}&-&-& -& -& - & 66.5& 70.7 \\
MVP\cite{mvp}& &\textcolor[HTML]{708090}{\tiny{NeurIPS2021}}&-&-& -& -& - & 66.4& 70.5 \\
Centerfusion\cite{nabati2021centerfusion}& &\textcolor[HTML]{708090}{\tiny{WACV2021}}&-&-& -& -& - & -& -  \\
EPNet++\cite{epnetpami}& &\textcolor[HTML]{708090}{\tiny{TPAMI2022}}&-&-& 91.37& 81.96& 76.71 & -& - \\
MSF\cite{fusionpatingtgrs}& &\textcolor[HTML]{708090}{\tiny{TGRS2024}}&63&V100& -& -& -& 68.2& 71.6 \\
PPF-Det\cite{xie2024ppfDet}& &\textcolor[HTML]{708090}{\tiny{TITS2024}}&29& TITANX  & 89.51& 84.46& 78.91 & -& - \\

\cmidrule{1-10}

Cont Fuse\cite{liang2018deep}&\multirow{8}{*}{F.P.} &\textcolor[HTML]{708090}{\tiny{ECCV2018}}&60&-& 82.54 & 66.22 & 64.04 & -& - \\
MMF\cite{mmf}& &\textcolor[HTML]{708090}{\tiny{CVPR2019}}&80&-& 86.81& 76.75& 68.41 & -& - \\
Focals Conv\cite{focalconv}& &\textcolor[HTML]{708090}{\tiny{CVPR2022}}&125&2080Ti& 90.55& 82.28& 77.59 & 67.8 &71.8 \\
VFF\cite{vff}& &\textcolor[HTML]{708090}{\tiny{CVPR2022}}&-&-& 89.50 &82.09& 79.29 & 68.4 &72.4  \\
LargeKernel3D\cite{LargeKernel3D}& &\textcolor[HTML]{708090}{\tiny{CVPR2023}}&145&2080ti& -& -& - & 71.2& 74.2 \\
SupFusion\cite{SupFusion}& &\textcolor[HTML]{708090}{\tiny{ICCV2023}}&-&-& -& -& - & 56.6& 64.6 \\
VoxelNextFusion\cite{VoxelNextFusion}& &\textcolor[HTML]{708090}{\tiny{TGRS2023}}&54&A6000& 90.90& 82.93& 80.6 & 68.8& 72.5 \\
RoboFusion\cite{song2024robofusion}& &\textcolor[HTML]{708090}{\tiny{IJCAI2024}}&322&A100& 91.75& 84.08 &80.71 &69.9 & 72.0 \\
\cmidrule{1-10}

PI-RCNN\cite{PI-RCNN}&\multirow{8}{*}{A.P.} &\textcolor[HTML]{708090}{\tiny{AAAI2020}}& 90&TITAN& 84.37& 74.82& 70.03 &-& - \\
3D-CVF\cite{3dcvf}& &\textcolor[HTML]{708090}{\tiny{ECCV2020}}&75&1080Ti&89.20 &80.05& 73.11& -& - \\
3D Dual-Fusion\cite{3DDualFusion}& &\textcolor[HTML]{708090}{\tiny{Arxiv2022}}&-&-& 91.01& 82.40& 79.39 &  70.6& 73.1\\
AutoAlignV2\cite{autoalignv2}& &\textcolor[HTML]{708090}{\tiny{ECCV2022}}& 208&V100& -&-&- & 68.4& 72.4 \\
HMFI\cite{HMFI}& &\textcolor[HTML]{708090}{\tiny{ECCV2022}}&-&-& 88.90& 81.93& 77.30 & -& - \\
LoGoNet\cite{Logonet}& &\textcolor[HTML]{708090}{\tiny{ICCV2023}}&-&-& 91.80& 85.06& 80.74 & -& - \\
GraphAlign\cite{graphalign}& &\textcolor[HTML]{708090}{\tiny{ICCV2023}}&26&A6000& 90.96& 83.49& 80.14 & 66.5& 70.6 \\
GraphAlign++\cite{graphalign++}& &\textcolor[HTML]{708090}{\tiny{TCSVT2024}}&149&V100& 90.98& 83.76& 80.16 & 68.5&72.2 \\
\cmidrule{1-10}
CLOCs\cite{CLOCs}&\multirow{8}{*}{D.P.} &\textcolor[HTML]{708090}{\tiny{IROS2020}}&-&-& 83.68& 68.78& 61.67 & -&- \\
AVOD\cite{avod}& &\textcolor[HTML]{708090}{\tiny{IROS2018}}&100&TITANXp& 81.94& 71.88 &66.38 & -&- \\
MV3D\cite{mv3d}& &\textcolor[HTML]{708090}{\tiny{CVPR2017}}&240& TitanX & 71.09& 62.35& 55.12 & -&- \\
F-PointNets\cite{Frustumpointnets}& &\textcolor[HTML]{708090}{\tiny{CVPR2018}}&-&-& 81.20& 70.39& 62.19& -&- \\
F-ConvNet\cite{Frustumconvnet}& &\textcolor[HTML]{708090}{\tiny{IROS2019}}&-&-& 82.11& 71.70& 67.08& 46.4& 58.1 \\
F-PointPillars\cite{Frustum-pointpillars}& &\textcolor[HTML]{708090}{\tiny{ICCVW2021}}&-&-& 88.90& 79.28& 78.07  & -&- \\
Fast-CLOCs\cite{Fast-CLOCs}& &\textcolor[HTML]{708090}{\tiny{WACV2022}}&-&-& 89.11& 80.34& 76.98  & -&-  \\
Graph R-CNN\cite{Graph-RCNN}& &\textcolor[HTML]{708090}{\tiny{ECCV22022}}&13&1080Ti& 91.89& 83.27& 77.78 & -&- \\
\cmidrule{1-10}
TransFusion\cite{Transfusion}&\multirow{6}{*}{Q.L.} &\textcolor[HTML]{708090}{\tiny{CVPR2022}}&265&V100& -& -& - & 68.9& 71.7 \\
DeepInteraction\cite{DeepInteraction}& &\textcolor[HTML]{708090}{\tiny{NeuriPS2022}}&204&A100& -& -& - & 70.8& 73.4 \\
SparseFusion\cite{sparsefusion}& &\textcolor[HTML]{708090}{\tiny{ICCV2023}}&188&A6000& -& -& - & 72.0&73.8\\
AutoAlign\cite{autoalign}& &\textcolor[HTML]{708090}{\tiny{IJCAI2022}}&-&-& -& -& -&  65.8&70.9 \\
SparseLIF\cite{zhang2024sparselif}& &\textcolor[HTML]{708090}{\tiny{arXiv2024}}&340&A100& -& -& -& 75.9& 77.7\\
FusionFormer\cite{hu2023fusionformer}& &\textcolor[HTML]{708090}{\tiny{arXiv2024}}&263&A100& -& -& -& 71.4& 74.1\\
FSF\cite{FSF}& &\textcolor[HTML]{708090}{\tiny{TPAMI2024}}&141&3090& -& -& -& 70.6 &74.0\\
\cmidrule{1-10}
BEVFusion-PKU\cite{BEVFusion}&\multirow{18}{*}{U.F.} &\textcolor[HTML]{708090}{\tiny{NeuriPS2022}}&-&-& -& -& - & 69.2& 71.8\\
BEVFusion-MIT\cite{bevfusion-mit}& &\textcolor[HTML]{708090}{\tiny{ICRA2023}}&119&3090& -& -& - & 70.2 &72.9\\
EA-BEV\cite{EA-BEV}& &\textcolor[HTML]{708090}{\tiny{arXiv2023}}&195&V100& -& -& - & 71.2 &73.1 \\
BEVFusion4D\cite{cai2023bevfusion4d}& &\textcolor[HTML]{708090}{\tiny{arXiv2023}}&500&V100& -& -& - & 72.0 &73.5 \\
FocalFormer3D\cite{FocalFormer3D}& &\textcolor[HTML]{708090}{\tiny{ICCV2023}}&109&V100& -& -& - & 71.6 &73.9 \\
FUTR3D\cite{FUTR3D}& &\textcolor[HTML]{708090}{\tiny{CVPR2023}}&-&-& -& -& - & 69.4& 72.1 \\
UniTR\cite{UniTR}& &\textcolor[HTML]{708090}{\tiny{ICCV2023}}&107&A100& -& -& - & 70.9 & 74.5\\
VirConv\cite{virconv}& &\textcolor[HTML]{708090}{\tiny{CVPR2023}}&92&V100&92.48& 87.20& 82.45& 68.7& 72.3\\
MSMDFusion\cite{MSMDFusion}& &\textcolor[HTML]{708090}{\tiny{CVPR2023}}&265&V100& -& -& - & 71.5 &74.0 \\
SFD\cite{sfd}& &\textcolor[HTML]{708090}{\tiny{CVPR2022}}&10&2080Ti&91.73& 84.76& 77.92 & -& - \\
CMT\cite{cmt}& &\textcolor[HTML]{708090}{\tiny{ICCV2023}}&167&A100& -& -& - & 72.0& 74.1 \\
UVTR\cite{UVTR}& &\textcolor[HTML]{708090}{\tiny{NeuriPS2022}}&-&-& -& -& - & 67.1& 71.1 \\
ObjectFusion\cite{ObjectFusion}& &\textcolor[HTML]{708090}{\tiny{ICCV2023}}&274&V100& -& -& - & 71.0 &73.3 \\
GraphBEV\cite{song2024graphbev}& &\textcolor[HTML]{708090}{\tiny{arXiv2024}}&141&A100& -& -& - & 71.7 &73.6 \\
ContrastAlign\cite{song2024contrastalign}& &\textcolor[HTML]{708090}{\tiny{arXiv2024}}&154&A100& -& -& - & 71.8& 73.8 \\
IS-Fusion\cite{yin2024isfusion}& &\textcolor[HTML]{708090}{\tiny{CVPR2024}}&-&-& -& -& - &  73.0 &75.2 \\

\bottomrule
  \end{tabular} }
  \label{tab:multimodel_latency}
\end{table}

\subsection{Monocular 3D object detection}

Monocular 3D object detection refers to performing 3D object detection using only one camera, which aims to infer the 3D positions, sizes, and orientations of objects from a single image \cite{peng2022survey}. 
In recent years, monocular 3D object detection has gained increasing attention due to its advantages of low cost, low power consumption, and ease of deployment in real-world applications.
However, monocular methods face many challenges, owing to the insufficient 3D information in monocular pictures, such as accurately localizing \textcolor{black}{3D positions}, handling occluded scenes, \textcolor{black}{and so on.}
Overcoming these challenges relies on leveraging depth information to supplement the missing 3D information in monocular images. Typically, most approaches employ depth estimation tasks to acquire depth information from images. However, monocular depth estimation is an ill-posed and highly challenging task, prompting researchers to dedicate significant efforts to \textcolor{black}{optimizing} the accuracy and stability of depth estimation.
%

\subsubsection{\textbf{Prior-guided monocular 3D object detection}} 
In recent years, prior-guided monocular methods \cite{zeeshan2014cars,chabot2017deep,he2019mono3dplus,kundu20183d,manhardt2019roi,beker2020monocular,zakharov2020autolabeling,ku2019monocular,xiang2015data,mousavian20173d,brazil2019m3d,naiden2019shift,li2020rtm3d,shi2020distance,cai2020monocular,lu2021geometry,zhang2021objects,shi2021geometry,chen2020monopair,liu2019deep} have continuously explored how to utilize the hidden prior knowledge of object shapes and scene geometry in images to address the challenges of monocular 3D object detection. \textcolor{black}{The effective integration of this prior knowledge is crucial for mitigating the uncertainty and ill-posed nature of inherent in monocular 3D object detection problems.} By introducing pre-trained subnetworks or auxiliary tasks, prior knowledge can provide additional information or constraints to assist in the accurate localization of 3D objects and enhance detection precision and robustness.

Widely adopted prior knowledge in 3D objects includes object shapes~\cite{chen2021monorun,Ku_Pon_Waslander_2019,DeepSDF,beker2020monocular,zakharov2020autolabeling,he2019mono3dplus}, geometric consistency~\cite{MonocularLoss,naiden2019shift,shi2020distance,li2020rtm3d,shi2020distance,cai2020monocular,zhang2021objects}, temporal constraints~\cite{hu2019joint,brazil2020kinematic}, and segmentation information~\cite{beker2020monocular}. 
Object shape provides insights into the appearance and structure of an object, aiding in more accurate inference of the spatial position and pose of the object. 
Geometric consistency knowledge assists the model in better understanding the relative positional relationships between objects in the scene, thereby improving detection consistency and robustness. 
Temporal constraints consider the continuity and stability of an object across different frames, providing vital clues for object detection. 
Additionally, leveraging segmentation information enables the model to better comprehend semantic information in images, facilitating precise localization and identification of objects. 
As a result, current works are dedicated to further exploring and utilizing prior knowledge to enhance the performance and robustness of monocular 3D object detection by integrating prior knowledge with deep learning approaches, thus driving continuous development and innovation in this field.

\subsubsection{\textbf{Camera-only monocular 3D object detection}} 
Camera-only monocular 3D object detection \cite{liu2020smoke,brazil2019m3d,brazil2020kinematic,liu2019deep,mousavian20173d,wang2021fcos3d,wang2022probabilistic,li2020rtm3d,zhang2021objects,CaDDN,simonelli2020towards,simonelli2019disentangling,li2019gs3d,qin2019monogrnet,shi2021geometry,lu2021geometry} is a kind of methods that utilize images captured by a single camera to detect and localize 3D objects. Camera-only monocular methods employ convolutional neural networks (CNNs) to directly regress 3D bounding box parameters from images, enabling the estimation of the spatial dimensions and poses of objects in three dimensions. \textcolor{black}{Inspired by} 2D detection networks, this direct regression methods can be trained end-to-end, facilitating holistic learning and inference for 3D objects. The unique challenge of monocular 3D object detection lies in inferring objects' 3D positions, dimensions, and orientations solely from a single image without relying on additional depth maps or point cloud data. Consequently, the direct regression approaches demonstrate practicality and broad applicability. By learning features from images, CNNs can predict the 3D information of  objects. The network gradually optimizes its parameters through end-to-end training to enhance the accurate extraction of 3D information. These direct regression methods streamline the entire detection process and reduce the reliance on supplementary information, improving the algorithms' robustness and generalization capability. Nevertheless, monocular 3D object detection still presents challenges, such as occlusion, viewpoint variations, and lighting conditions, which may \textcolor{black}{affect} the accuracy of 3D detection. The representative work Smoke\cite{liu2020smoke} abandons the regression of 2D bounding boxes and predicts the 3D box for each detected object by combining the estimation of individual key points with the regression of 3D variables.

\subsubsection{\textbf{Depth-assisted monocular 3D object detection}} 
Depth estimation plays a crucial role in depth-assisted monocular 3D object detection. To achieve more accurate monocular detection results, numerous studies \cite{RPLR,park2021pseudo,chang2019deep,ma2019accurate} leverage pre-trained auxiliary depth estimation networks. Specifically, the process begins by transforming monocular images into depth images using pre-trained depth estimators, such as MonoDepth\cite{godard2017unsupervised}. Subsequently, two primary methodologies are employed to handle depth images and monocular images.
Remarkable progress has been made in Pseudo-LiDAR detectors that use a pre-trained depth estimation network to generate Pseudo-LiDAR representations\cite{Neighbor-Vote,RPLR}. However, there is a \textcolor{black}{significant} performance gap between Pseudo-LiDAR and LiDAR-only detectors \textcolor{black}{due to} the errors in image-to-LiDAR generation. 
Thus,
Hong \textit{et al.}\cite{hong2022cross} attempted to transfer deeper structural information from point clouds to assist monocular image detection. 
By leveraging the mean-teacher framework, they aligned the outputs of the LiDAR-only teacher model and the Camera-only student model at both the feature-level and the response-level, aiming to achieve cross-modal knowledge transfer. 
Such depth-assisted monocular 3D object detection, by effectively integrating depth information, not only enhances detection accuracy but also extends the applicability of monocular vision to tasks involving 3D scene understanding. 

\subsection{Stereo-based 3D object detection}
Stereo-based 3D object detection is designed to identify and localize 3D objects using a pair of stereo images. Leveraging the inherent capability of stereo cameras to capture dual perspectives, stereo-based methods excel in acquiring highly accurate depth information through stereo matching and calibration. This is a feature that \textcolor{black}{distinguishes them} from monocular camera setups. Despite these advantages, stereo-based methods still face a considerable performance gap when compared to LiDAR-only counterparts. Furthermore, the realm of 3D object detection from stereo images remains relatively underexplored, with only limited research \textcolor{black}{efforts} dedicated to this domain. Specifically, these approaches involve the utilization of image pairs captured from distinct viewpoints to estimate the 3D spatial depth of each object.

\subsubsection{\textbf{2D-detection-based methods}}
Traditional 2D object detection frameworks can be modified to address stereo detection problems. Stereo R-CNNs \cite{li2019stereo} \textcolor{black}{employ} an image-based 2D detector to predict 2D proposals, generating left and right regions of interest (RoIs) for the corresponding left and right images. Subsequently, in the second stage, \textcolor{black}{they} directly \textcolor{black}{estimate} the parameters of 3D objects based on the previously generated RoIs. This paradigm has been widely adopted by \textcolor{black}{subsequent} works~\cite{disprcnn,qin2019triangulation,xu2020zoomnet,chen2021shape,peng2020ida,liu2021yolostereo3d,peng2022side}.


\subsubsection{\textbf{Pseudo-LiDAR-only methods}}

The disparity map predicted from stereo images can be transformed into a depth map and further converted into pseudo-LiDAR points. Consequently, similar to monocular detection methods, pseudo-LiDAR representations can also be employed in stereo-based 3D object detection approaches. These methods aim to enhance disparity estimation in stereo matching to achieve more accurate depth predictions. Regarding the contribution of depth in 3D detection, Wang et al. \cite{pseudoLiDAR} are pioneers in introducing the Pseudo-LiDAR representation. This representation is generated by using an image with a depth map, requiring the model to perform a depth estimation task to assist in detection.
Subsequent works have followed this paradigm and made optimizations by introducing additional color information to augment pseudo point cloud~\cite{ma2019accurate}, auxiliary tasks (instance segmentation~\cite{weng2019monocular}, foreground and background segmentation~\cite{wang2020task} and domain adaptation~\cite{ye2020monocular}) and coordinate transform scheme~\cite{wang2021progressive,RPLR}. 
\textcolor{black}{
To achieve both high accuracy and high responsiveness, Meng et al. \cite{meng2023efficient} propose a lightweight Pseudo-LiDAR 3D detection system.}
These studies indicate that the power of the pseudo LiDAR representation stems from the coordinate transformation rather than the point cloud representation itself. 

\subsubsection{\textbf{\textcolor{black}{Volume-based methods}}}
\textcolor{black}{
The general procedure of volume-based methods is to generate a cost volume from the left and the right images to represent disparity information, which is then utilized in the subsequent detection process. Volume-based methods bypass the pseudo-LiDAR representation and perform 3D object detection directly on 3D stereo volumes. These methods have inherited the traditional matching idea, but most computations now rely on 3D convolutional networks, \textcolor{black}{such as those found in references} \cite{kendall2017end,guo2019groupgwc,cheng2020hierarchical,liu2022local,xu2022acvnet,chen2020dsgn,guo2021liga,gao2022esgn}. For example, the pioneering work GC-Net\cite{kendall2017end} uses an end-to-end neural network for stereo matching, obviating the need for any post-processing steps, and \textcolor{black}{regressively computes} disparity from a cost volume constructed by a pair of stereo features. GwcNet\cite{guo2019groupgwc} employs a proposed group-wise correlation method to construct the cost volume. LEAStereo\cite{cheng2020hierarchical} utilizes NAS technology to select the optimal structure for the 3D cost volume. GANet\cite{liu2022local} designs a semi-global aggregation layer and a local guidance aggregation layer to further improve accuracy. ACVNet\cite{xu2022acvnet} introduces an attention concatenation unit to generate more accurate similarity metrics. DSGN \cite{chen2020dsgn} proposes a 3D geometric volume derived from stereo matching networks and applies a grid-based 3D detector on the volume for 3D object detection. LIGA-Stereo\cite{guo2021liga} uses a LiDAR-based detector as a teacher model to guide geometry-aware feature learning. ESGN\cite{gao2022esgn} achieves efficient stereo matching through the efficient geometry-aware feature generation (EGFG) module. Due to the benefits of large-scale training data and end-to-end training, deep learning-based stereo methods have achieved outstanding results\cite{shi2024rethinking}.
}
\subsection{Multi-view 3D object detection}

Recently, multi-view 3D object detection has demonstrated superior accuracy and robustness compared to monocular and stereo 3D object detection approaches. In contrast to LiDAR-only 3D object detection, the latest panoramic Bird's Eye View (BEV) approaches eliminate the need for high-precision maps, elevating the detection from 2D to 3D. This advancement has led to significant developments in multi-view 3D object detection. In comparison to previous reviews \cite{wang2023multi,mao20233d,wang2023multi_ijcv,wang2023multi_TITS,peng2022survey}, there has been extensive research on effectively leveraging multi-view images for 3D object detection. A key challenge in multi-camera 3D object detection is recognizing the same object across different images and aggregating object features from multiple view inputs. The current approach, a common practice, involves uniformly mapping multi-view to the Bird's Eye View (BEV) space. 
Therefore, multi-view 3D object detection, also called BEV-camera-only 3D object detection, revolves around the core challenge of unifying 2D views into the BEV space. 
Based on different spatial transformations, this can be categorized into two main methods. Ones are depth-based methods \cite{bevdepth,huang2021bevdet,lss,huang2022bevdet4d,yang2023bevheight++,yang2023bevheight,chi2023bev,wang2023frustumformer,liu2022multi,huang2022tig,cao2019multi,xu2023nerf,wang2023towards}, represented by the LSS \cite{lss}, also known as 2D to 3D transformation. The others are \textcolor{black}{query-based methods }\cite{jiang2023polarformer,liu2023sparsebev,bevformer,liu2022petr,liu2023petrv2,M3DETR,wang2023frustumformer,luo2022detr4d,wang2021multi,lin2022sparse4d,lin2023sparse4dv2,lin2023sparse4dv3,park2022time,xiong2023cape,chen2023viewpoint,chen2022graph,shu20233dppe,chen2022bevdistill,wang2023exploring,jiang2023far3d}, represented by DETR3D \cite{wang2022detr3d}, \textcolor{black}{making} a query from 3D to 2D.

\subsubsection{\textbf{Depth-based Multi-view methods}}
The direct transformation from 2D to BEV space poses a significant challenge. LSS \cite{lss} was the first to propose a depth-based method, utilizing 3D space as an intermediary. This approach involves initially predicting the grid depth distribution of 2D features and then elevating these features to voxel space. This method holds promise for achieving the transformation from 2D to BEV space more effectively. Following LSS \cite{lss}, CaDDN \cite{CaDDN} adopted a similar depth representation approach. It employed a network structure akin to LSS, primarily for predicting categorical depth distribution. By compressing voxel-space features \textcolor{black}{into BEV space}, it performed the final 3D detection. It is worth noting that CaDDN is not part of multi-view 3D object detection but rather single-view 3D object detection, which has influenced subsequent research on depth. The main distinction between LSS \cite{lss} and CaDDN \cite{CaDDN} lies in CaDDN's use of actual ground truth depth values to supervise its prediction of categorical depth distribution, resulting \textcolor{black}{in a superior depth} network capable of more accurately extracting 3D information from 2D space. This line of research has sparked a series of subsequent studies, such as BEVDet \cite{huang2021bevdet}, its temporal version BEVDet4D \cite{huang2022bevdet4d}, and BEVDepth \cite{bevdepth}. These studies are significant in advancing the transformation from 2D to 3D space and enabling more accurate object detection in the BEV space, providing valuable insights and directions for the field's development. Furthermore, some studies have addressed the issue of insufficient depth solely by encoding height information. These studies have found that with increasing distance, the depth disparity between the car and the ground rapidly diminishes \cite{yang2023bevheight,yang2023bevheight++}.


\subsubsection{\textbf{Query-based Multi-view methods}}
Under the influence of Transformer \textcolor{black}{technology, such as in the works} \cite{transformer,Swintransformer,vit,deformabledetr}, query-based Multi-view methods retrieve 2D spatial features from 3D space. Inspired by Tesla's perception system, DETR3D\cite{wang2022detr3d} introduces 3D object queries to address the aggregation of multi-view features. It \textcolor{black}{achieves} this by extracting image features from different perspectives and projecting them into 2D space using learned 3D reference points, thus obtaining image features in the Bird's Eye View (BEV) space. Query-based Multi-view methods, \textcolor{black}{as opposed} to Depth-based Multi-view methods, acquire sparse BEV features by employing a reverse querying technique, fundamentally impacting subsequent query-based developments \cite{jiang2023polarformer,liu2023sparsebev,bevformer,liu2022petr,liu2023petrv2,M3DETR,wang2023frustumformer,luo2022detr4d,wang2021multi,lin2022sparse4d,lin2023sparse4dv2,lin2023sparse4dv3,park2022time,xiong2023cape,chen2023viewpoint,chen2022graph,shu20233dppe,chen2022bevdistill,wang2023exploring,jiang2023far3d}. However, due to the potential inaccuracies associated with explicit 3D reference points, PETR \cite{liu2022petr}, influenced by DETR \cite{detr} and DETR3D \cite{wang2022detr3d}, adopts an implicit positional encoding method for constructing the BEV space, influencing subsequent works\cite{liu2023petrv2,wang2023exploring}.


\subsection{Analysis: Accuracy, Latency, Robustness}
\label{Sec:Analysis:Image}
Currently, the 3D object detection solutions based on Bird's Eye View (BEV) perception are rapidly advancing. Despite the existence of numerous reviews\cite{wang2023multi,mao20233d,wang2023multi_ijcv,wang2023multi_TITS,peng2022survey}, a comprehensive review of this field remains inadequate. It is noteworthy that Shanghai AI Lab and SenseTime Research have provided a thorough review \cite{bevsurvey} of the technical roadmap for BEV solutions. However, unlike existing reviews \cite{wang2023multi,mao20233d,wang2023multi_ijcv,wang2023multi_TITS,peng2022survey}, which primarily focus on the technical roadmap and the current state of the art, we consider crucial aspects such as autonomous driving safety perception. Following an analysis of the technical roadmap and the current state of development for Camera-only solutions, we intend to base our discussion on the foundational principles of `Accuracy, Latency, and Robustness'. We will integrate the perspectives of safety perception to guide the practical implementation of safety perception in autonomous driving.

\begin{figure*}[htbp]
\centering
\includegraphics[width=0.968\linewidth]{image/zhu.pdf}
\caption{
(a) The $AP_{\text{3D}}$ comparison of monocular-based methods \cite{liu2020smoke,liu2022learning,qin2019monogrnet,chen2020monopair,monodistill,MonoDTR,zhang2022monodetr,lian2022monojsg,yang2023mix,shi2021geometry}  and stereo-based methods \cite{pon2020object,xu2020zoomnet,li2021rts3d,liu2021yolostereo3d,gao2022esgn,disprcnn,chen2022dmf,chen2022dsgn++,chen2020dsgn,garg2020cdn}  on KITTI test dataset. 
(b) The mAP (left) and NDS (right) comparison of monocular-based methods \cite{Centernet,simonelli2019disentangling,wang2022probabilistic,park2021pseudo,wang2021fcos3d}  and Multi-view methods \cite{wang2022detr3d,liu2022petr,liu2023petrv2,bevformer,bevformerv2,park2022time,bevdepth,lin2023sparse4dv3,wang2023exploring,jiang2023far3d}  on the nuScenes test dataset.
(c) The $AP_{\text{3D}}$ comparison of View-based methods \cite{rangecd,RangeDet,rangeioudet,rangercnn}, Voxel-based methods \cite{Second,Pointpillars,part2,Voxelrcnn,PDV,voxeltransformer,PG-RCNN,TED}, Point-based \cite{PI-RCNN,Pointrcnn,3DIoU-Net,Point-gnn,3dssd,IA-SSD,sasa}, and Point-Voxel-based methods \cite{Std,lidarrcnn,HVPR,Pv-rcnn,pyramidrcnn}  on KITTI test dataset.
(d) The mAP (left) and NDS (right) comparison of Voxel-based methods~\cite{Pointpillars,Centernet,Centerpoint,focalconv,UVTR,pillarnet,voxelnext,Transfusion,FocalFormer3D} and Point-based methods~\cite{3dssd} on the nuScenes test dataset.
(e) The $AP_{\text{3D}}$ comparison of Point-Projection-based (P.P.) methods \cite{Pointpainting,sindagi2019mvx, Epnet,epnetpami}, Feature-Projection-based (F.P.) methods \cite{mmf,focalconv,SupFusion}, Auto-Projection-based (A.P.) methods \cite{PI-RCNN,3dcvf,HMFI,3DDualFusion,Robust-FusionNet,graphalign,Logonet}, Decision-Projection-based (D.P.) methods \cite{mv3d,avod,Frustum-pointpillars,Frustumconvnet,roifusion,Fast-CLOCs,CLOCs}, and Query-Learning-based (Q.L.) methods \cite{CAT-Det} on KITTI test dataset. 
(f) The mAP (left) and NDS (right) comparison of Point-Projection-based (P.P.) methods \cite{wang2021pointaugmenting}, Feature-Projection-based (F.P.) methods \cite{LargeKernel3D}, Auto-Projection-based (A.P.) methods \cite{autoalignv2,graphalign}, Query-Learning-based (Q.L.) methods \cite{autoalign,Transfusion,DeepInteraction} and Unified-Feature-based (U.F.) methods \cite{UVTR,FUTR3D,sparsefusion,BEVFusion,MSMDFusion,cmt,UniTR,cai2023bevfusion4d,FocalFormer3D,EA-BEV} on the nuScenes test dataset.
}
\label{fig:zhu_all}
\end{figure*}

\subsubsection{\textbf{Accuracy}} 
Accuracy is a focal point of interest in most research articles and reviews and is indeed of paramount importance. While accuracy can be reflected through AP (average precision), considering AP alone for comparison may not provide a comprehensive view, as different methodologies may exhibit substantial differences due to differing paradigms.

As shown in Fig. \ref{fig:zhu_all} (a), we selected ten representative methods (including classic and latest research) for comparison, and it is evident that there are significant metric disparities between monocular 3D object detection \cite{liu2020smoke,qin2019monogrnet,chen2020monopair,MonoDTR,zhang2022monodetr,lian2022monojsg,yang2023mix,shi2021geometry} and stereo-based 3D object detection \cite{pon2020object,xu2020zoomnet,li2021rts3d,liu2021yolostereo3d,gao2022esgn,chen2022dmf,chen2022dsgn++,chen2020dsgn,garg2020cdn}. The current scenario indicates that the accuracy of monocular 3D object detection is far lower than that of stereo-based 3D object detection. Stereo-based 3D object detection leverages the capture of images from two different perspectives of the same scene to obtain depth information. The greater the baseline between cameras, the wider the range of depth information captured. As shown in Fig. \ref{fig:zhu_all} (b), there were monocular 3D object detection methods \cite{Centernet,simonelli2019disentangling,wang2022probabilistic,park2021pseudo,wang2021fcos3d} on the \textcolor{black}{nuScenes dataset \cite{nuscenes}}, but no related research on stereo-based 3D object detection.

Starting from 2021, monocular methods have gradually been supplanted by multi-view (bird's-eye-view perception) 3D object detection methods \cite{wang2022detr3d,liu2022petr,liu2023petrv2,bevformer,bevformerv2,park2022time,bevdepth,lin2023sparse4dv3,wang2023exploring,jiang2023far3d}, leading to a significant improvement in mAP. The emergence of the novel bird's-eye-view paradigm and the increase in sensor quantity have substantially impacted mAP. It can be observed that initially, the disparity between DD3D \cite{park2021pseudo} and DETR3D \cite{wang2022detr3d} is not prominent, but with the continuous enhancement of multi-view 3D object detection, particularly with the advent of novel works such as Far3D \cite{jiang2023far3d}, the gap has widened. In other words, camera-only 3D object detection methods on multi-camera datasets like nuScenes \cite{nuscenes} are predominantly based on bird's-eye-view perception. If we consider accuracy solely from this single dimension, the increase in sensor quantity has significantly improved accuracy metrics (including mAP, NDS, AP, etc.).
\subsubsection{\textbf{Latency}}
In 3D object detection, latency (frames er second, FPS) and accuracy are critical metrics for evaluating algorithm performance \cite{smith2019super}. As shown in Table \ref{tab:camera_latency}, Monocular-based 3D object detection, \textcolor{black}{which relies} on data from a single camera, typically achieves higher FPS due to lower computational requirements. However, its accuracy is often inferior to stereo or multi-view systems due to the absence of depth information. Stereo-based detection, leveraging disparity information from dual cameras, enhances depth estimation accuracy but introduces greater computational complexity, potentially reducing FPS. Multi-view detection provides richer scene information and improved accuracy but demands extensive data processing, computational power, and algorithmic optimization for reasonable FPS levels. Notably, the nuScenes dataset lacks representation of stereo-based methods, with the monocular method FCOS3D\cite{wang2021fcos3d} standing out as emblematic, introduced in 2021. Over time, multi-view 3D object detection has rapidly evolved \textcolor{black}{in terms of accuracy and latency.} \textcolor{black}{In practice, real-time performance is also an important consideration when deploying a robust 3D object detection system. For example, ER3D\cite{meng2023er3d} takes stereo images as input and predicts 3D bounding boxes, which leverages a fast but inaccurate method of semi-global matching for depth estimation. Li et al. \cite{li2022real} propose a lightweight Pseudo-LiDAR 3D detection system that achieves high accuracy and responsiveness. RTS3D \cite{li2021rts3d} proposes a novel framework for faster and more accurate 3D object detection using stereo images. FastFusion \cite{meng2023fastfusion}, a three-stage stereo-LiDAR deep fusion scheme, integrates LiDAR priors into each step of the classical stereo-matching taxonomy, thereby gaining high-precision dense depth sensing in \textcolor{black}{ real-time.}
}
In conclusion, achieving safe autonomous driving necessitates balancing latency and accuracy in 3D object detection algorithms. While monocular detection is faster, it lacks precision. Stereo and multi-view methods are accurate but slower. Future research should focus on maintaining high precision while emphasizing increased FPS and reduced latency to meet the dual requirements of real-time responsiveness and safety in autonomous driving.

\subsubsection{\textbf{Robustness}}
\label{sec:camera_Robustness}
\begin{figure}[htbp]
\centering
\includegraphics[width=\linewidth]{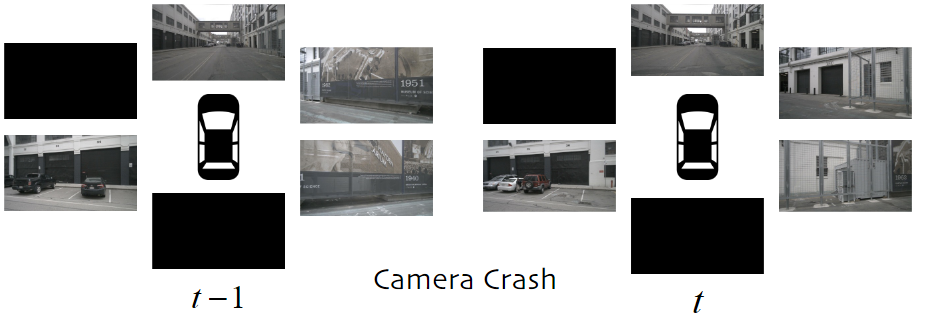}
\caption{Corruption examples in the RoboBEV\cite{RoboBEV} benchmark: simulating camera malfunction.}
\label{fig:RoboBEV_sensor_miss}
\end{figure}

Robustness constitutes a pivotal factor in the safety perception of autonomous driving, representing a topic of significant attention \textcolor{black}{that has been} previously overlooked in comprehensive reviews. In the current meticulously designed clean datasets and benchmarks, such as KITTI \cite{kitti}, nuScenes \cite{nuscenes}, and Waymo \cite{Waymo}, this aspect is not commonly addressed. Presently, research works\cite{RoboBEV,BR3D,zhu2023understanding,BRLCF,kong2023robo3d,zhang2023comprehensive,Tanet} like RoboBEV\cite{RoboBEV}, Robo3D\cite{kong2023robo3d} on 3D object detection \textcolor{black}{incorporate} considerations of robustness, exemplified by factors such as sensor misses, as illustrated in Fig. \ref{fig:RoboBEV_sensor_miss}. They have adopted a methodology involving the introduction of disturbances into datasets relevant to 3D object detection to assess robustness. This includes introducing various types of noise, such as variations in weather conditions, sensor malfunctions, motion disturbances, and object-related perturbations, aimed at unraveling the distinct impacts of different noise sources on the model. Typically, most papers investigating robustness conduct evaluations by introducing noise to the validation sets of clean datasets, such as KITTI \cite{kitti}, nuScenes \cite{nuscenes}, and Waymo \cite{Waymo}. Additionally, we highlight findings from Ref. \cite{BR3D}, where KITTI-C\cite{BR3D} and nuScenes-C\cite{BR3D} are emphasized as examples to illustrate the results of Camera-Only 3D object detection methods. Tables \ref{tab:kittic-car-moderate} and \ref{tab_nuscenes_C} provide an overall comparison, revealing that, in general, Camera-Only methods are less robust compared to LiDAR-Only and multi-modal fusion methods. They are highly susceptible to various types of noise. In KITTI-C, three representative works—SMOKE\cite{liu2020smoke}, PGD\cite{wang2022probabilistic}, and ImVoxelNet\cite{rukhovich2022imvoxelnet}—show consistently lower overall performance and reduced robustness to noise. In nuScenes-C, noteworthy methods such as DETR3D\cite{wang2022detr3d} and BEVFormer\cite{bevformer} exhibit greater robustness compared to FCOS3D \cite{wang2021fcos3d} and PGD\cite{wang2022probabilistic}, suggesting that as the number of sensors increases, overall robustness improves. In conclusion, future Camera-Only methods need to consider not only cost and accuracy metrics (mAP, NDS, etc.) but also factors related to safety perception and robustness. Our analysis aims to provide valuable insights for the safety of future autonomous driving systems.

\begin{table*}[!ht]
\scriptsize
\caption{Comparison with SOTA methods on \textbf{KITTI-C validation} set. The results are evaluated based on the \textbf{car} class with AP of $R_{\text{40}}$ at \textbf{moderate} difficulty. `RCE' denotes Relative Corruption Error from Ref.\cite{BR3D}.
}
\label{tab:kittic-car-moderate}
\renewcommand\arraystretch{0.75}
\newcommand{\tabincell}[2]{\begin{tabular}{@{}#1@{}}#2\end{tabular}}
\setlength{\tabcolsep}{0.1mm}{
\begin{tabular}{cc|cccccc|ccc|cccc}
\toprule
\multicolumn{2}{c|}{\multirow{3}{*}{\textbf{Corruptions}}}  & \multicolumn{6}{c|}{\textbf{LiDAR-Only}} & \multicolumn{3}{c|}{\textbf{Camera-Only}} & \multicolumn{4}{c}{\textbf{Multi-modal}}\\
& &
\multirow{1}{*}{SECOND $^{\dagger}$}  & 
\multirow{1}{*}{\tabincell{c}{PointPillars $^{\dagger}$}}  &
\multirow{1}{*}{\tabincell{c}{PointRCNN $^{\dagger}$}} & 
\multirow{1}{*}{PV-RCNN $^{\dagger}$} & 
\multirow{1}{*}{Part-A$^2 $ $^{\dagger}$} & 
\multirow{1}{*}{3DSSD $^{\dagger}$} & 
\multirow{1}{*}{SMOKE $^{\dagger}$} & 
\multirow{1}{*}{PGD $^{\dagger}$} & 
\multirow{1}{*}{\tabincell{c}{ImVoxelNet $^{\dagger}$}} & 
\multirow{1}{*}{EPNet $^{\dagger}$} & 
\multirow{1}{*}{\tabincell{c}{Focals Conv $^{\dagger}$}} & 
\multirow{1}{*}{LoGoNet *} & 
\multirow{1}{*}{\tabincell{c}{VirConv-S *}}\\

\midrule
\multicolumn{2}{c|}{\textbf{None}($\text{AP}_{\text{clean}}$)}  & 81.59 & 78.41 & 80.57 & 84.39 &82.45 &80.03 & 7.09 &8.10& 11.49 & 82.72 & 85.88 & 86.07 & 91.95  \\
\midrule
\multicolumn{1}{c|}{}                          & Snow           & 52.34 & 36.47 & 50.36 & 52.35 &42.70&27.12& 2.47&0.63 & 0.22  & 34.58 & 34.77 & 51.45 & 51.17  \\
\multicolumn{1}{c|}{}                          & Rain           & 52.55 & 36.18 & 51.27 & 51.58 &41.63&26.28& 3.94&3.06 & 1.24  & 36.27 & 41.30 & 55.80 & 50.57  \\
\multicolumn{1}{c|}{}                          & Fog            & 74.10 & 64.28 & 72.14 & 79.47 & 71.61&45.89&5.63&0.87 & 1.34  & 44.35 & 44.55 & 67.53 & 75.63 \\
\multicolumn{1}{c|}{\multirow{-4}{*}{Weather}} & Sunlight       & 78.32 & 62.28 & 62.78 & 79.91&76.45&26.09 & 6.00&7.07 & 10.08 & 69.65 & 80.97 & 75.54 & 63.62  \\
\midrule
\multicolumn{1}{c|}{}                          & Density        & 80.18 & 76.49 & 80.35 & 82.79&80.53&77.65 & -    & -   & -   & 82.09 & 84.95 & 83.68 & 80.70       \\
\multicolumn{1}{c|}{}                          & Cutout         & 73.59 & 70.28 & 73.94 & 76.09&76.08&73.05 & -    & -     & -    & 76.10 & 78.06 & 77.17 & 75.18      \\
\multicolumn{1}{c|}{}                          & Crosstalk      & 80.24 & 70.85 & 71.53 & 82.34&79.95&46.49 & -    & -     & -    & 82.10 & 85.82 & 82.00 & 75.67       \\
\multicolumn{1}{c|}{}                          & Gaussian (L)    & 64.90 & 74.68 & 61.20 & 65.11&60.73&59.14 & -     & -     & -     & 60.88 & 82.14 & 61.85 & 63.16     \\ 
\multicolumn{1}{c|}{}                          & Uniform (L)    & 79.18 & 77.31 & 76.39 & 81.16&77.77&74.91 &  -    &  -    & -     & 79.24 & 85.81 & 82.94 & 70.74     \\ 
\multicolumn{1}{c|}{}                          & Impulse (L)    & 81.43 & 78.17 & 79.78 & 82.81&80.80&78.28 &  -    &  -     & -    & 81.63 & 85.01 & 84.66 & 80.50    \\
\multicolumn{1}{c|}{}                          & Gaussian (C)    & -     & -     & -     & -  & -     & -     & 1.56 & 1.71& 2.43  & 80.64 & 80.97 & 84.29 & 82.55      \\
\multicolumn{1}{c|}{}                          & Uniform (C)     & -     & -     &      & -  & -     & -     & 2.67 & 3.29 & 4.85  & 81.61 & 83.38 & 84.45 & 82.56    \\
\multicolumn{1}{c|}{\multirow{-9}{*}{Sensor}}  & Impulse (C)    & -     & -     & -  & -   & -     & -     & 1.83 & 1.14& 2.13  & 81.18 & 80.83 & 84.20 & 82.54    \\
\midrule
\multicolumn{1}{c|}{}                          & Moving Obj.    & 52.69 & 50.15 & 50.54 & 54.60 &79.57&77.96& 1.67& 2.64 & 5.93  & 55.78 & 49.14 & 14.44 & 32.28       \\
\multicolumn{1}{c|}{\multirow{-2}{*}{Motion}}  & Motion Blur    & -     & -     & - & -  & - & -    & 3.51 & 3.36& 4.19  & 74.71 & 81.08 & 84.52 & 82.58       \\
\midrule
\multicolumn{1}{c|}{}                          & Local Density  & 75.10 & 69.56 & 74.24 & 77.63&79.57&77.96 & -    & - & -    & 76.73 & 80.84 & 78.63 & 78.73      \\
\multicolumn{1}{c|}{}                          & Local Cutout   & 68.29 & 61.80 & 67.94 & 72.29&75.06&73.22 & -    & - & -    & 69.92 & 76.64 & 64.88 & 71.01     \\
\multicolumn{1}{c|}{}                          & Local Gaussian & 72.31 & 76.58 & 69.82 & 70.44&77.44&75.11 & -    & -  & -    & 75.76 & 82.02 & 55.66 & 72.85     \\
\multicolumn{1}{c|}{}                          & Local Uniform  & 80.17 & 78.04 & 77.67 & 82.09&80.77&78.64 & -    & -  & -    & 81.71 & 84.69 & 79.94 & 79.61    \\
\multicolumn{1}{c|}{}  & Local Impulse  & 81.56 & 78.43 & 80.26 & 84.03 &82.25&79.53& -    & -  & -    & 82.21 & 85.78 & 84.29 & 82.07    \\
\multicolumn{1}{c|}{}                          & Shear & 41.64 & 39.63 & 39.80 & 47.72&37.08&26.56 & 1.68 & 2.99   & 1.33     & 41.43 & 45.77 & - & -   \\
\multicolumn{1}{c|}{}                          & Scale & 73.11 & 70.29 & 71.50 & 76.81&75.90&75.02 & 0.13  & 0.15   &0.33 & 69.05&69.48 & - & -   \\
\multicolumn{1}{c|}{\multirow{-8}{*}{Object}}                          & Rotation & 76.84 & 72.70 & 75.57 & 79.93&75.50&76.98   & 1.11&2.14   &2.57 & 74.62&77.76 & - & -   \\
\midrule
\multicolumn{1}{c|}{\multirow{-1}{*}{Alignment}}  & Spatial  & - & - & - & - &-&-& -  & -   & -     &35.14 & 43.01 & - & -    \\
\midrule
\multicolumn{2}{c|}{$\textbf{Average} (\text{AP}_{\text{cor}})$}& 70.45& 65.48& 67.74 & 72.59&69.92&60.55 &  2.68 &  2.42 & 3.05 & 67.81& 71.87 & 80.93 & 85.66 \\

\multicolumn{2}{c|}{RCE (\%) $\downarrow$ }& 13.65 & 16.49& 15.92 & 13.98 &  15.20 &  24.34 & 62.20 & 70.12 & 73.46& 22.03&18.02& 5.97&6.84  \\
\bottomrule
\end{tabular}
}
\begin{tablenotes}
\footnotesize
\item[1] $^{\dagger}$: Results from Ref. \cite{BR3D}.
\item[2] * denotes the result of our re-implementation.

\end{tablenotes}
\end{table*}

\begin{table*}[!ht]
\scriptsize
\caption{Comparison with SOTA methods on \textbf{nuScenes-C validation} set with \textbf{mAP}. `D.I.' refers to DeepInteraction~\cite{DeepInteraction}. `RCE' denotes Relative Corruption Error from Ref.\cite{BR3D}.}
\label{tb:nusc-map}
\renewcommand\arraystretch{0.75}
\newcommand{\tabincell}[2]{\begin{tabular}{@{}#1@{}}#2\end{tabular}}
\setlength{\tabcolsep}{1.5mm}{
\begin{tabular}{cc|ccc|cccc|cccc}
\toprule
\multicolumn{2}{c|}{\multirow{3}{*}{\textbf{Corruptions}}}  & 
\multicolumn{3}{c|}{\textbf{LiDAR-Only}} & 
\multicolumn{4}{c|}{\textbf{Camera-Only}} & 
\multicolumn{4}{c}{\textbf{Multi-modal}}\\
&
& 
\multirow{1}{*}{\tabincell{c}{PointPillars$^{\dagger}$}}  & 
\multirow{1}{*}{\tabincell{c}{SSN$^{\dagger}$}}  & 
\multirow{1}{*}{\tabincell{c}{CenterPoint$^{\dagger}$}} & 
\multirow{1}{*}{FCOS3D$^{\dagger}$} & 
\multirow{1}{*}{PGD$^{\dagger}$} & 
\multirow{1}{*}{DETR3D$^{\dagger}$} & 
\multirow{1}{*}{\tabincell{c}{BEVFormer$^{\dagger}$}} & 
\multirow{1}{*}{FUTR3D$^{\dagger}$} & 
\multirow{1}{*}{\tabincell{c}{TransFusion$^{\dagger}$}} & 
\multirow{1}{*}{BEVFusion$^{\dagger}$} & 
\multirow{1}{*}{\tabincell{c}{D.I.*}}  \\
\midrule
\multicolumn{2}{c|}{\textbf{None}($\text{AP}_{\text{clean}}$)}  & 27.69 & 46.65 & 59.28 & 23.86& 23.19 & 34.71 & 41.65 & 64.17 & 66.38 & 68.45 & 69.90  \\
\midrule
\multicolumn{1}{c|}{}                          & Snow           & 27.57 & 46.38 & 55.90 & 2.01 & 2.30  & 5.08  & 5.73  & 52.73 & 63.30 & 62.84 & 62.36   \\
\multicolumn{1}{c|}{}                          & Rain           & 27.71 & 46.50 & 56.08 & 13.00 & 13.51& 20.39 & 24.97 & 58.40 & 65.35 & 66.13 & 66.48    \\
\multicolumn{1}{c|}{}                          & Fog            & 24.49 & 41.64 & 43.78 & 13.53& 12.83 & 27.89 & 32.76 & 53.19 & 53.67 & 54.10 & 54.79    \\
\multicolumn{1}{c|}{\multirow{-4}{*}{Weather}} & Sunlight       & 23.71 & 40.28 &54.20 & 17.20 &22.77 & 34.66 & 41.68 & 57.70 & 55.14 & 64.42 & 64.93    \\
\midrule
\multicolumn{1}{c|}{}                          & Density        & 27.27 &46.14 & 58.60 & -     & -     & -  & -   & 63.72 & 65.77 & 67.79 &68.15      \\
\multicolumn{1}{c|}{}                          & Cutout         & 24.14 &40.95 & 56.28 & -     & -     & -  & -   & 62.25 & 63.66 & 66.18 &66.23      \\
\multicolumn{1}{c|}{}                          & Crosstalk      & 25.92 & 44.08&56.64 & -     & -     & -  & -   & 62.66 & 64.67 & 67.32 &68.12      \\
\multicolumn{1}{c|}{}                          & FOV lost       &  8.87 &15.40 &20.84 & -     & -     & -  & -   & 26.32 & 24.63 & 27.17 &  42.66  \\
\multicolumn{1}{c|}{}                          & Gaussian (L)    & 19.41&39.16  & 45.79 & -     & -     & - & -    & 58.94 & 55.10 & 60.64 & 57.46        \\
\multicolumn{1}{c|}{}                          & Uniform (L)     & 25.60 & 45.00&56.12 & -     & -     & -  & -   & 63.21 & 64.72 & 66.81 &  67.42       \\
\multicolumn{1}{c|}{}                          & Impulse (L)    & 26.44 &45.58 &57.67 & -     & -     & -   & -  & 63.43 & 65.51 & 67.54 & 67.41       \\
\multicolumn{1}{c|}{}                          & Gaussian (C)    &- &-&- & 3.96 &4.33 & 14.86 & 15.04 & 54.96 & 64.52 & 64.44 &  66.52    \\
\multicolumn{1}{c|}{}                          & Uniform (C)     & - & -   & -     & 8.12 &8.48 & 21.49 & 23.00 & 57.61 & 65.26 & 65.81 &  65.90     \\
\multicolumn{1}{c|}{\multirow{-9}{*}{Sensor}}  & Impulse (C)    & - & -   & -     & 3.55 &3.78 & 14.32 & 13.99 & 55.16 & 64.37 & 64.30 &  65.65   \\
\midrule
\multicolumn{1}{c|}{}                          & Compensation   & 3.85 &10.39 & 11.02 & -     & -     & -   & -   & 31.87 & 9.01 & 27.57 &  39.95    \\
\multicolumn{1}{c|}{}                          & Moving Obj.    & 19.38 &35.11& 44.30 & 10.36 &10.47& 16.63 & 20.22 & 45.43& 51.01& 51.63 &   -       \\
\multicolumn{1}{c|}{\multirow{-3}{*}{Motion}}  & Motion Blur    & -  & -  & -     & 10.19 & 9.64&11.06 & 19.79 & 55.99 & 64.39 & 64.74 &   65.45     \\
\midrule
\multicolumn{1}{c|}{}                          & Local Density  & 26.70 &45.42 & 57.55 & -     & -     & -  & -   & 63.60 & 65.65 & 67.42 &   67.71    \\
\multicolumn{1}{c|}{}                          & Local Cutout   & 17.97 &32.16 & 48.36 & -     & -     & -  & -     & 61.85 & 63.33 & 63.41 &   65.19   \\
\multicolumn{1}{c|}{}                          & Local Gaussian & 25.93 &43.71 &51.13 & -     & -     & -  & -     & 62.94 & 63.76 & 64.34 &   64.75    \\
\multicolumn{1}{c|}{}                          & Local Uniform  & 27.69 & 46.87&57.87 & -     & -     & -   & -    & 64.09 & 66.20 & 67.58 &   66.44     \\
\multicolumn{1}{c|}{} & Local Impulse  & 27.67 & 46.88 &58.49 & -     & -     & -  & -     & 64.02 & 66.29 & 67.91 &   67.86     \\
\multicolumn{1}{c|}{} & Shear  & 26.34 & 43.28& 49.57 &17.20 &16.66& 17.46 &24.71  & 55.42 &62.32& 60.72 &   -     \\
\multicolumn{1}{c|}{} & Scale  & 27.29& 45.98 &51.13 & 6.75& 6.57& 12.02 &17.64   & 55.42& 62.32& 60.72 &   -     \\
\multicolumn{1}{c|}{\multirow{-8}{*}{Obeject}} & Rotation  & 27.80 &46.93 & 54.68 & 17.21& 16.84& 27.28& 33.97    & 59.64& 63.36& 65.13 &   -     \\
\midrule
\multicolumn{1}{c|}{} & Spatial  & -& -&- &-&- & -&-   & 63.77 &66.22& 68.39&-     \\
\multicolumn{1}{c|}{\multirow{-2}{*}{Alignment}} & Temporal  & - &- & - & -& -& -& -    & 51.43& 43.65& 49.02&   -     \\
\midrule
\multicolumn{2}{c|}{$\textbf{Average} (\text{AP}_{\text{\text{cor}}})$} & 23.42&40.37 &49.81& 10.26& 10.68& 18.60 &22.79& 56.99 &58.73& 61.03 & 62.92  \\

\multicolumn{2}{c|}{$\text{RCE}(\%)\downarrow$}& 15.42 &13.46 & 15.98  & 57.00  & 53.95  & 46.89  & 46.41  & 11.45  & 11.52  & 10.84  &11.09  \\
\bottomrule
\end{tabular}
}
\begin{tablenotes}
\footnotesize
\item[1] $^{\dagger}$: Results from Ref. \cite{BR3D}.
\item[2] * denotes the result of our re-implementation.

\end{tablenotes}
\label{tab_nuscenes_C}
\end{table*}

\begin{figure*}[htbp]
	\centering
	\includegraphics[width=\linewidth]{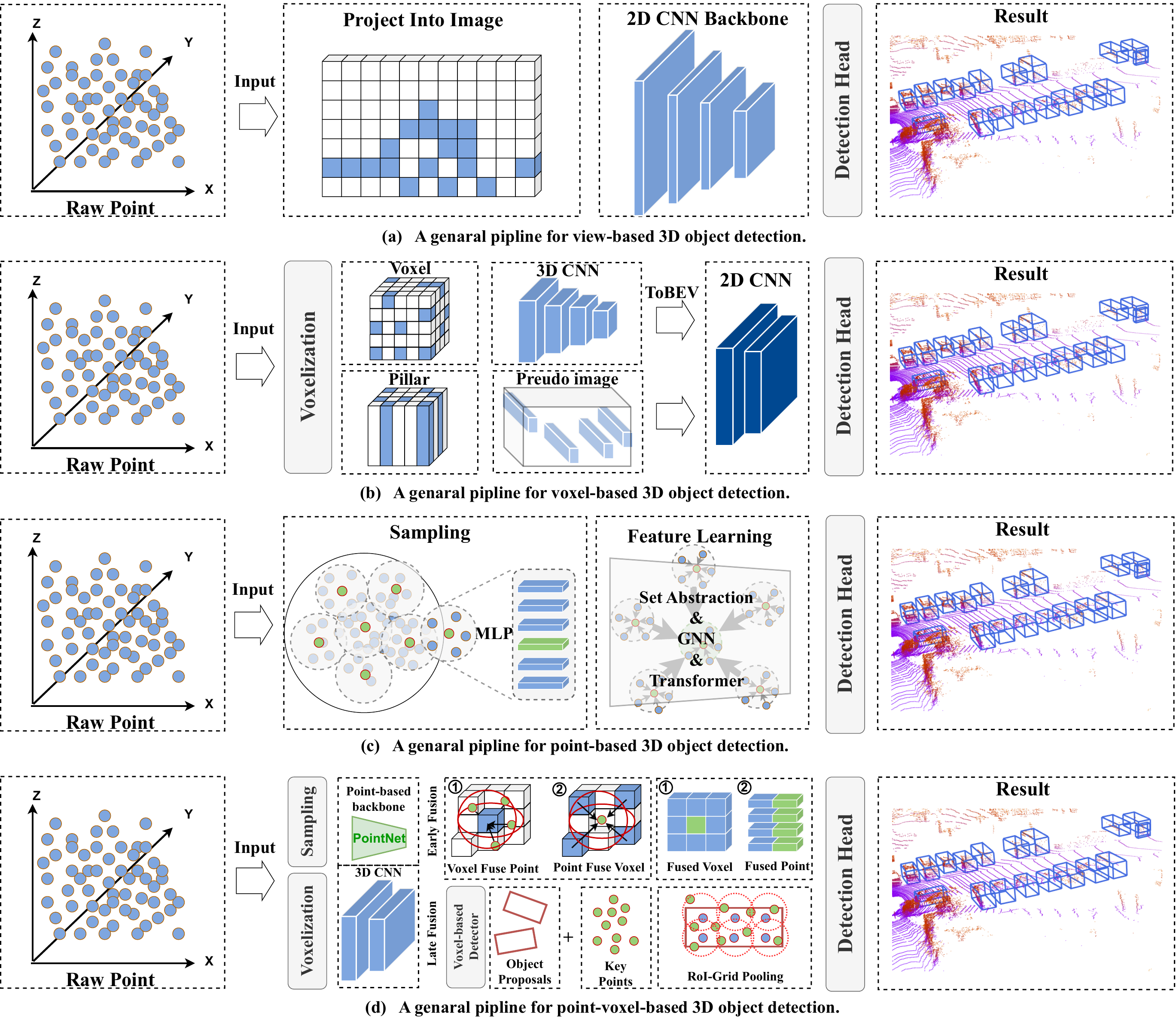}
	\caption{The genaral piplines for LiDAR-only 3D object detection.}
	\label{fig:allpipline}
\end{figure*}

\section{LiDAR-only 3D Object Detection}
\label{section:LiDAR-only}

LiDAR-only methods capture precise 3D information, leading to higher detection accuracy and robustness, particularly in extreme weather conditions~\cite{BR3D}.
Because in comparison to optical radiation, the laser beams emitted by LiDAR systems can penetrate certain weather disturbances, such as raindrops and haze, with slight interference. 
However, the high cost of LiDAR remains one of the main barriers to large-scale adoption of LiDAR-only methods. 
Generally, as shown in Fig.~\ref{fig:allpipline}, LiDAR-only methods can be categorized into four types: (1) view-based 3D object detection, (2) voxel-based 3D object detection, (3) point-based 3D object detection, (4) point-voxel-based 3D object detection. In contrast to previous reviews \cite{wang2023multi,mao20233d,wang2023multi_ijcv,wang2023multi_TITS,peng2022survey}, our survey extends beyond the conventional classifications of LiDAR-only methods. We adopt a more foundational idea to class LiDAR-only methods based on their core \textbf{data representations} (BEV, Voxel, Pillars.) and underlying \textbf{model structure} (CNNs, Transformers, PointNet). We provide a comprehensive understanding of the technological paradigms at LiDAR-only methods, analyzing and classifying these systems from a more essential, technical lineage perspective.

\subsection{View-based 3D object detection}

View-based methods transform point clouds into pseudo-images using BEV and range views. Based on the different data representation views, the view-based methods can be divided into two categories: 1) \textbf{Range View}, 2) \textbf{BEV View}. In these representations, each pixel contains 3D spatial information rather than RGB values. Due to the dense representation of pseudo-images, traditional or specialized 2D convolutions can be seamlessly applied to range images, making the feature extraction process highly efficient. However, compared to other LiDAR-only methods, \textcolor{black}{detection using range views is more susceptible to occlusion and scale variations.}

\subsubsection{\textbf{Range View}}
Due to the sparsity of point cloud data, projecting it directly onto an image plane results in a sparse 2D point map. Therefore, most methods~\cite{rangeioudet,rangercnn,RangeDet,lasernet,laserflow,rangecd} project point clouds into cylinder coordinates to generate a dense front-view representation by using the following projection fuction:

\begin{equation}
\begin{aligned}
\label{euqation:project}
&\theta  = atan2(y, x), \\
&\phi  = arcsin(z/\sqrt{x^{2}  + y^{2}  + z^{2}}), \\
&r = \left \lfloor \theta/\Delta \theta \right \rfloor, \\
&c = \left \lfloor \phi / \Delta \phi  \right \rfloor ,   \\
\end{aligned}
\end{equation}

\noindent where $p = (x, y, z)\Tr$ denotes a 3D point and $(r, c)$ denotes the 2D map position of its projection. 
$\theta$ and $\phi$ denote the azimuth and elevation angle when observing the point.
$\Delta \theta$ and $\Delta \phi$ are the average horizontal and vertical angle resolution between consecutive beam emitters, respectively.
VeloFCN~\cite{li2016vehicle} is an influential work that first introduces the projection method in cylindrical coordinates. It has then followed by~\cite{rangecd,RangeDet,rangeioudet,rangercnn}. LaserNet~\cite{lasernet} utilizes DLA-Net~\cite{DLA-Net} to obtain multi-scale features and detect 3D objects from this representation. Inspired by LaserNet, some works have borrowed models from 2D object detection to handle range images. For example, U-Net~\cite{U-net} is applied in~\cite{laserflow,rsn,rangercnn}, RPN~\cite{RPN} is employed in~\cite{rangercnn,rangecd}, and FPN~\cite{FPN} is leveraged in~\cite{RangeDet}. Considering the limitations of traditional 2D CNNs in extracting features from range images, some works have resorted to novel operators, including range dilated convolutions~\cite{rangecd}, graph operators~\cite{chai2021point}, and meta-kernel convolutions~\cite{RangeDet}. Furthermore, some works have focused on addressing issues of occlusion and scale variation in range view. Specifically, these methods~\cite{rangeioudet,rangercnn} \textcolor{black}{construct feature transformation structures from the range view to the point view and from the point view to the BEV (Bird's Eye View) perspective to convert range features into 
BEV perspective.}

\subsubsection{\textbf{BEV View}}
Comparison to range view detection, BEV-based detection is more robust to occlusion and scale variation challenges. Hence, feature extraction from the range view and object detection from the \textcolor{black}{BEV become} the most practical solution for range-based 3D object detection. The BEV representation is encoded by height, intensity, and density. Point clouds are discretized into a regular 2D grid. To encode more detailed height information, point clouds are evenly divided into $M$ slices, resulting in $M$ height maps where each grid cell stores the maximum height value of the point clouds. The intensity feature represents the reflectance value of the point within each grid cell, and the point cloud density indicates the number of points in each cell. PIXOR~\cite{pixor}, which outputs oriented 3D object estimates decoded from pixel-wise neural network predictions, is a pioneering work in this field, followed by~\cite{birdnet,birdnet+,rangeioudet,RangeDet}. These methods usually entail three stages. First, point clouds are projected into a novel cell encoding for BEV projection. Next, both the object's location on the plane and its heading are estimated through a convolutional neural network originally designed for image processing. Considering scale variation and occlusion, RangeRCNN~\cite{rangercnn} and RangeIOUDet~\cite{rangeioudet} introduce a point view that serves as a bridge from RV to BEV, which provides pointwise features for \textcolor{black}{the models}.

\subsection{Voxel-based 3D object detection}
Voxel-based methods segment sparse point clouds into regular voxels, achieving \textcolor{black}{a} dense representation through voxelization. Despite spatial convolution enhancing 3D information perception, challenges persist in achieving high detection accuracy. These challenges include 1) \textbf{high computational complexity}, \textcolor{black}{which demands} substantial memory and computational resources due to the numerous voxels representing 3D space, 2) \textbf{spatial information loss} that occurs during voxelization, leading to difficulties in accurately detecting small objects, 3) \textbf{inconsistencies in scale and density}, inherent to specific voxel grids, which pose challenges in adapting to diverse scenes with varying object scales and point cloud densities. Overcoming these challenges requires addressing \textcolor{black}{limitations in} data representation, enhancing network feature capacity, improving object localization accuracy, and enhancing the model's understanding of complex scenes. Ensuring safety perception in autonomous driving is crucial, and despite varying optimization strategies, these methods converge on common perspectives of model optimization, focusing on 1) \textbf{data representation} and 2) \textbf{model structure}.

\subsubsection{\textbf{Data representation}}
Voxel-based methods first rasterize point clouds into discrete grid representations. Grid representations are closely related to accuracy, computational complexity, and memory requirements. \textcolor{black}{Using a voxel size that is too large results in significant information loss, while using a voxel size that is too small increases the burdens of computation and memory. As shown in Fig.~\ref{fig:allpipline} (b), according to the height along the z-axis, the types of grid representations can be categorized into voxels and pillars.}


\paragraph{\textit{\textbf{Voxel}}}
Voxel process divides the 3D space into regular voxel grids with size (${d_{L}}\times{d_{W}}\times{d_{H}}$) in the x, y, and z directions, respectively.
Only non-empty voxel units that contain points are stored and used for feature extraction. However, due to the sparse distribution of point clouds, the majority of voxel units are empty. 
As a pioneering work in voxel-based methods~\cite{voxelnext,VoxelFPN,Voxelrcnn,Second,LargeKernel3D,Link,3dssd,Cia-ssd,SE-SSD,vpnet,CenterNet3D,AFDetV2,Centerpoint,casa,afdet,TED,UVTR,Transfusion,Voxelnet}, VoxelNet~\cite{Voxelnet} proposes a novel voxel feature encoding (VFE) layer to extract features from the points inside a voxel cell.
Then, following works~\cite{Voxelrcnn,CBGS,afdet,AFDetV2,Centerpoint,DGCNN} have extended the VoxelNet network by adopting similar voxel encoding approaches. Existing methods often perform local partitioning and feature extraction uniformly across all positions in the point cloud. This approach limits the receptive field for distant regions and information truncation.
Therefore, some works have proposed different approaches to voxel partitioning: 1) \textbf{Different coordinate systems}: 
some approaches have reexamined voxel partitioning from different coordinate system perspectives, e.g.~\cite{cylinder3d, cvcnet2020} from cylindrical and~\cite{lai2023spherical} from spherical coordinate systems. 
Sphereformer~\cite{lai2023spherical} facilitates the aggregation of information from sparsely distant points by dividing the 3D space into multiple non-overlapping radial windows using spherical coordinates ($r$, $\theta$, $\phi$), thereby enhancing information integration from dense point regions. 2) \textbf{Multi-scale voxels}: some works generate voxels of different scales~\cite{VoxelFPN,hvnet} or use reconfigurable voxels~\cite{Reconfigurable_Voxels}, e.g., HVNet~\cite{hvnet} proposes a hybrid voxel network which integrates different scales in the point-level voxel feature encoder (VFE).

\paragraph{\textit{\textbf{Pillars}}}

Pillars can be considered a special form of voxels. Specifically, point clouds are discretized into a grid uniformly distributed on the x-y plane without binning along the z-axis. Pillar features can be aggregated from points through a PointNet~\cite{Pointnet} and then scattered back to construct a 2D BEV image for feature extraction. As the pioneering work in this series~\cite{pillarnet,Pointpillars,SWFormer,pillarNeXt,Seformer,Sparsetransformer}, PointPillar~\cite{Pointpillars} first introduces the pillar representation. \textcolor{black}{Following} works have extended the ideas from 2D detection to PointPillars. PillarNet~\cite{pillarnet} adopts the 'encoder-neck-head' detection architecture to enhance the performance of pillar-based methods. SWFormer~\cite{SWFormer} and ESS~\cite{Sparsetransformer} draw inspiration from the Swin Transformer~\cite{Swintransformer} and apply a hierarchical window mechanism to pseudo-images, thereby enabling the network to maintain a global receptive field. PillarNeXt~\cite{pillarNeXt} integrates a series of mature 2D detection techniques and achieves performance comparable to voxel-based methods.
 
\subsubsection{\textbf{Model Structure}}

There are three major types of neural networks \textcolor{black}{in} voxel-based methods: 1) 2D CNNs for processing BEV feature maps and pillars. 2) 3D Sparse CNNs for processing voxels. 3) Transformers for handling both voxels and pillars.

\paragraph{\textit{\textbf{2D CNNs}}}
2D CNNs \textcolor{black}{are} primarily used to detect 3D objects from a bird's-eye view perspective, including processing BEV (Bird's Eye View) feature maps and pillars~\cite{Pointpillars,pillarnet,pillarNeXt,Seformer,SWFormer}. Specifically, the 2D CNNs used for processing BEV feature maps often come from well-developed 2D object detection networks, such as Darknet~\cite{yolov3}, ResNet~\cite{resnet}, FPN~\cite{FPN}, and RPN~\cite{RPN}. One significant advantage of 2D CNNs compared to 3D CNNs is their faster speed. However, due to their difficulty in capturing spatial relationships and shape information, 2D CNNs typically \textcolor{black}{exhibit} lower accuracy.

\paragraph{\textit{\textbf{3D Sparse CNNs}}}
3D Sparse CNNs \textcolor{black}{consist} of two core operators: sparse convolution and submanifold convolution~\cite{sparsecnn}, which ensure that the convolutional operation is performed only on non-empty voxels. SECOND~\cite{Second} implements efficient computation of sparse convolution~\cite{sparsecnn} and submanifold convolution~\cite{Submanifoldcnn} operators to gain fast inference speed by constructing a hash table. It is followed by~\cite{voxelnext,Centerpoint,Voxelrcnn,VoxelNextFusion,afdet}. However, the limited receptive field of 3D Sparse CNNs, which \textcolor{black}{leads} to information truncation, restricts the model's feature extraction capabilities. Meanwhile, the sparse representation of features makes it challenging for the model to capture fine-grained object boundaries and detailed information. To optimize these issues, main optimization strategies have emerged: 1) Expanding the model's receptive field. Some methods ~\cite{Link,LargeKernel3D} extend the concept of large kernel convolution from 2D to 3D space or introduce additional downsampling layers in the model~\cite{voxelnext}. 2) Combining sparse and dense representations. Methods in this category typically utilize dense prediction heads to prevent information loss~\cite{Centerpoint,Voxelrcnn,Voxelnet,Second,part2} or retrieve lost 3D information from the detection process~\cite{Voxelrcnn,pyramidrcnn,part2,CenterNet3D,Centerpoint}, or they add additional auxiliary tasks to the model~\cite{part2,SegVoxelNet,AFDetV2,afdet,Cia-ssd}. Methods employing dense prediction heads typically require high-resolution Bird's Eye View (BEV) feature maps for conducting dense predictions on them. Considering computational complexity, some recent methods aim to establish global sparse and local dense prediction relationships~\cite{FSD}.

\paragraph{\textit{\textbf{Transformer}}}
Due to the amazing performance of transformers~\cite{Swintransformer,vit}, many efforts have been made to adapt Transformers to 3D object detection. Particularly, recent studies~\cite{RoboBEV,BR3D} have confirmed the excellent robustness of transformer-based models, which will further advance research in the domain of safety perception for autonomous driving. Compared with CNNs, the query-key-value design and the self-attention mechanism allow transformers to model global relationships, resulting in a larger receptive field. However, the primary limitation for efficiently applying Transformer-based models is the quadratic time and space complexity of the global attention mechanism. Hence, designing specialized attention mechanisms for Transformer-based 3D object detectors is critical. Transformer~\cite{transformer}, DETR~\cite{detr}, and ViT~\cite{vit} are the works that have most significantly influenced 3D transformer-based methods~\cite{voxeltransformer,Sparsetransformer,SWFormer,Seformer,Transfusion,sparsefusion,vsettransformer}. They have each inspired subsequent 3D detection works in various aspects: the design of attention mechanisms, the architecture of encoders and decoders, and the development of patch-based inputs and architectures similar to visual transformers. Inspired by transformer~\cite{transformer}, VoTr~\cite{voxeltransformer} is the first work to incorporate a transformer into a voxel-based backbone network, composed of sparse attention and sparse submanifold attention modules. Subsequent works~\cite{vsettransformer} have continued to build on the foundation of voxel-transformer, further optimizing the temporal complexity of the attention mechanism. DETR~\cite{detr} has inspired a range of networks to adopt an encoder-decoder structure akin to DETR's. TransFusion~\cite{Transfusion} is a notable work that generates object queries from initial detections, applying cross-attention to LiDAR and image features within the Transformer decoder for 3D object detection. Meanwhile, many papers~\cite{Seformer,SWFormer,Sparsetransformer} are exploring and refining the patch-based input mechanism from ViT~\cite{vit} and the window attention mechanism from Swin Transformer~\cite{Swintransformer}, e.g., SST~\cite{Sparsetransformer} and SWFormer~\cite{SWFormer} group local regions of voxels into patches, apply sparse regional attention, and then apply region shift to change the grouping. Notably, SEFormer~\cite{Seformer} is the first to introduce object structure encoding into the transformer module.

\subsection{Point-based 3D object detection}
Unlike \textcolor{black}{voxel-based methods, point-based methods retain the original information to the maximum extent, facilitating fine-grained feature acquisition. However, the performance of point-based methods is still affected by two crucial factors: 1) the number of contextual points in the point cloud sampling stage and 2) the context radius used in the point-based backbone. These factors significantly impact the speed and accuracy of point-based methods, including the detection of small objects, which is critical for safety considerations. Therefore, optimizing these two factors is paramount, based on existing literature.} In this regard, we primarily focus on elucidating 1) \textbf{Point Cloud Sampling} and 2) \textbf{Point-based Backbone}.

\subsubsection{\textbf{Point Cloud Sampling}}

As an extensively utilized method, FPS (Farthest Point Sampling) aims to select a set of representative points from the raw points, such that their mutual distances are maximized, thereby optimally covering the entire spatial distribution of the point cloud.

\textcolor{black}{PointRCNN~\cite{Pointrcnn}, a pioneering two-stage detector in point-based methods, utilizes the PointNet++~\cite{Pointnet++} with multi-scale grouping as the backbone network. In the first stage, it generates 3D proposals from point clouds in a bottom-up manner. The second stage network refines the proposals by combining semantic features and local spatial features. However, existing methods relying on FPS still face several issues:}
1) Points irrelevant to detection also participate in the sampling process, leading to additional computational burden.
2) The distribution of points across different parts of an object is uneven, resulting in suboptimal sampling strategies. Subsequent works have attempted various optimization strategies, such as segmentation-guided background point filtering~\cite{sasa}, random sampling~\cite{StarNet}, feature space sampling~\cite{3dssd}, voxel-based sampling~\cite{Graph-RCNN,Point-gnn}, coordinate refinement~\cite{3dobwithpointformer}, and ray-based grouping sampling~\cite{rbgnet}.

\subsubsection{\textbf{Point-based Backbone}}

\textcolor{black}{The feature learning stage in point-based methods aims to extract discriminative feature representations from raw points. The neural network used in the feature learning phase should possess the ability of to local be awareness locally aware and integrating to integrate contextual information.}
Based on the aforementioned motivations, a multitude of detectors have been designed for processing raw points. However, most methods can be categorized according to the core operators they utilize: 1) PointNet-based methods~\cite{Pointrcnn,Std,sasa,3D-CenterNet}. 2) Graph Neural Network-based methods~\cite{Point-gnn,PointRGCN,StarNet,RGN,svganet}. 3) Transformer-based methods~\cite{3dobwithpointformer,Group-Free}.

\paragraph{\textit{\textbf{PointNet-based}}}
PointNet-based methods~\cite{Pointrcnn,Std,sasa,3D-CenterNet} primarily rely on the Set Abstraction~\cite{Pointnet} to perform downsampling on raw points, aggregation of local information, and integration of contextual information, while preserving the symmetry invariance of the raw points.
Point-RCNN~\cite{Pointrcnn}, as the first two-stage work in point-based methods, achieved amazing performance at its time; however, it still faces the issue of high computational cost.
Subsequent work~\cite{ipod,sasa} has addressed this issue by introducing an additional semantic segmentation task during the detection process to filter out background points that contribute minimally to detection.
Furthermore, some efforts have focused on resolving the issue of the uncontrolled receptive field in PointNet \& PointNet++, such as through the use of GNN~\cite{DGCNN} or Transformer~\cite{3dobwithpointformer} techniques.

\paragraph{\textit{\textbf{Graph-based}}}
GNNs (Graph Neural Networks) possess key elements such as an adaptive structure, dynamic neighborhood, the capability to construct both local and global contextual relationships, and robustness against irregular sampling. These characteristics naturally endow GNNs with an advantage in handling irregular point clouds. Point-GNN~\cite{Point-gnn}, a pioneering work, designs a one-stage graph neural network to predict objects with an auto-registration mechanism, merging, and scoring operations, which demonstrate the potential of using graph neural networks as a new approach for 3D object detection. Most graph-based, point-based methods~\cite{Point-gnn,PointRGCN,StarNet,RGN,PC-RGNN} aim to fully utilize contextual information. This motivation has led to further improvements in subsequent works~\cite{RGN,PC-RGNN}.

\paragraph{\textit{\textbf{Transformer-based}}}

Up to this point, a series of methods~\cite{pointtransformer,Group-Free,3dobwithpointformer,3D-QueryIS} have explored the use of transformers for feature learning in point clouds, \textcolor{black}{achieving excellent results}. Pointformer~\cite{3dobwithpointformer} introduced local and global attention modules for processing 3D point clouds. The local transformer module models interactions among points within local areas, with the aim of learning contextually relevant regional features at the object level. The global transformer, on the other hand, focuses on learning context-aware representations at the scene level. Subsequently, the local-global Transformer combines local features with high-resolution global features to further capture dependencies between multi-scale representations. Group-free~\cite{Group-Free} adapted the Transformer to suit 3D object detection, enabling it to model both object-to-object and object-to-pixel relationships and to extract object features without manual grouping. Moreover, by iteratively refining the spatial encoding of objects at different stages, the detection performance is further enhanced. Point-based transformers directly process unstructured and unordered raw point clouds, which results in significantly higher computational complexity compared to structured voxel data.

\subsection{Point-Voxel based 3D object detection}

 Point-voxel methods aim to leverage the fine-grained information capture capabilities of point-based methods and the computational efficiency of voxel-based methods. By integrating these methods, point-voxel based methods enable a more detailed processing of point cloud data, capturing both the global structure and micro-geometric details. This is critically important for safety perception in autonomous driving, as the accuracy of decisions made by autonomous driving systems depends on high-precision detection results.

The key goal of point-voxel methods is to enable feature interplay between voxels and points via point-to-voxel or voxel-to-point transformations. The idea \textcolor{black}{of leveraging point-voxel} feature fusion in backbones has been explored by many works~\cite{pvcnn,tang2020searching,vtop,PVGNet,M3DETR,pvrcnn++,HVPR,Pv-rcnn,fastpoint_rcnn,PDV,lidarrcnn,pyramidrcnn,CT3D}. These methods fall into two categories: 1) \textit{\textbf{Early Fusion.}} Early fusion methods ~\cite{pvcnn,tang2020searching,vtop,PVGNet,M3DETR,HVPR} fuse voxel features and point features within the backbone network. 2) \textit{\textbf{Late Fusion.}} Late fusion methods~\cite{pvrcnn++,Pv-rcnn,fastpoint_rcnn,PDV,lidarrcnn,pyramidrcnn,CT3D} typically employ a two-stage detection approach, \textcolor{black}{using} voxel-based methods for initial proposal box generation, followed by sampling and refining key point features from the point cloud to enhance 3D proposals.

\paragraph{\textit{\textbf{Early Fusion}}}
Some methods~\cite{pvcnn,tang2020searching,vtop,PVGNet,HVPR,M3DETR} have explored using new convolutional operators to fuse voxel and point features, with PVCNN~\cite{pvcnn} potentially being the first work in this direction. In this method, the voxel-based branch initially converts points into a low-resolution voxel grid and aggregates neighboring voxel features through convolution. Then, \textcolor{black}{the} voxel-level features are transformed back into point-level features and fused with the features obtained from the point-based branch. Following closely, SPVCNN~\cite{tang2020searching}, which builds upon PVCNN, extends PVCNN to the domain of object detection. Other methods attempt to make improvements from other perspectives, such as auxiliary tasks~\cite{vtop} or \textcolor{black}{multi-scale} feature fusion~\cite{PVGNet,HVPR,M3DETR}.


\paragraph{\textit{\textbf{Late Fusion}}}

The methods in this series predominantly adopt a two-stage detection framework. Initially, voxel-based methods are employed to generate preliminary object proposals. This is followed by a refinement phase, where point-level features are leveraged for the precise delineation of detection boxes. As a milestone in PV-based methods, PV-RCNN~\cite{Pv-rcnn} utilizes SECOND~\cite{Second} as the first-stage detector and proposes a second-stage refinement stage with a RoI grid pool for the fusion of keypoint features. Subsequent works \textcolor{black}{have followed} the aforementioned paradigm, focusing on advancements in second-stage detection. Notable developments include \textcolor{black}{the use of} attention mechanisms~\cite{infofocus,pyramidrcnn,CT3D}, scale-aware pooling~\cite{lidarrcnn}, and point density-aware refinement modules~\cite{PDV}.

PV-based methods simultaneously possess the computational efficiency of voxel-based approaches and the capability of point-based methods to capture fine-grained information. However, constructing point-to-voxel or voxel-to-point relationships, along with the feature fusion of voxels and points, incurs additional computational overhead. Consequently, compared to voxel-based methods, PV-based methods can achieve better detection accuracy and robustness, but at the cost of increased inference time.

\subsection{Analysis: Accuracy, Latency, Robustness}
\label{Sec:Analysis:Lidar}
In the autonomous driving sector, the development of LiDAR-only 3D object detection solutions is advancing rapidly. A series of works~\cite{wang2023multi,mao20233d,wang2023multi_ijcv,wang2023multi_TITS,peng2022survey,arnold2019survey} have comprehensively summarized the current technological roadmaps, such as the extensive review of LiDAR-only solutions by the Shanghai AI Lab and SenseTime Research~\cite{mao20233d}. However, there is a lack of summarization and guidance from the perspective of safety perception and cost impact in autonomous driving. Therefore, in this section, following an analysis of the technological roadmaps and the current state of LiDAR-only solutions, we intend to base our discussion on the fundamental principles of `Accuracy, Latency, and Robustness.' It aims to guide the practical implementation of economically efficient and safe sensing in autonomous driving.


\subsubsection{\textbf{Accuracy}}
\textcolor{black}{
Referring to Section~\ref{section:camera-based} on Camera-only methods, we investigated the core factors influencing LiDAR-only methods. Representative methods from each category underwent comparative performance analysis on the KITTI \cite{kitti} and nuScenes \cite{nuscenes} datasets, as shown in Fig~\ref{fig:zhu_all} (c, d). The current scenario indicates that the latest view-based methods exhibit lower performance compared to other categories. View-based approaches transform point clouds into pseudo-images for processing with 2D detectors, which favors inference speed but sacrifices 3D spatial information. Therefore, an effective representation of 3D spatial information is pivotal for LiDAR-only methods.
Initially, point-based and PV-based methods outperformed voxel-based approaches in LiDAR-only detection. Over time, methods like Voxel RCNN~\cite{Voxelrcnn}, which utilize ROI pool modules for fine-grained information aggregation, have achieved comparable or superior performance. Voxel RCNN's ROI pooling module effectively addresses the loss of detailed 3D spatial information resulting from voxelization.}

\subsubsection{\textbf{Latency}}
Section~\ref{section:camera-based} highlights latency's importance in autonomous driving safety and user experience. 
While Camera-only methods tend to outperform LiDAR-only methods in terms of inference speed, the latter still maintain a competitive edge due to their accurate 3D perception. We conducted tests using an A100 graphics card to measure the FPS of significant LiDAR-only approaches, and evaluated their performance using the original research's AP and NDS metrics.
As shown in Table \ref{tab:lidar_latency}, it indicates that view-based methods excel in model latency due to the reduction in point cloud dimensions and the efficiency of 2D CNNs. Voxel-based methods achieve exceptional inference speed due to the use of structured voxel data and well-optimized 3D sparse convolutions. However, point-based methods face challenges in applying efficient operators during data preprocessing and feature extraction stages due to the irregular representation of point clouds. Point-GNN~\cite{Point-gnn} is an extreme example of this, with model latency nearly several times that of contemporary voxel algorithms.
Transformer-based methods~\cite{CT3D} face significant challenges in real-time inference. The current research trend in transformer-based methods is the development of efficient attention operators, like~\cite{Sparsetransformer,SWFormer,Swintransformer}. Moreover, for PV-based methods, the construction of point-to-voxel or voxel-topoint relationships, along with the feature fusion of voxels and points, incurs additional computational overhead.
To conclude, common accuracy optimization strategies, such as two-stage optimization or attention mechanisms, typically compromise inference speed in autonomous driving models. Achieving a balance between accuracy and speed is an evolving challenge in this field. Future studies should prioritize the simultaneous improvement of accuracy, as well as the reduction of FPS (frames per second) and latency, in order to meet the urgent requirements of real-time response and safety in autonomous driving.

\subsubsection{\textbf{Robustness}}
\label{sec:point_robustness}

Previous comprehensive reviews have not focused significantly on the topic of robustness. Presently, research works\cite{RoboBEV,BR3D,zhu2023understanding,BRLCF,kong2023robo3d,zhang2023comprehensive,Tanet} like RoboBEV\cite{RoboBEV}, Robo3D\cite{kong2023robo3d}  on 3D object detection incorporate considerations of robustness, exemplified by factors such as sensor misses. Robo-LiDAR~\cite{robotrans} represents the first comprehensive exploration solely dedicated to the robustness of LiDAR-only methods. In a manner akin to BR3D~\cite{BR3D}, this method evaluates robustness by integrating disturbances into datasets pertinent to 3D object detection, such as KITTI~\cite{kitti}. The method involves proposing a variety of noise types and 25 typical degradations associated with object and scene-level natural weather conditions, noise interferences, density variations, and object transformations. In this section, we will combine the work of Ref.~\cite{BR3D} and Robo-LiDAR~\cite{robotrans} with the aim of systematically analyzing the robustness of LiDAR-only methods. As shown in the Table~\ref{tab:kittic-car-moderate}, generally, LiDAR-only methods exhibit higher robustness to noise compared to Camera-only methods. In Multi-modal methods~\cite{epnetpami,focalconv,Transfusion,virconv,FUTR3D,Logonet,DeepInteraction}, the complementary interplay of data types becomes evident when disturbances are limited to LiDAR sensor data. In such scenarios, image data can partially mitigate the impact on  point cloud integrity, consequently elevating the performance of fusion methods above that of methods relying solely on LiDAR. when disturbances affect both image and point cloud data concurrently, the efficacy of most Multi-modal methods significantly diminishes. \textcolor{black}{It is worth noting that DTS \cite{DTS} and Bi3D \cite{Bi3D} enhance model robustness through domain adaptation methods.}

As shown in  Table~\ref{table:lidarnoise}, under various noise conditions, LiDAR-only methods experience varying degrees of accuracy decline, with the most significant reduction observed in extreme weather noise scenarios. These results indicate an urgent need in the field of autonomous driving to address the robustness issue of point cloud detectors. For most types of corruptions, voxel-based methods generally exhibit greater robustness than point-based methods, as shown in Table~\ref{tab:kittic-car-moderate},~\ref{tb:nusc-map},~\ref{table:lidarnoise}. A plausible explanation is that voxelization, through the spatial quantization of a group of adjacent points, mitigates the local randomness and spatial information disruption caused by noise and density degradation. Specifically, for severe corruptions (e.g., shear, FFD in the transformation), the point-voxel-based method~\cite{Pv-rcnn} exhibits greater robustness. PointRCNN~\cite{Pointrcnn} does not show the highest robustness against any form of corruption, highlighting potential limitations inherent in point-based methods. In conclusion, future works should explore robustness optimization from the perspectives of data representation and model architecture. The above analysis aims to offer valuable insights for future work related to robustness.

\begin{table}[!htbp]
\scriptsize
\caption{Comparsion with LiDAR-only detectors on corrupted validation sets of \textbf{KITTI} from Ref.~\cite{robotrans} on \textbf{Car} detection with $\text{CE}_{\text{AP}}(\%)$. $\text{CE}_{\text{AP}}(\%)$ denotes \textbf{Corruption Error} from Ref.~\cite{robotrans}. The best one is highlighted in \textbf{bold}. `T.F.' denotes Transformation.}
\label{table:lidarnoise}
\renewcommand\arraystretch{0.75}
\setlength{\tabcolsep}{0.3mm}{
\begin{tabular}{c|c|c|c|c|ccc|c}
\toprule
\multicolumn{3}{c|}{\multirow{2}{*}{\textbf{Corruption}}} & \multicolumn{1}{c|}{\textbf{PV}} & \multicolumn{1}{c|}{\textbf{Point}} & \multicolumn{3}{c|}{\textbf{Voxel}} & \multirow{2}{*}{\textbf{Avg.}} \\ \cmidrule{4-8}
\multicolumn{3}{c|}{} & \multicolumn{1}{c|}{PV-RCNN} & \multicolumn{1}{c|}{PointRCNN} & \multicolumn{1}{c}{SECOND} & \multicolumn{1}{c}{SE-SSD} & \multicolumn{1}{c|}{CenterPoint} & \\ \cmidrule{1-9}
\multirow{14}{*}{\begin{turn}{90}\textbf{Scene-level}\end{turn}} & \multirow{3}{*}{\textbf{Weather}} & rain & 25.11 & 23.31  &\textbf{21.81}  &29.51 &25.83 &26.45 \\
 &  & snow & 44.23 & 37.74  &\textbf{34.84}  &49.19 &38.74 &45.64 \\
 &  & fog & 1.59 & 3.52  &1.60  &1.59 &\textbf{1.11} &1.88 \\ \cmidrule{2-9}
 & \multirow{5}{*}{\textbf{Noise}} & uniform\_rad & 10.19 & 8.32  &9.51  &9.34 &8.15 &7.82 \\
 &  & gaussian\_rad & 13.02 & 9.98  &12.13  &11.02 &10.17 &9.65 \\
 &  & impulse\_rad & 2.20 & 3.86  &2.23  &\textbf{1.18} &1.8 &2.46 \\
 &  & background & 2.93 & 6.49  &2.41  &2.14 &1.86 &2.46 \\
 &  & upsample & 0.81 & 1.84  &\textbf{0.31}  &0.55 &0.46 &0.75 \\ \cmidrule{2-9}
 & \multirow{5}{*}{\textbf{Density}} & cutout & 3.75 & 3.97  &4.27  &4.26 &4.11 &4.0 \\
 &  & local\_dec & 14.04 & -  &13.88  &17.01 &14.64 &14.44 \\
 &  & local\_inc & 1.40 & 3.34  &1.33 &\textbf{0.90} &0.95 &1.68 \\
 &  & beam\_del & 0.58 & 0.79  &0.73  &1.07 &\textbf{0.47} &0.73 \\
 &  & layer\_del & 2.94 & 3.46  &3.10  &3.37 &\textbf{2.67} &3.17 \\ \midrule
\multirow{12}{*}{\begin{turn}{90}\textbf{Object-level}\end{turn}} & \multirow{4}{*}{\textbf{Noise}} & uniform & 15.44 & 12.95 & 9.48  & 6.99 & 6.51 & 8.94 \\
 &  & gaussian & 20.48 & 17.62 & 12.98 & 9.56 & 9.49 & 12.42 \\
 &  & impulse & 3.3 & 4.7 & 2.53 & 2.2 & \textbf{2.11} & 3.26 \\ 
 &  & upsample & 1.12 & 1.95 & 0.67 & 0.22 & 0.16 & 0.74 \\ \cmidrule{2-9}
 & \multirow{3}{*}{\textbf{Density}} & cutout & 15.81 & 15.62 & 14.99  & 16.51 & \textbf{14.06} & 15.47 \\
 &  & local\_dec & 14.38 & 14.16 & 13.23 & 15.08 & \textbf{12.52} & 13.84 \\
 &  & local\_inc & 13.93 & 14.19 & 13.74 & \textbf{11.03} & 11.64 & 12.81 \\ \cmidrule{2-9}
 & \multirow{5}{*}{\textbf{T.F.}} & shear & \textbf{37.27} & 40.96 & 40.35 & 40.35 & 40.0 & 39.71 \\
 &  & FFD & \textbf{32.42} & 38.88 & 33.15 &  37.96 & 32.86 & 34.93 \\
 &  & rotation & 0.60 & 0.47 & 0.31 &\textbf{0.27} & 0.38 & 0.52 \\
 &  & scale & \textbf{5.78} & 8.13 & 6.96 & 6.53 & 7.50 & 6.97 \\
 &  & translation & 3.82 & 3.03 & 3.24 &  \textbf{1.37} & 3.91 & 3.77 \\ \midrule
\multicolumn{3}{c|}{\textbf{mCE}} & 11.49 & 11.64 & 10.60 & 11.17 & \textbf{10.09} & 11.01 \\ \bottomrule
\end{tabular}}
\end{table}

\section{Multi-modal 3D Object Detection}
\label{section:multimodal-based}

\begin{figure}[t]
    \centering
\includegraphics[width=\linewidth]{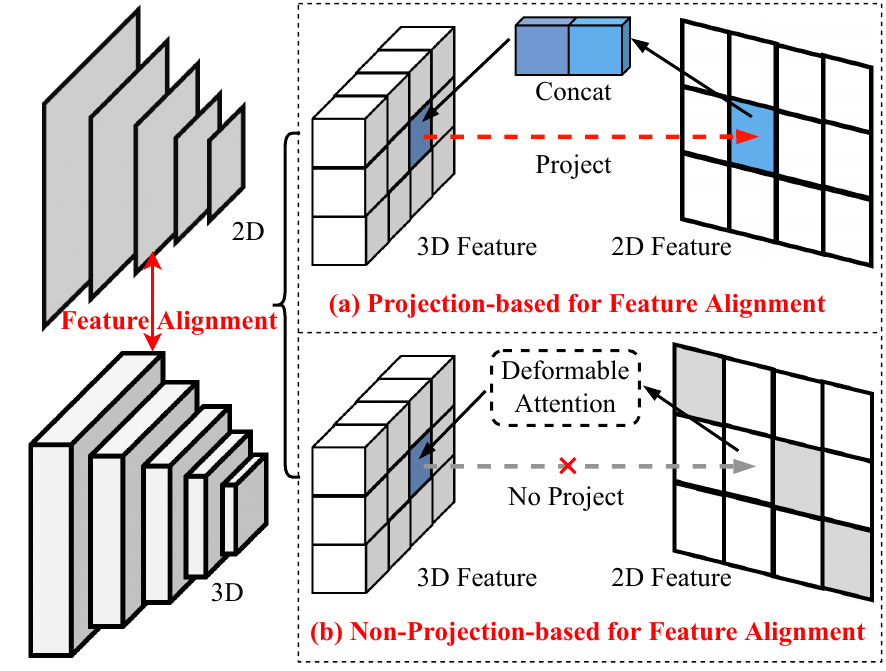}
    \caption{ Projection-based for feature alignment vs. Non-Projection-based for feature alignment.}
    \label{fig:multimodal_pipeline}
\end{figure}

Multi-modal 3D object detection refers to the technique of using data features from different sensors and integrating these features to achieve complementarity, thus enabling the detection of 3D objects. As shown in Fig. \ref{fig:multimodal_pipeline}, the approach particularly emphasizes the combination of image data and point cloud data. Image data is rich in semantic features, such as color and texture, but often lacks depth information. In contrast, point cloud data provides depth information and geometric structure, which is crucial for accurately perceiving and interpreting the 3D characteristics of a scene. Since a single type of sensor cannot fully and accurately perceive the 3D environment, multi-modal 3D object detection acquires features with rich semantic information by fusing various types of data.

In the field of autonomous driving, there are a variety of fusion methods for multi-modal 3D object detection. Previous reviews \cite{wang2023multi,mao20233d,wang2023multi_ijcv,wang2023multi_TITS,peng2022survey} have mostly classified these methods based on different stages of fusion (early, middle, late), but this classification is overly simplistic and does not fully consider the special requirements of autonomous driving. Given the fundamental differences between the two heterogeneous modalities of point clouds and images, the alignment step in multi-modal fusion is particularly critical. It ensures the consistency and accuracy of information from different sensors and data sources during the fusion process. In autonomous driving, the key to achieving feature alignment lies in whether to use a calibration matrix (also known as a projection matrix). However, the inherent error of the calibration matrix, being a type of prior knowledge, poses a challenge. Some works, like \cite{autoalign,DeepFusion}, avoid using the projection matrix and reduce projection errors by adopting learning methods.

Therefore, based on different methods of feature alignment, we can categorize multi-modal 3D object detection methods into two types: (1) projection-based for feature alignment, and (2) model-based for feature alignment. This taxonomy is more detailed and scientific, better reflecting the characteristics and progress of multi-modal 3D object detection methods in the field of autonomous driving.


\subsection{Projection-based 3D object detection}

\begin{figure*}[htbp]
	\centering
\includegraphics[width=\linewidth]{image/multimodalprojectionV2.pdf}
\caption{Projection-based 3D object detection: (a) Point-Projection-based methods \cite{Pointpainting,wang2021pointaugmenting,xu2021fusionpainting,fusionpatingtgrs,nabati2021centerfusion,shin2019roarnet,simon2019complexer,lasernet,Epnet,epnetpami,sindagi2019mvx,mvp,xie2024ppfDet}, (b) Feature-Projection-based methods \cite{liang2018deep,VoxelNextFusion,focalconv,xu2018pointfusion,vff,LargeKernel3D,SupFusion,C2BN,mmf,lin2022cl3d,song2024robofusion}, 
(c) Auto-Projection-based methods \cite{PI-RCNN,3dcvf,graphalign,graphalign++,autoalignv2,Robust-FusionNet,HMFI,Logonet,3DDualFusion,song2024re,CAT-Det,liu2024menet,wu2024robust}, (d) Decision-Projection-based methods \cite{CLOCs,avod,mv3d,roifusion,Frustumpointnets,Frustumconvnet,Frustum-pointpillars,Fast-CLOCs,Graph-RCNN}. 
Non-Projection-based 3D object detection: 
(e) Query-Learning-based  methods \cite{Transfusion, DeepFusion, DeepInteraction,sparsefusion, autoalign, MixedFusion, QTNet,liu2023multi,zhang2024sparselif}, 
(f) Unified-Feature-based methods \cite{EA-BEV,zheng2023rcfusion, BEVFusion, cai2023bevfusion4d, FocalFormer3D, FUTR3D, SparseFusion3D, UniTR,  virconv, MSMDFusion, sfd, cmt,zeng2021cross, UVTR,ObjectFusion,song2024graphbev,song2024contrastalign,ge2023metabev,yin2024isfusion,zhou2023bridging}.}

\label{fig:projection_pipeline}
\end{figure*}

Projection-based 3D object detection refers to the use of projection matrices during the feature fusion stage to achieve the integration of point cloud and image features. It is important to clarify that the focus here is on projection during the feature fusion period, rather than \textcolor{black}{projections in other stages} of the fusion process, which \textcolor{black}{include} projections needed for processes such as data augmentation. As shown in Fig.\ref{fig:projection_pipeline}, we have developed a more detailed classification of projection-based 3D object detection based on the different types of projection used in the fusion stage, including Point-Projection-based \cite{Pointpainting,wang2021pointaugmenting,xu2021fusionpainting,nabati2021centerfusion,shin2019roarnet,simon2019complexer,lasernet,Epnet,epnetpami,sindagi2019mvx,mvp}, Feature-Projection-based \cite{liang2018deep,VoxelNextFusion,focalconv,xu2018pointfusion,vff,LargeKernel3D,SupFusion,C2BN,mmf}, Auto-Projection-based \cite{PI-RCNN,3dcvf,graphalign,graphalign++,autoalignv2,Robust-FusionNet,HMFI,Logonet,3DDualFusion}, and Decision-Projection-based methods \cite{CLOCs,avod,mv3d,roifusion,Frustumpointnets,Frustumconvnet,Frustum-pointpillars,Fast-CLOCs,Graph-RCNN}.

\subsubsection{\textbf{Point-Projection-based 3D object detection}}

Point-Projection-based 3D object detection methods \cite{Pointpainting,wang2021pointaugmenting,xu2021fusionpainting,nabati2021centerfusion,shin2019roarnet,simon2019complexer,lasernet,mvp} involve projecting image features onto raw point clouds to enhance the representational capability of the original point cloud data, as shown in Fig. \ref{fig:projection_pipeline} (a). The initial step in these methods is to establish a strong correlation between LiDAR points and image pixels, which is achieved using calibration matrices. Following this, the point cloud features are enhanced by augmenting them with additional data. This augmentation takes two forms: either through the incorporation of segmentation scores \cite{Pointpainting,xu2021fusionpainting,fusionpatingtgrs} or by using CNN features \cite{sindagi2019mvx, mvp, Epnet,epnetpami,wang2021pointaugmenting} from the correlated pixels. PointPainting \cite{Pointpainting} and PointAugmenting \cite{wang2021pointaugmenting} represent advancements in multi-modal 3D object detection methods by enhancing the traditional cut-and-paste augmentation. These techniques aim to seamlessly integrate data from different domains, such as point clouds and 2D imagery, while carefully managing potential overlaps or collisions between objects in both domains. PointPainting enhances LiDAR points by appending segmentation scores. However, it has limitations in effectively capturing the color and texture details present in images. To address these shortcomings, more sophisticated approaches like FusionPainting \cite{xu2021fusionpainting} have been developed, following a similar paradigm. MVP \cite{mvp} builds upon the concept of PointPainting \cite{Pointpainting}. It initially utilizes image instance segmentation and establishes an alignment between the segmentation masks and the point cloud using a projection matrix. The key distinction of MVP lies in its approach to sampling: it randomly selects pixels within each range, ensuring consistency with the points in the point cloud. These selected pixels are then linked to their nearest neighbors in the point cloud. The depth value of the LiDAR point in this linkage is assigned as the depth of the corresponding pixel. Subsequently, these points are projected back to the LiDAR coordinate system, resulting in the generation of virtual LiDAR points.

\subsubsection{\textbf{Feature-Projection-based 3D object detection}}

In contrast to Point-Projection-based methods, Feature-Projection-based 3D object detection methods \cite{ContFuse,VoxelNextFusion,focalconv,xu2018pointfusion,vff,LargeKernel3D,SupFusion,C2BN,mmf}, as shown in Fig. \ref{fig:projection_pipeline} (b), primarily focus on fusing point cloud features with image features during the feature extraction phase of the point clouds. During this fusion process, point cloud features are projected onto corresponding image features, and subsequently, these image and point cloud features are integrated together.
This process is achieved by applying a calibration matrix to transform the voxel's three-dimensional coordinate system into the pixel coordinate system of the image, thereby facilitating the effective fusion of point cloud and image modalities. Specifically, the projection of a three-dimensional point cloud onto the image plane can be articulated as follows:
\begin{equation}\label{equ3d2d}
 z_{c}\left[\begin{array}{c} u \\ v \\ 1 \end{array}\right] = h \mathcal{K}\left[\begin{array}{ll} R & T \end{array}\right]\left[\begin{array}{c} P_{x} \\ P_{y} \\ P_{z} \\ 1 \end{array}\right], 
\end{equation}
where, \( P_{x} \), \( P_{y} \), and \( P_{z} \) represent the three-dimensional spatial coordinates of the LiDAR points, while \( u \) and \( v \) denote the corresponding two-dimensional coordinates. The term \( z_{c} \) indicates the depth of the point's projection on the image plane. Additionally, \( \mathcal{K} \) represents the intrinsic parameters of the camera, and \( R \) and \( T \) signify the rotation and translation of the LiDAR relative to the camera's reference frame, respectively. The factor \( h \) accounts for the scale change due to downsampling.

A quintessential example of the Feature-Projection-based method, ContFuse~\cite{ContFuse}, employs continuous convolution to amalgamate multi-scale convolutional feature maps from each sensor. Within this technique, the projection of the point cloud facilitates the correspondence between the image and the Bird's Eye View (BEV). In essence, Feature-Projection-based 3D object detection method is accomplished during the point cloud feature extraction phase. \textcolor{black}{Compared to Point-Projection-based methods, they do not perform fusion on the original point cloud but achieve a profound depth feature fusion, resulting in more robust performance.}

\subsubsection{\textbf{Auto-Projection-based 3D object detection}}
\begin{figure}[htbp]
	\centering
	\includegraphics[width=0.4\linewidth]{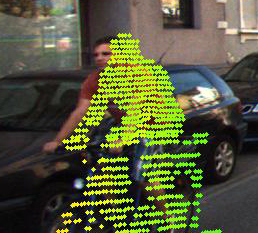}
	\caption{Examples of misalignment between point clouds and images.}
	\label{fig:misalignment}
\end{figure}

As shown in Fig. \ref{fig:misalignment}, a partial image from the KITTI \cite{kitti} dataset exemplifies that projection inaccuracies persist even in this classic clean dataset. Consequently, the issue of projection errors cannot be completely eliminated through manual calibration; instead, they can only be mitigated. This is a frequent challenge in practical dataset deployments. Many studies, like Point \& Feature-Projection-based methods, have performed fusion through direct projection without addressing the projection error issue. A few works \cite{3dcvf,graphalign,graphalign++,autoalignv2,HMFI,Robust-FusionNet,Logonet,3DDualFusion}, have sought to mitigate these errors through approaches such as projection offsets and neighboring projections. For instance, Deformable Cross Attention\cite{deformabledetr} has been employed to learn offsets in the context of already projected data. We have systematically reviewed and synthesized methods that tackle projection errors, designating them as Auto-projection-based 3D object detection methods, as shown in Fig. \ref{fig:projection_pipeline} (c). As representative works addressing feature alignment, HMFI\cite{HMFI}, GraphAlign\cite{graphalign}, and GraphAlign++\cite{graphalign++} utilize a priori knowledge of projection calibration matrices to project onto corresponding images for local graph modeling. This approach simulates intermodal relationships, enabling Multi-modal 3D object detectors to effectively identify more appropriate alignment relationships, thereby achieving faster and more accurate feature alignment between modalities. AutoAlignV2 \cite{autoalignv2} focuses on sparse learnable sampling points for cross-modal relational modeling, enhancing calibration error tolerance and significantly accelerating feature aggregation across different modalities. In summary, Auto-Projection-based 3D object detection methods mitigate errors arising from feature alignment by leveraging neighbor relationships or neighbor \textcolor{black}{offsets}, thereby enhancing robustness in Multi-modal 3D object detection.

\subsubsection{\textbf{Decision-Projection-based 3D object detection}}

Decision-Projection-based 3D object detection methods \cite{CLOCs,avod,mv3d,roifusion,Frustumpointnets,Frustumconvnet,Frustum-pointpillars,Fast-CLOCs,Graph-RCNN}, as early implementations of Multi-modal 3D object detection schemes, use projection matrices to align features in Regions of Interest (RoI) or specific results, as shown in Fig. \ref{fig:projection_pipeline} (d). These methods are primarily focused on the alignment of features in localized areas of interest or specific detection outcomes.

Graph-RCNN \cite{Graph-RCNN} projects the graph node to the location in the camera image and collects the feature vector at that pixel in the camera image through bilinear interpolation. F-PointNet \cite{Frustumpointnets} performs detection on the 2D image to determine the class and localization of the object, and for each detected object, the corresponding point clouds in 3D space are obtained through the conversion matrix of calibrated sensor parameters and 3D space. MV3D \cite{mv3d} employs a transformation of the LiDAR point cloud into Bird's Eye View (BEV) and Front View (FV) projections for generating proposals. During this process, a specialized 3D proposal network is used to create precise 3D candidate boxes. These 3D proposals are then projected onto feature maps from multiple perspectives to facilitate feature alignment between the two modalities. Differing from MV3D \cite{mv3d}, AVOD \cite{avod} streamlines this approach by omitting the FV component and introducing a more refined region proposal mechanism. In summary, Decision-Projection-based 3D object detection methods primarily achieve feature fusion at a high level through projection, with limited \textcolor{black}{interaction} between heterogeneous modalities. This often leads to the alignment and fusion of erroneous features, resulting in issues of reduced accuracy and robustness.

\subsection{Non-Projection-based 3D object detection}

Non-Projection-based 3D object detection methods  achieve fusion without relying on feature alignment, thereby yielding robust feature representations. They circumvent the limitations of camera-to-LiDAR projection, which often reduces the semantic density of camera features and impacts the effectiveness of techniques like Focals Conv \cite{focalconv} and PointPainting \cite{Pointpainting}. Non-Projection-based methods typically employ cross-attention mechanisms or the construction of a unified space to address the inherent misalignment issues in direct feature projection. These methods are primarily divided into two categories: (1) Query-Learning-based \cite{Transfusion, DeepFusion, DeepInteraction, autoalign, CAT-Det, MixedFusion} and (2) Unified-feature-based \cite{EA-BEV,BEVFusion,cai2023bevfusion4d,FocalFormer3D,FUTR3D,UniTR,virconv,MSMDFusion,sfd,cmt,UVTR,sparsefusion}. Query-Learning-based methods entirely negate the need for alignment during the fusion process. Conversely, Unified-Feature-based methods, though constructing a unified feature space, do not completely avoid projection; it usually occurs within a single modality context. For example, BEVFusion \cite{BEVFusion} utilizes LSS \cite{lss}  for camera-to-BEV projection. This process, taking place before fusion, demonstrates considerable robustness in scenarios with feature misalignment.

\subsubsection{\textbf{Query-Learning-based 3D object detection}}

Query-Learning-based 3D object detection methods, as exemplified by works such as \cite{Transfusion, DeepFusion, DeepInteraction, autoalign, CAT-Det, MixedFusion,jiang2024sparseinteraction}, eschew the necessity for projection within the feature fusion process, as shown in Fig. \ref{fig:projection_pipeline} (e). Instead, they attain feature alignment through cross-attention mechanisms before engaging in the fusion of features. Point cloud features are typically employed as queries, while image features serve as keys and values, facilitating a global feature query to acquire highly robust Multi-modal features. Furthermore, DeepInteraction \cite{DeepInteraction} incorporates multimodality interaction, wherein point cloud and image features are utilized as distinct queries to enable further feature interaction. In comparison to the exclusive use of point cloud features as queries, the comprehensive incorporation of image features leads to the acquisition of more resilient Multi-modal features. Overall, Query-Learning-based 3D object detection methods employ a transformer-based structure for feature querying to achieve feature alignment. Ultimately, the Multi-modal features are integrated into LiDAR-only pipelines, such as CenterPoint\cite{Centerpoint}.

\subsubsection{\textbf{Unified-Feature-based 3D object detection}}

Unified-feature-based 3D object detection methods, represented by works such as \cite{EA-BEV,BEVFusion,cai2023bevfusion4d,FocalFormer3D,FUTR3D,UniTR,virconv,MSMDFusion,sfd,cmt,UVTR,sparsefusion}, generally employ projection before feature fusion, achieving the pre-fusion unification of heterogeneous modalities, as shown in Fig. \ref{fig:projection_pipeline} (f). In the BEV fusion series, which utilizes LSS for depth estimation \cite{EA-BEV,BEVFusion,cai2023bevfusion4d,sparsefusion}, the front-view features are transformed into BEV features, followed by the fusion of BEV image and BEV point cloud features. Alternatively, CMT \cite{cmt} and UniTR \cite{UniTR} employ transformers for tokenization of point clouds and images, constructing an implicit unified space through transformer encoding. CMT \cite{cmt} utilizes projection in the position encoding process, but entirely avoids dependency on projection relations at the feature learning level.
FocalFormer3D \cite{FocalFormer3D}, FUTR3D \cite{FUTR3D}, and UVTR \cite{UVTR} leverage transformers' queries to implement schemes similar to DETR3D\cite{wang2022detr3d}, constructing a unified sparse BEV feature space through queries, thus mitigating the instability introduced by direct projection. VirConv\cite{virconv}, MSMDFusion\cite{MSMDFusion}, and SFD\cite{sfd} construct a unified space through pseudo-point clouds, with the projection occurring before feature learning. The issues introduced by direct projection are addressed through subsequent feature learning. In summary, Unified-feature-based 3D object detection methods \cite{EA-BEV,BEVFusion,cai2023bevfusion4d,FocalFormer3D,FUTR3D,UniTR,virconv,MSMDFusion,sfd,cmt,UVTR,sparsefusion} currently represent high-precision and robust solutions. Although they incorporate projection matrices, such projection does not occur between Multi-modal fusion, distinguishing them as Non-Projection-based 3D object detection methods. Unlike Auto-Projection-based 3D object detection approaches, they do not directly address projection error issues but instead opt for unified space construction, considering multiple dimensions for Multi-modal 3D object detection, thereby obtaining highly robust Multi-modal features.

\subsection{Analysis: Accuracy, Latency, Robustness}


In the preceding Sections \ref{Sec:Analysis:Image}, \ref{Sec:Analysis:Lidar}, we have conducted a comprehensive analysis of `Accuracy, Latency, Robustness' for camera-only and LiDAR-only approaches. Subsequently, we extend our examination to multi-modal 3D object detection methods, employing a similar analytical framework.
\subsubsection{\textbf{Accuracy}}

As shown in Fig.\ref{fig:zhu_all} (e) and (f), we conducted comparative evaluations on both the KITTI and nuScenes test datasets. The majority of Projection-based 3D object detection methods have predominantly undergone experimentation on the KITTI dataset, with only a minority extending their evaluation to nuScenes. As shown in Fig.\ref{fig:zhu_all} (e), it is evident that Feature-Projection-based and Auto-Projection-based methods exhibit superior overall performance, while Decision-Projection-based methods, primarily dated prior to 2020, tend to manifest relatively lower Average Precision (AP) metrics. A scant few Non-Projection-based 3D object detection methods, such as CAT-Det \cite{CAT-Det}, have been experimented with on the KITTI dataset. As shown in Fig.\ref{fig:zhu_all} (f), the latest methods predominantly belong to the Unified-Feature-based methods, underscoring the suitability of the panoramic camera offered by nuScenes for achieving modality-unifying strategies like BEVFusion\cite{BEVFusion}. Overall, it is discernible that Non-Projection-based methods present more effective solutions in terms of Accuracy metrics (e.g., AP, mAP, NDS, etc.).
\subsubsection{\textbf{Latency}}
As shown in Table \ref{tab:multimodel_latency}, we conducted a comparative analysis of mono-modal 3D object detection methods (LiDAR-only and Camera-only) and Multi-modal 3D object detection on the KITTI and nuScenes datasets, presenting scatter plots for Latency (FPS) and Accuracy metrics (AP, mAP, NDS, etc.). It is noteworthy that, in comparison to mono-modal 3D object detection methods (LiDAR-only and Camera-only), Multi-modal 3D object detection approaches generally exhibit lower FPS. The results on the KITTI dataset indicate that GraphAlign excels in both AP and FPS metrics. \textcolor{black}{Additionally, LoGoNet \cite{Logonet}, Focals Conv \cite{focalconv}, and EP-Net \cite{Epnet} demonstrate outstanding performance.} GraphAlign \cite{graphalign} maintains its position as having the highest FPS, but its NDS performance is suboptimal on the nuScenes dataset. In contrast, UniTR performs exceptionally well in both NDS and FPS metrics. Overall, it can be observed that within Projection-based methods, Auto-Projection-based and Feature-Projection-based methods exhibit superior overall performance, while within Unified-Feature-based methods, the overall performance is more outstanding. In the meticulous evaluation of the KITTI and nuScenes datasets, emphasis is placed on the trade-off between FPS and NDS metrics.
\subsubsection{\textbf{Robustness}}
In the previous sections \ref{sec:camera_Robustness} and \ref{sec:point_robustness}, we analyzed the robustness of mono-modal 3D object detection (Camera-only and LiDAR-only). In this section, based on Tables \ref{tab:kittic-car-moderate} and \ref{tab_nuscenes_C}, we analyze the robustness of Multi-modal 3D object detection. From KITTI-C \cite{BR3D} and nuScenes-C \cite{BR3D}, it can be seen that Multi-modal 3D object detection is more robust compared to mono-modal 3D object detection (Camera-only and LiDAR-only), with smaller RCE. In KITTI-C, representative articles LoGoNet \cite{Logonet} for Auto-Projection-based and VirConv \cite{virconv} for Unified-Feature-based exhibit greater robustness, while EPNet \cite{Epnet} for Point-Projection-based and Focals Conv \cite{focalconv} for Feature-Projection-based show slightly weaker performance. Additionally, in nuScenes-C, among Non-Projection-based methods, FUTR3D \cite{FUTR3D}, TransFusion \cite{Transfusion}, BEVFusion \cite{BEVFusion}, and DeepInteraction \cite{DeepInteraction} all demonstrate strong robustness. \textcolor{black}{It is worth noting that MetaBEV \cite{ge2023metabev} explores the problem of modal loss caused by feature misalignments and sensor failures in BEV features of LiDAR and camera through deformable attention based on BEVFusion \cite{BEVFusion}. ObjectFusion \cite{ObjectFusion} proposes a novel object-centric fusion  to align object-centric features of different modalities. GraphBEV  \cite{song2024graphbev} mitigates misalignment issues by matching neighbor depth features through graph matching.}
\textcolor{black}{
\section{Future outlooks}
}
\textcolor{black}{
Through reviewing all the literature and analyzing the research trends of the past few years, we make some predictions on the future research direction of 3D object detection from the perspective of robustness.
}
\textcolor{black}{
\subsection{3D Object Detection with Large Models}
Inspired by the success of large language models (LLMs) such as ChatGPT \cite{achiam2023gpt} and vision foundation models (VFMs) like SAM, many researchers have focused on related research of large models. Compared to conventional methods, a large autonomous driving model based on LLM can mainly solve the following two problems. Firstly, it is the endless corner case problem. LLMs have a common sense ability and may become a new paradigm for solving corner cases in autonomous driving problems. Secondly, the current methods lack intuitive reasoning and provide textual explanations, and LLMs happen to be the best in this direction. It is worth researching how to combine large models with 3D object detection to enhance robustness and generalization and improve the ability of corner cases. However, there is currently limited research on combining large models and 3D object detection.
For example, RoboFusion \cite{song2024robofusion} has integrated TransFusion \cite{Transfusion} and Focals Conv \cite{focalconv} with VFMs like SAM \cite{sam} to enhance its ability in harsh weather conditions. SEAL \cite{seal} uses VFMs like SAM \cite{sam} to segment different car point cloud sequences and can segment any car point cloud by encouraging spatial and temporal consistency during the representation learning stage. CLIP-BEVFormer \cite{pan2024clipbevformer} combines CLIP \cite{clip} and BEVFormer \cite{bevformer}, leveraging the universal capabilities of CLIP \cite{clip} to enhance generalization on corner cases. VisLED \cite{greer2024language} is a language-driven active learning framework for open-set 3D object detection, which utilizes active learning techniques to query various information-rich data samples from unlabeled pools. Almost all existing works are proposed and evaluated on close-range datasets. Although these datasets may be large and diverse, they are still insufficient for real-world applications. In the real world, the generalization and robustness of corner cases are of utmost importance, and 3D object detection with large models is a good starting point for solving open-set 3D object detection. In addition, current 3D object detection algorithms lack interpretability, and LLMs can bring hope for more robust 3D object detection and avoid unexpected situations caused by black box detectors.
}
\textcolor{black}{
\subsection{3D Object Detection in End-to-End Autonomous Driving}
UniAD\cite{uniad} undoubtedly brought another hot topic to the field of autonomous driving after winning the CVPR Best Paper Award: end-to-end autonomous driving. End-to-end autonomous driving is a fully differentiable machine learning system that takes raw sensor input data and other metadata as prior information and directly outputs the control signals or trajectory planning for vehicles \cite{xu2024m2da}. Generally, the autonomous drive system integrates multiple tasks, such as detection, tracking, online mapping, motion prediction, and planning. 3D object detection is closely related to other perception tasks and downstream tasks such as prediction and planning. Therefore, pursuing high accuracy in 3D object detection may not be optimal when considering the autonomous drive system. Although impressive progress has been made in end-to-end research, we believe three areas can be further improved in current 3D object detection for end-to-end autonomous driving. First, 3D object detection can guide more effective multi-modal environmental perception, allowing for better data integration from multi-modal sources. Second, the current inference capabilities of end-to-end autonomous driving are concerning. Third, 3D object detection enhanced with large language models (LLMs) provides stronger explanatory power, leading to enhanced explanatory power for subsequent tasks.
}
\section{Conclusion}
\label{section:conclusion}
3D object detection plays a crucial role in autonomous driving perception. In recent years, this field has witnessed rapid development, yielding many research 
results. Based on the diverse data forms generated by sensors, these methods are primarily categorized into three types: image-based, point cloud-based, and multi-modal. The primary metrics for evaluation in these methods are high accuracy and low latency. Numerous reviews have summarized these approaches, focusing on the core principles of `high accuracy and low latency' in delineating their technical trajectories.
However, in the transition of autonomous driving technology from breakthroughs to practical applications, existing reviews have not prioritized safety perception as a central concern, failing to encompass the current technological pathways related to safety perception. For instance, recent Multi-modal fusion methods typically undergo robustness testing during the experimental phase, a facet not adequately considered in current reviews.
Therefore, in this study, we re-examine 3D object detection algorithms with a central focus on the key aspects of `Accuracy, Latency, and Robustness'. We reclassify previous reviews, placing particular emphasis on re-segmenting from the perspective of safety perception. We aim for this work to offer new insights for future research in 3D object detection, transcending the confines of high-accuracy exploration.

\addtolength{\textheight}{-0cm}



\bibliographystyle{IEEEtran}
\bibliography{IEEEabrv,IROS}\ 
%
%
%

\begin{IEEEbiography}[{\includegraphics[width=1in,height=1.25in,clip,keepaspectratio]{{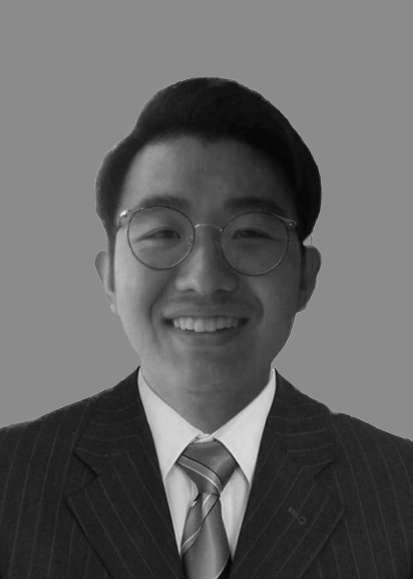}}}]{Ziying Song}, was born in Xingtai, Hebei Province, China in 1997. He received the B.S. degree from Hebei Normal University of Science and Technology (China) in 2019. He received a master's degree major in Hebei 
University of Science and Technology (China) in 2022. He is now a PhD student majoring in Computer Science and Technology at Beijing Jiaotong University (China), with a research focus on Computer Vision. 
\end{IEEEbiography}

\begin{IEEEbiography}
[{\includegraphics[width=1in,height=1.25in,clip,keepaspectratio]{{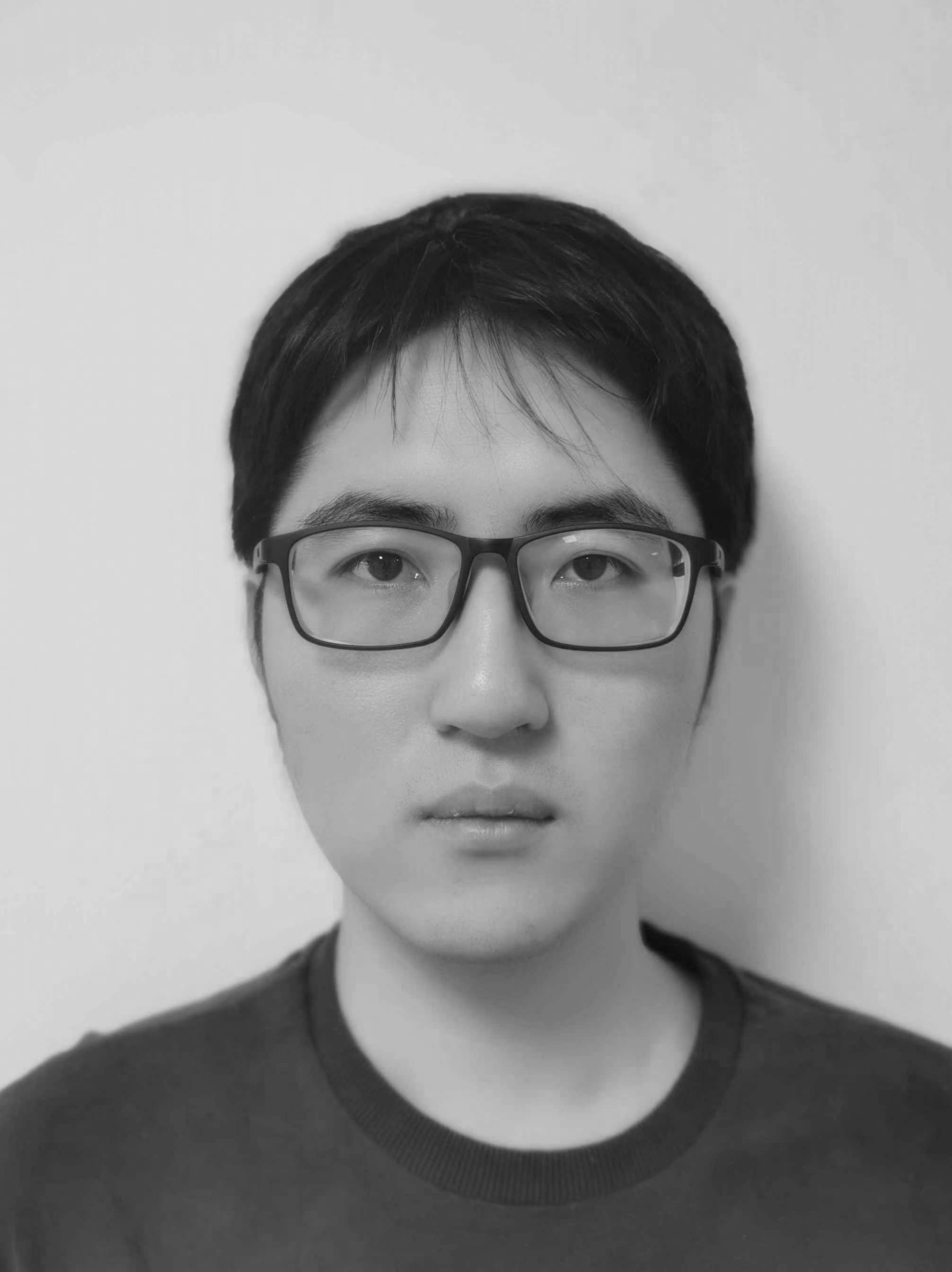}}}] {Lin Liu} was born in Jinzhou, Liaoning Province, China, in 2001. He is now a college student majoring in Computer Science and Technology at China University of Geosciences(Beijing).
Since Dec. 2022, he has been recommended for a master's degree in Computer Science and Technology at Beijing Jiaotong University. His research interests are in computer vision.
\end{IEEEbiography}

\begin{IEEEbiography}
[{\includegraphics[width=1in,height=1.25in,clip,keepaspectratio]{{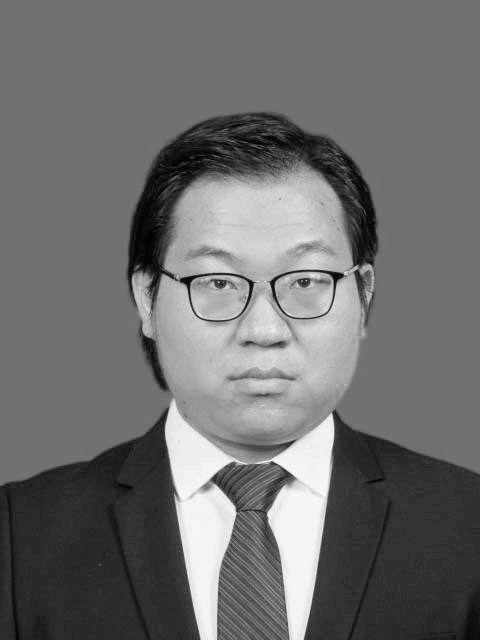}}}]
{Feiyang Jia} was born in Yinchuan, Ningxia Province, China, in 1998. He received his B.S. degree from Beijing Jiaotong University (China) in 2020. He received a master's degree from Beijing Technology and Business University (China) in 2023. He is now a Ph.D. student majoring in Computer Science and Technology at Beijing Jiaotong University (China), with research focus on Computer Vision.
\end{IEEEbiography}

\begin{IEEEbiography}[{\includegraphics[width=1in,height=1.25in,clip,keepaspectratio]{{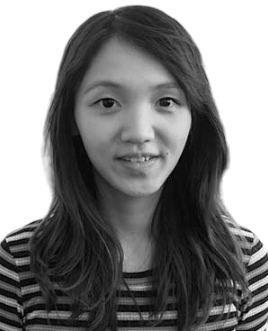}}}]{Yadan Luo} (Member, IEEE) received the BS degree in computer science from the University of Electronic Engineering and Technology of China, and the PhD degree from the University of Queensland. Her research interests include machine learning, computer vision, and multimedia data analysis. She is now a lecturer with the University of Queensland.
\end{IEEEbiography}

\begin{IEEEbiography}[{\includegraphics[width=1in,height=1.25in,clip,keepaspectratio]{{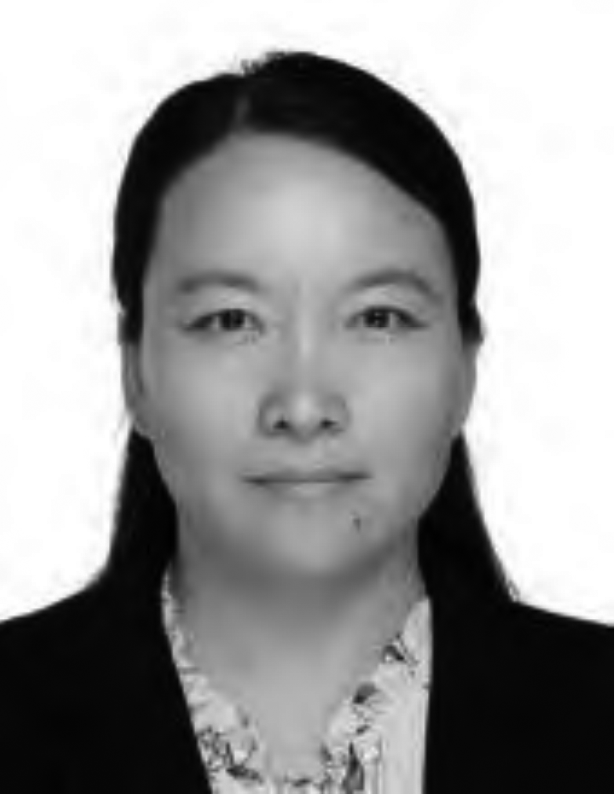}}}]{Caiyan Jia}, born on March 2, 1976, is a lecturer and a postdoctoral fellow of the Chinese Computer Society. she graduated from Ningxia University in 1998 with a bachelor's degree in mathematics, Xiangtan University in 2001 with a master's degree in computational mathematics, specializing in intelligent information processing, and the Institute of Computing Technology of the Chinese Academy of Sciences in 2004 with a doctorate degree in engineering, specializing in data mining. she has received her D. degree in 2004. She is now a
professor in School of Computer Science and
Technology, Beijing Jiaotong University, Beijing,
China.
\end{IEEEbiography}

\begin{IEEEbiography}[{\includegraphics[width=1in,height=1.25in,clip,keepaspectratio]{{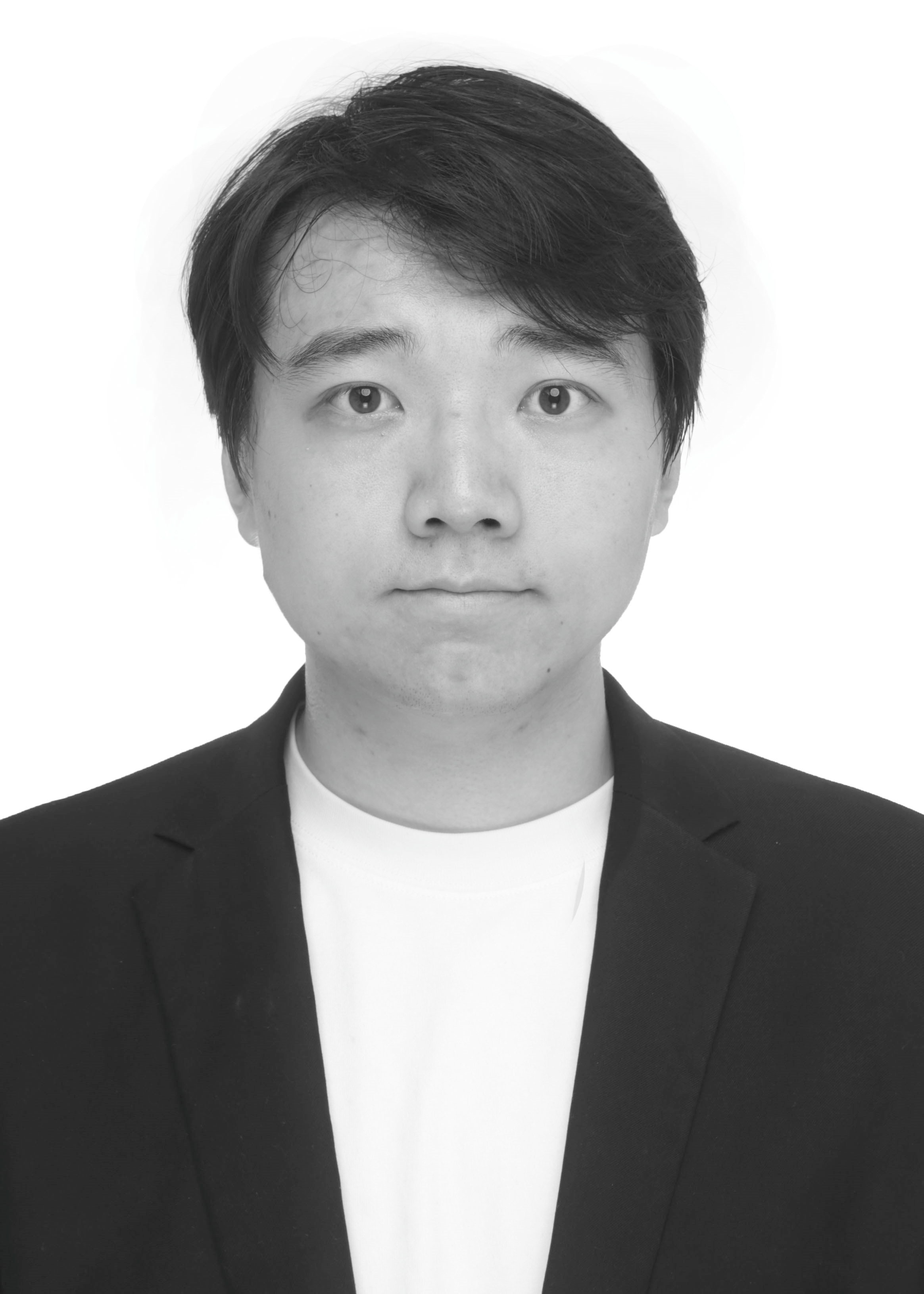}}}]{Guoxin Zhang}, was born in 1998 in Xingtai, Hebei Province, China. He received his bachelor's and Master's degrees from Hebei University of Science and Technology in 2021 and 2024, respectively. He is now a Ph.D. student in the School of Computer Science at Beijing University of Posts and Telecommunications (China) since 2024. His research interests are in computer vision.
\end{IEEEbiography}

\begin{IEEEbiography}[{\includegraphics[width=1in,height=1.25in,clip,keepaspectratio]{{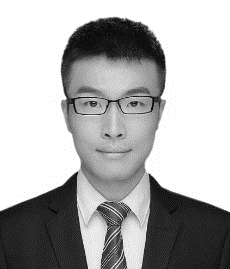}}}]{Lei Yang (Graduate Student Member, IEEE)} received his B.E. degree from Taiyuan University of Technology, Taiyuan, China,
and the M.S. degree from the Robotics Institute at Beihang University, in 2018. Then he joined the Autonomous Driving R$\&$D Department of JD.COM as an algorithm researcher from 2018 to 2020. He is now a Ph.D. student in the School of Vehicle and Mobility at Tsinghua University since 2020.
His current research interests include computer vision, 3D scene understanding and autonomous driving.
\end{IEEEbiography}

\begin{IEEEbiography}[{\includegraphics[width=1in,height=1.25in,clip,keepaspectratio]{{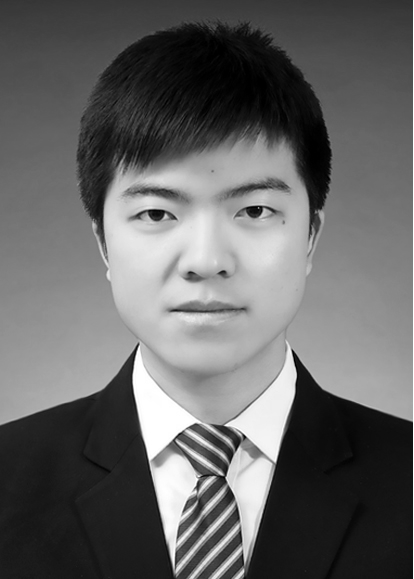}}}]{Li Wang} was born in Shangqiu, Henan Province, China in 1990. He received his Ph.D. degree in mechatronic engineering at the State Key Laboratory of Robotics and System, Harbin Institute of Technology, in 2020. He was a visiting scholar at Nanyang Technology of University for two years. He was a postdoctoral fellow in the State Key Laboratory of Automotive Safety and Energy, and the School of Vehicle and Mobility, Tsinghua University. Currently, he is an assistant professor at School of Mechanical Engineering, Beijing Institute of Technology. His research interests include autonomous driving perception, 3D robot vision, and Multi-modal fusion.
\end{IEEEbiography}



\end{document}